\documentclass{article} %
\usepackage{iclr2027_conference,times}

\usepackage{amsmath,amsfonts,bm}

\def\eqref#1{equation~\ref{#1}}

\def\1{\bm{1}}

\def\rvz{{\mathbf{z}}}

\DeclareMathAlphabet{\mathsfit}{\encodingdefault}{\sfdefault}{m}{sl}
\SetMathAlphabet{\mathsfit}{bold}{\encodingdefault}{\sfdefault}{bx}{n}

\usepackage[utf8]{inputenc}
\usepackage[T1]{fontenc}
\usepackage{booktabs}
\usepackage{amsfonts}
\usepackage{microtype}
\usepackage{xcolor}
\usepackage{graphicx}
\usepackage{colortbl}
\usepackage{caption}
\usepackage{wrapfig}
\usepackage{algorithm}
\usepackage{algorithmic}
\usepackage{hyperref}
\usepackage{url}
\usepackage{etoolbox}

\definecolor{darkblue}{rgb}{0, 0, 0.5}
\definecolor{oodgray}{gray}{0.93}
\newcommand{\cifont}{\scriptsize}
\newcommand{\ci}[3]{#1\,{\cifont[#2,\,#3]}}
\newcommand{\cis}[3]{\begin{tabular}[c]{@{}r@{}}#1\\[-2.5pt]{\cifont[#2,\,#3]}\end{tabular}}
\hypersetup{colorlinks=true, citecolor=darkblue, linkcolor=darkblue, urlcolor=darkblue}

\iclrfinalcopy
\makeatletter
\patchcmd{\@maketitle}{Published as a conference paper at ICLR 2027}{Preprint}{}{\PackageWarning{main}{ICLR header patch failed}}
\makeatother

\title{Probabilistic Calibration Is a Trainable Capability in Language Models}

\author{Davide Baldelli$^{1,2,3,4}$\thanks{Corresponding author: \texttt{davide.baldelli@mila.quebec}.} \quad
  Sruthi Kuriakose$^{6}$ \quad
  Maryam Hashemzadeh$^{1,2,5}$ \\
  \bf Amal Zouaq$^{2,3,4}$ \quad
  Sarath Chandar$^{1,2,4}$ \\[0.5em]
  \normalfont $^{1}$Chandar Research Lab \quad $^{2}$Mila -- Quebec AI Institute \quad $^{3}$LAMA-WeST Lab \\
  \normalfont $^{4}$Polytechnique Montr\'eal \quad $^{5}$Universit\'e de Montr\'eal \quad $^{6}$Independent researcher}

\begin{document}

\maketitle

\begin{abstract}
Language models are increasingly used in settings where outputs should follow an intended distribution, yet their generation probabilities are often poorly calibrated to it. In many such requests the distribution is implicit and the valid outputs cannot be listed in advance, so the task cannot be passed to an external random number generator. We study whether this capability can be improved directly through fine-tuning. Concretely, we fine-tune language models on synthetic prompts that require sampling from mathematical distributions, and compare two Calibration Fine-Tuning variants: a soft-target method that converts the desired output distribution into trie-derived next-token targets, and a hard-target method that trains on sampled completions from the same target distribution. Across 12 models spanning four families, both methods substantially improve structured-sampling fidelity on held-out distribution families and unseen parameter settings, showing that training on cheap synthetic mathematical distributions improves sampling on unseen distributions, even in natural language. The gains also transfer to broader stochastic generation benchmarks, including open-ended random generation, where fine-tuned models broaden their first-token support by one to two orders of magnitude, and NoveltyBench. The gains sometimes reduce downstream capability, especially arithmetic reasoning, with costs varying by model. Overall, our results show that probabilistic calibration can be improved through fine-tuning with either objective, and that calibration learned on mathematical distributions transfers to natural-language stochastic generation. Code is available at \url{https://github.com/chandar-lab/calibration-finetuning}.

\end{abstract}

\section{Introduction}

Language models are increasingly used as general-purpose interfaces for generation and decision-making \citep{chatterji2025people}. In many such uses, producing a valid answer is not enough: when the request involves randomness, diversity, or balanced choice, the model should place probability mass according to the intended stochastic behavior. In formal spaces, calibrated sampling is a standard primitive: one calls routines like \texttt{np.random.normal(...)} to obtain draws from a specified distribution. There is no equally reliable primitive for language space. In practice, language models are the natural candidate interface for requests such as ``tell me a random city'' or ``tell me a random number from 1 to 10,'' but their induced output distributions are poorly controlled. For such requests the distribution can be implicit and the valid outputs cannot be listed in advance, so the task cannot be passed to an external random number generator.

Current pretraining and post-training objectives do not incentivize calibrated output probabilities: the former optimizes next-token prediction on observed text, and the latter rewards task success, compliance, or preferred responses. Unsurprisingly, language models collapse onto narrow output subsets, violate simple support constraints, and induce distributions that are far from the requested target \citep{renda2023random,zhao2026dice,gu2026illusion}. Prompting workarounds can help but add inference-time cost and prompt sensitivity \citep{misaki2025stringseed,xiao2025flipping}.

This raises a more specific question: can fine-tuning make language models better calibrated probabilistic samplers, so that their samples reflect the intended randomness of the request? In this work, we study a simple setting in which this intended randomness is specified exactly by a known target distribution. We train language models on synthetic prompts that require sampling from mathematical distributions, and compare two Calibration Fine-Tuning strategies for teaching this behavior. The first is \emph{soft-target} fine-tuning: we enumerate the valid numeric outputs for a prompt, assign each output its target probability, and convert this distribution over full answers into supervised probabilities for the next token at each prefix. The second is \emph{hard-target} fine-tuning: we draw many numeric answers from the target distribution and train on them with standard next-token cross-entropy.

Our main finding is that fine-tuning on cheap synthetic mathematical distributions improves sampling on unseen distributions, even in natural language. Across 12 models, both soft-target and hard-target fine-tuning substantially improve structured-sampling fidelity on held-out distribution families and unseen parameter settings. The gains also extend beyond the training domain. Both variants broaden first-token support in open-ended random generation and improve NoveltyBench utility for most models, while perplexity on text from diverse domains is largely preserved.

In summary, we make the following contributions:
\begin{itemize}
    \item We study probabilistic calibration as a trainable capability of language models, framing the problem as learning both the valid output space and the probability mass assigned across it.
    \item We introduce and compare two fine-tuning strategies for this setting: a soft-target method based on trie-induced next-token supervision, and a hard-target method based on repeated sampled completions from the same target distribution.
    \item Across 12 models, we show that both methods substantially improve structured-sampling fidelity on held-out families and unseen parameter settings, reducing normalized Wasserstein error by a median factor of two to four.
    \item We show that these gains transfer beyond the synthetic training domain, to open-ended random generation and NoveltyBench \citep{zhang2025noveltybench} and partly to multiple-choice question (MCQ) answer-position balance \citep{zhao2026dice}, while retention and perplexity results reveal model-dependent tradeoffs under the selected training budgets.
\end{itemize}

\begin{figure*}[t]
\centering
\includegraphics[width=\textwidth]{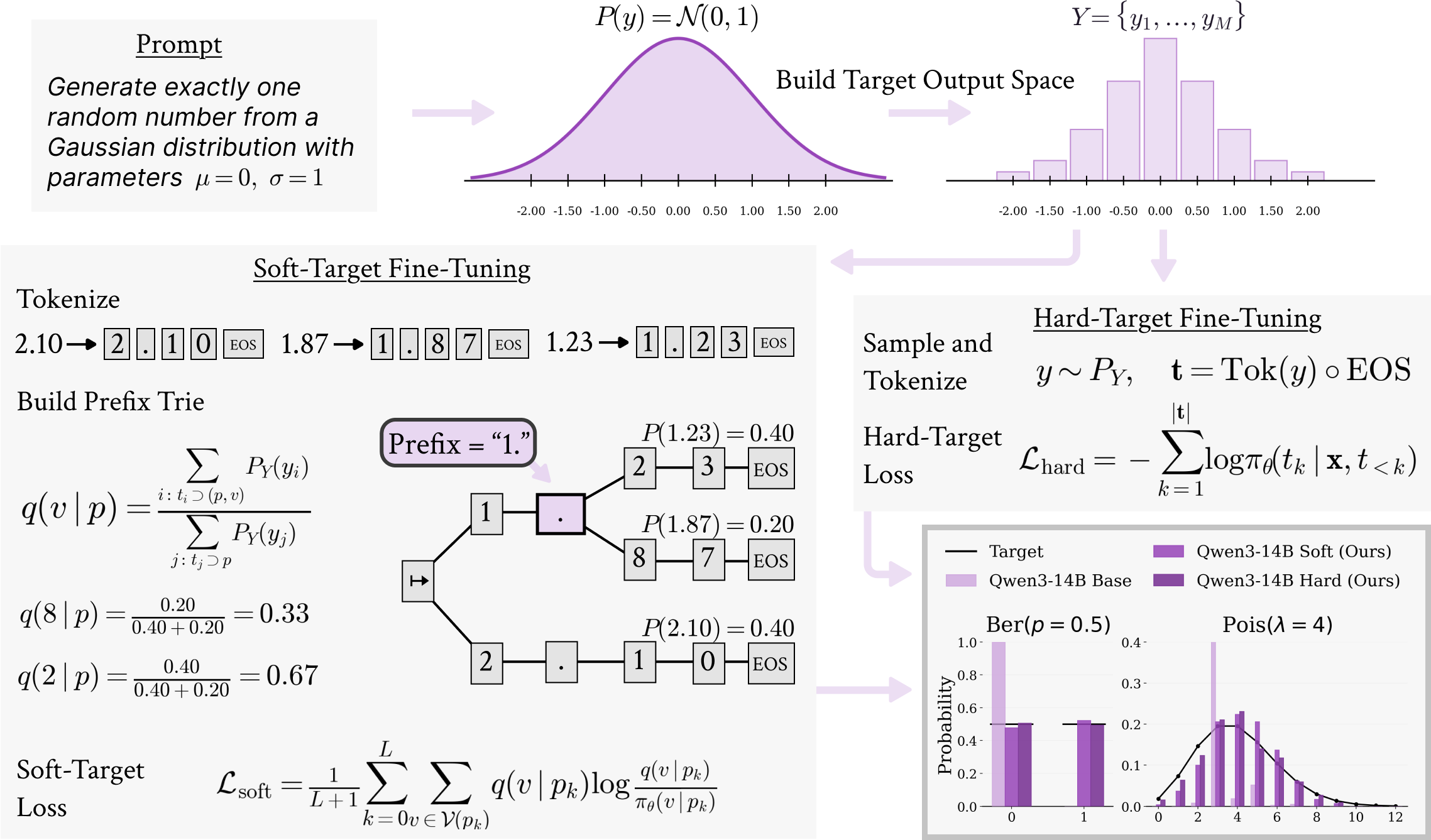}

\caption{\textbf{Calibration Fine-Tuning pipeline.} A prompt defines a target law \(P\), discretized into a canonical output space \(Y\) with masses \(P_Y\). Soft-target fine-tuning builds a token trie and matches the induced next-token distribution with \(\mathcal{L}_{\mathrm{soft}}\). Hard-target fine-tuning samples \(y \sim P_Y\), tokenizes it, and applies masked completion cross-entropy \(\mathcal{L}_{\mathrm{hard}}\). In both variants, only the LoRA adapter is updated.}
\label{fig:method}
\end{figure*}

\section{Related Work}
\label{sec:related}

\paragraph{Language models as poor native samplers, and prompting workarounds.}
A growing line of work shows that autoregressive language models are weak stochastic generators. They deviate from target distributions even in simple synthetic settings \citep{renda2023random}, more so under independent single-sample prompting \citep{zhao2026dice}, and frontier models that can transform provided random seeds into a target distribution still fail to sample from it directly \citep{gu2026illusion}. Prompting-based interventions inject extra reasoning or synthetic entropy \citep{misaki2025stringseed,xiao2025flipping}, but at inference time and with added prompt sensitivity and cost. We instead train the model itself to match a known target distribution.

\paragraph{Calibration, soft targets, and probability-shaping objectives.}
We use calibration in a distributional sense, meaning alignment of output probabilities with a target statistical distribution rather than epistemic uncertainty calibration \citep{kapoor-etal-2024-calibration}. Our objectives belong to a family that shapes predictive distributions through soft supervision or calibration-aware training \citep{lee2022adaptivelabel,kim2025teachercalibration,kirk2023rlhf,parikh2026catto}, and \citet{hu2024amortizing} fine-tune language models to match intractable posterior distributions with generative flow networks. Closest to our setting, \citet{zhang2024forcing} fine-tune language models with sequence-level targets to produce diffuse distributions over valid outputs, and we run their method as a baseline (Appendix~\ref{app:zhang}). We extend distributional fine-tuning to a broad family of probability distributions and evaluate transfer to natural-language stochastic behavior together with capability retention.

\paragraph{Diversity, creativity, and epistemic breadth.}
Our transfer evaluations connect to work on diversity, creativity, and epistemic breadth in generative models \citep{mohammadi2024creativity,doshi2024collectivediversity,holzner2025genaicreativity,luo2026creativitydiversity,wright2025epistemicdiversity,zhang2025noveltybench}, which we use as a downstream test of whether calibration learned on mathematical distributions transfers to natural-language random generation.

\section{Calibration Fine-Tuning}
\label{sec:method}

Figure~\ref{fig:method} summarizes Calibration Fine-Tuning. Both variants start from the same synthetic sampling prompts and the same induced discrete target over a canonical output space, but differ in supervision: soft-target fine-tuning matches trie-induced next-token distributions, while hard-target fine-tuning imitates sampled canonical completions from the same discrete target.

\paragraph{Task setup.}

We train on prompts of the form:
\begin{quote}
\small
Generate exactly ONE random number from a [distribution] distribution with parameters [params]. Output ONLY the number.
\end{quote}
When applicable, this request is wrapped in the model's native chat template; for reasoning-capable models such as Qwen3 and GPT-OSS, we also disable reasoning traces so that training targets only the sampled numeric output.

Each prompt specifies a distribution family and parameters, inducing a target probability law $P$ over the reals or integers. To make supervision tractable, we discretize this law into a finite canonical output space $Y = \{y_1, \dots, y_M\}$ of valid numeric strings together with an induced discrete distribution $P_Y$ over $Y$. For integer-valued families, $Y$ is the finite support when available and otherwise a quantile-truncated support set. For continuous families, we quantize a quantile-bounded interval at fixed decimal precision; when the resulting decimal grid exceeds the output-space cap, we keep evenly spaced grid points, compute bin masses from CDF differences between adjacent midpoints, and assign outside-interval tail mass to the edge bins. In the final training runs, both methods use five-decimal canonical outputs; soft-target fine-tuning caps $Y$ at 1001 bins, while hard-target fine-tuning uses 16384 bins, following the ablations in Appendix~\ref{sec:appendix-ablations}.

\paragraph{Soft-target fine-tuning.}

For each canonical output $y_i \in Y$, we take its induced mass $P_Y(y_i)$, tokenize it, append \textsc{eos}, and insert the resulting token sequence into a prefix trie. For a given trie prefix $p$ (a partial token sequence), we normalize the remaining mass over all continuations consistent with $p$ to obtain a target distribution over valid next tokens:

\begin{equation}
q(v \mid p)
=
\frac{\sum_{i: \, t_i \text{ extends } (p, v)} P_Y(y_i)}
{\sum_{j: \, t_j \text{ extends } p} P_Y(y_j)}.
\end{equation}

Let $\pi_\theta(v \mid p) := \mathrm{softmax}(\rvz_p / \tau)_v$ denote the model's next-token distribution at prefix $p$, where $\rvz_p$ are the model's logits at $p$ and $\tau$ is the sampling temperature. We train the model to match the trie-derived target $q(\cdot \mid p)$ at each visited prefix by minimizing
\begin{equation}
\label{eq:cal-loss}
\ell(p)
=
\mathrm{KL}\!\bigl(
  q(\cdot \mid p)
  \;\big\|\;
  \pi_\theta(\cdot \mid p)
\bigr)
=
\sum_{v \in \mathcal{V}(p)}
q(v \mid p)\,
\log \frac{q(v \mid p)}{\pi_\theta(v \mid p)} ,
\end{equation}
where $\mathcal{V}(p)$ is the set of trie children at $p$.
To form the training loss, we sample a canonical output $y \sim P_Y$, tokenize it as $\mathbf{t}=\mathrm{Tok}(y)\circ\textsc{eos}=(t_1,\dots,t_L,\textsc{eos})$, and average the prefix loss along this sampled path:
\begin{equation}
\label{eq:cal-loss-full}
\mathcal{L}_{\mathrm{soft}}
=
\frac{1}{L+1}\sum_{k=0}^{L}\ell\bigl((t_1,\dots,t_k)\bigr).
\end{equation}

\paragraph{Hard-target fine-tuning.}
The soft-target objective gives dense distributional supervision, but it requires constructing next-token targets for every trie prefix. We therefore also study a second strategy closer to standard supervised fine-tuning: can the model learn the same sampling behavior from completions sampled from the target distribution? For each instance, we draw $y \sim P_Y$, form $\mathbf{t}=\mathrm{Tok}(y)\circ\textsc{eos}$, and optimize masked autoregressive cross-entropy on the concatenated prompt-completion sequence:
\begin{equation}
\mathcal{L}_{\mathrm{hard}}
=
\frac{1}{|\mathbf{t}|}
\sum_{k=1}^{|\mathbf{t}|}
-\log \pi_\theta\!\left(t_k \mid \mathbf{x}, t_{<k}\right),
\end{equation}
where $\mathbf{x}$ are the prompt tokens and loss is applied only to completion tokens. Since each example provides only one sampled path, we repeat prompts multiple times per epoch with independently resampled completions.

Both objectives are implemented with frozen-base LoRA adapters; training details and procedural summaries are given in Section~\ref{sec:implementation-details}, Appendix~\ref{app:implementation}, and Appendix Algorithms~\ref{alg:calibrate-sft}--\ref{alg:hard-label-sft}. The key difference is supervision granularity: a soft-target example specifies the full target at every prefix, while a hard-target example is a single draw, so hard-target fine-tuning must revisit each prompt with fresh draws and uses roughly ten times more updates. Appendix~\ref{sec:appendix-ablations} reports the ablations that select each variant's discretization and the hard-target sampling budget, and Appendix~\ref{app:matched-budget} compares the two methods at a matched optimizer budget.

\section{Experimental Setup}
\label{sec:setup}

\subsection{Implementation Details}
\label{sec:implementation-details}

\paragraph{Data.}
The training data is fully synthetic. Our benchmark spans three tiers of increasing distributional complexity and includes 36 distribution families in total: 24 seen families used for training and 12 held-out families reserved for evaluation only, each evaluated at three parameter settings, for 36 held-out prompts. The held-out families are Bernoulli, Poisson, Maxwell, TruncNorm, Chi, Weibull, Gaussian mixtures with two, three, and four modes, a bimodal Poisson mixture, a bounded Zipf law, and a categorical distribution over six outcomes with specified probabilities. The held-out families are more or less similar to the training families. Some are variants of a training family, such as TruncNorm, which is a Gaussian truncated to an interval. Others are completely new, such as the four-mode Gaussian mixture: no training family has more than one peak inside its range, while this one has four, at locations given in the prompt. The full benchmark, parameter ranges, and train/test splits are listed in Appendix Table~\ref{tab:benchmark-distributions}.

For each seen training family, we discretize the parameter space on a fixed grid, yielding 1988 prompt configurations across all families. Each prompt is paired with a canonical output space derived from its target distribution (Section~\ref{sec:method}). Training batches use family-balanced ordering. Appendix~\ref{sec:appendix-ablations} discusses the corresponding discretization and training-budget ablations.

\paragraph{Models.}
We train and evaluate four model families spanning roughly 0.6B to 27B parameters: Qwen3 \citep{yang2025qwen3} at 0.6B, 1.7B, 4B, 8B, and 14B; Gemma-3-it \citep{gemmateam2025gemma3technicalreport} at 1B, 4B, 12B, and 27B; Llama-3.2-Instruct \citep{grattafiori2024llama3herdmodels} at 1B and 3B; and GPT-OSS \citep{agarwal2025gpt} at 20B. For every model, we report three conditions under the same prompting and decoding settings (temperature 1, top-$p=1$, independent single-sample requests): the original checkpoint (Base), a soft-target Calibration Fine-Tuning adapter (Soft), and a hard-target Calibration Fine-Tuning adapter (Hard).

\paragraph{Details.}
Both methods use frozen-base LoRA adapters with rank 16, alpha 32, and dropout 0.05, applied to the query, key, value, and output attention projections. We optimize with AdamW using learning rate $2\times 10^{-4}$, weight decay 0.01, and a cosine schedule with 3\% linear warmup; all runs use maximum sequence length 256, per-device batch size 8, and gradient accumulation 1 on 4 A100 GPUs. Soft-target fine-tuning trains for 3 epochs, corresponding to 189 optimizer steps per model. Because hard-target supervision is sparser, hard-target fine-tuning uses 16 sampled completions per prompt per epoch and trains for 2 epochs, yielding 1988 optimizer steps per model. Appendix~\ref{app:implementation} provides the remaining implementation details.

\subsection{Evaluation Axes}

We organize evaluation around six axes, moving from probabilistic sampling in mathematical spaces, to natural-language stochastic behavior, novelty and creativity, and capability retention.

\paragraph{Structured distribution sampling.}
This is the primary benchmark. Using the same prompt format introduced in Section~\ref{sec:method}, we evaluate each model along two complementary axes. At the \textit{logit level}, we compute the forward KL between the model's next-token distribution and the trie-derived target \(q(v \mid p)\), averaging over prefixes along 4 Monte Carlo sampled candidate paths per prompt. For all reported results, this target is built from a shared high-resolution evaluation output space with five-decimal canonical outputs and max bins \(=16384\). This common reference makes logit KL comparable across Base, Soft, and Hard, but is stricter than the 1001-bin output space used to train the soft-target configuration. At the \textit{sample level}, we draw 1{,}000 independent generations per prompt, parse valid numeric outputs, and compare the resulting empirical samples against the reference SciPy distribution using valid rate and Wasserstein-1 distance,
\begin{equation}
\label{eq:wasserstein}
W_1 = \frac{1}{N}\sum_{i=1}^{N} \left| x_{(i)} - {F^*}^{-1}\!\left(\tfrac{i - 0.5}{N}\right) \right|,
\end{equation}
where \(N\) is the number of valid samples, \(x_{(i)}\) is the \(i\)-th order statistic, and \({F^*}^{-1}\) is the target quantile function. When no valid numeric sample is obtained for a prompt configuration, the corresponding \(W_1\) is undefined. In aggregate summaries (Figure~\ref{fig:main-structured-sampling} and Appendix Tables~\ref{tab:main-ood}--\ref{tab:main-unseen}), we divide each prompt's \(W_1\) by the target \(Q_{95}-Q_{05}\) width, average finite estimates within each family, and report the median across families whenever every family has at least one finite estimate. We also evaluate String Seed of Thought (SSOT) prompting \citep{misaki2025stringseed} as an inference-time baseline under the same sample-level metric, with details reported in Appendix~\ref{sec:appendix-ssot}. As a training-time baseline, we compare against \citet{zhang2024forcing} in both directions: their recipe on our benchmark, and ours on theirs (Appendix~\ref{app:zhang}).

\paragraph{Open-ended random generation.}
We test whether fine-tuning broadens stochastic support and increases empirical diversity on a 102-prompt benchmark of open-ended random-generation requests that we constructed for this work, spanning categories such as names, cities, animals, foods, chemical elements, medical concepts, and landmarks. Prompt wording varies across formulations such as ``Think of,'' ``Choose,'' ``Name,'' and ``Pick,'' while consistently enforcing a strict output contract (``Output ONLY the answer''); the full prompt set is given in Appendix~\ref{app:prompts-open}. For each prompt, we collect 100 independent samples and measure two quantities: the top-90\% support size, i.e., the number of first-step next tokens required to cover 90\% of the model's probability mass, and the unique-output rate, i.e., the fraction of distinct samples after normalization.

\paragraph{NoveltyBench.}
We also evaluate open-ended generative breadth on NoveltyBench \citep{zhang2025noveltybench} using its curated and WildChat splits (100 and 1{,}000 prompts, respectively). To remain faithful to the original benchmark, we draw 10 generations per prompt with temperature 1 and top-$p=1$, partition responses with the benchmark's classifier-based procedure (a lexical short-circuit for very short answers and otherwise \texttt{yimingzhang/deberta-v3-large-generation-similarity}), and score them with \texttt{Skywork/Skywork-Reward-Gemma-2-27B-v0.2}. Raw rewards are converted to the benchmark's 1--10 ratings, and utility credits only the first response in each semantic partition under the benchmark's patience-based discounting. We report two split-level summary metrics: \emph{mean distinct}, the average number of semantic partitions per prompt, and \emph{mean utility}, the benchmark's reward-weighted summary.

\paragraph{MCQ generation with uniform answer positions.}
We adopt the MCQ answer-position balancing protocol of \citet{zhao2026dice}. Models are asked to generate medical multiple-choice questions under a prompt that encourages approximately uniform placement of the correct answer among A/B/C/D; the full prompt is given in Appendix~\ref{app:mcq-prompt}. We generate 1{,}000 independent MCQs per model and first report the MCQ parse rate, i.e., the fraction of generations that can be parsed as MCQs with a question, four A/B/C/D options, and a correct-answer field in A/B/C/D. Among these parseable generations, we then compute the total variation (TV) distance between the empirical answer-position frequencies and the uniform distribution.

\paragraph{Capability retention.}
We evaluate whether fine-tuning degrades general capabilities using the TinyBenchmarks suite \citep{polo2024tinybenchmarksevaluatingllmsfewer}: tinyMMLU, tinyHellaSwag, tinyTruthfulQA, tinyWinoGrande, and tinyGSM8K, each consisting of 100 items drawn from the full benchmark. We report the gp-IRT (Generalized Performance Item Response Theory) aggregate score, which combines observed per-item accuracy with IRT-based extrapolation to estimate performance on the full benchmark from the 100-item subset. All retention evaluations use greedy decoding.

\paragraph{PALOMA perplexity.}
We report perplexity on text from diverse domains using PALOMA \citep{magnusson2024paloma}, together with bits-per-byte for tokenizer-robust comparison; details in Appendix~\ref{sec:appendix-paloma}.

\section{Main Results}
\label{sec:results}

\begin{figure*}[t]
\centering
\includegraphics[width=\textwidth]{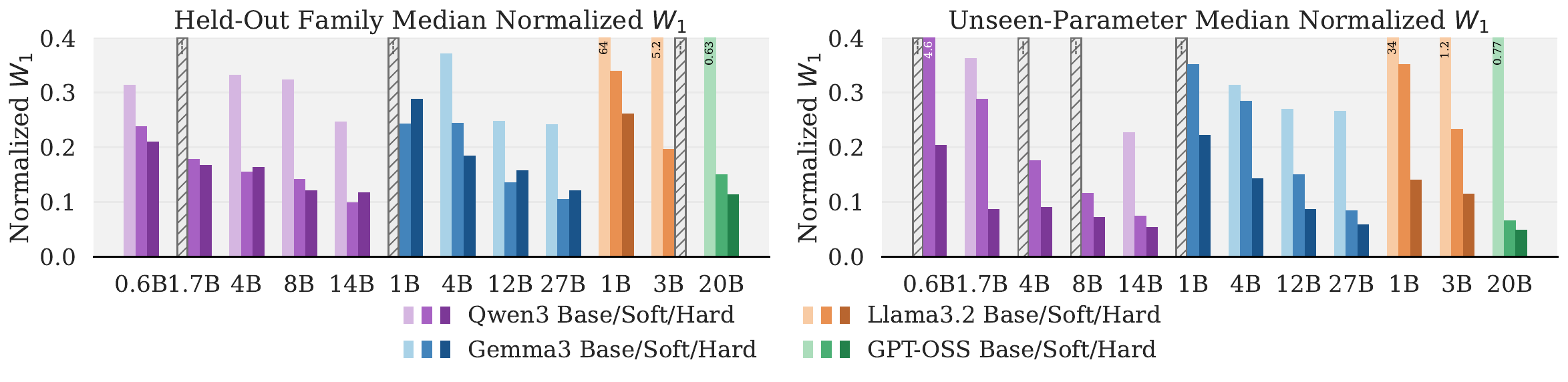}
\caption{\textbf{Main structured-sampling result.} Normalized $W_1$, averaged within family, median across families. \textbf{Left}: the 12 held-out families; \textbf{right}: unseen parameter settings of the 24 training families. Clipped bars are labeled with their value; hatched bars mark a model with no valid sample on some family.}
\label{fig:main-structured-sampling}
\end{figure*}

\paragraph{Structured sampling.}

\begin{figure*}[t]
\centering
\includegraphics[width=\textwidth]{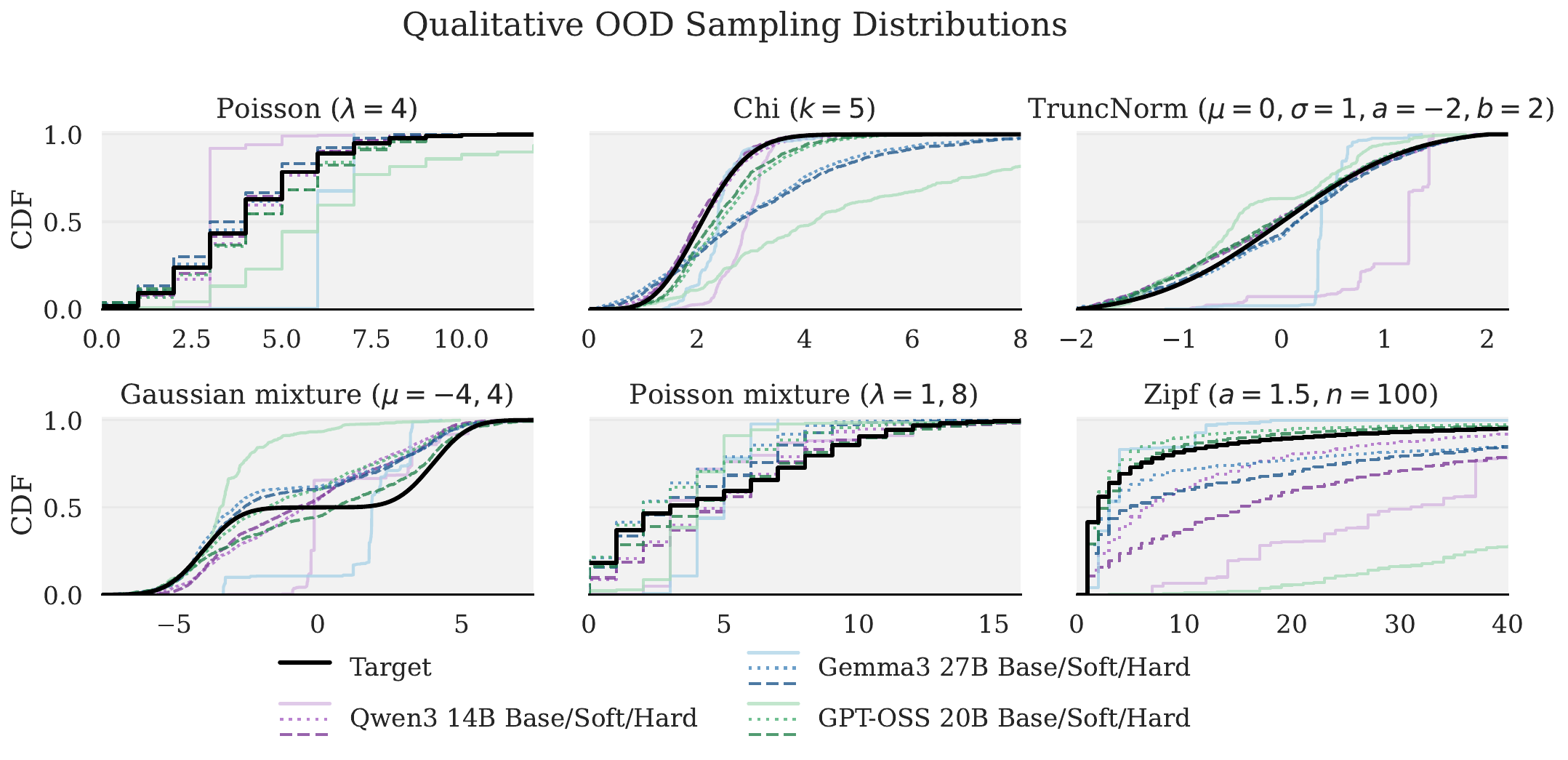}
\caption{\textbf{Qualitative held-out sampling examples.} Each panel overlays the target CDF with samples from the base checkpoint and from both Calibration Fine-Tuning variants, for three models, on held-out distribution families.}
\label{fig:main-ood-qualitative}
\end{figure*}

Figure~\ref{fig:main-structured-sampling} and Appendix Tables~\ref{tab:main-ood}--\ref{tab:main-unseen} show that Calibration Fine-Tuning strongly improves structured distribution sampling on both held-out families and unseen parameter settings. Family-median normalized \(W_1\) falls for nearly every model, by a median factor of two to four, and trie-based logit KL (Appendix Figure~\ref{fig:appendix-structured-kl}) falls by roughly an order of magnitude. Since \(W_1\) is computed on parsed values, these gains are not formatting effects: models already near-perfect in validity still show large reductions in \(W_1\). The two training variants perform similarly, with small differences that vary by split and model; Appendix~\ref{app:matched-budget} compares them at a matched optimizer budget, where they are statistically indistinguishable. Figure~\ref{fig:main-ood-qualitative} gives representative held-out examples, and the appendix reports the full per-distribution breakdowns. The gains do not depend on the training template: under four rewordings of the prompt that keep the distribution and its parameters fixed, the worst rewording of either fine-tuned model still beats the base model's best (Appendix~\ref{app:prompt-robustness}).

We compare against two inference-time baselines: String Seed of Thought (SSOT) prompting \citep{misaki2025stringseed} (Appendix~\ref{sec:appendix-ssot}) and rescaling the base model's sampling temperature (Appendix~\ref{app:temperature}). SSOT can improve over the base checkpoint when the model reliably follows the seed-and-reasoning protocol, but it is brittle, model-dependent, and more expensive per sample than direct sampling, and thus remains weaker than Calibration Fine-Tuning across our evaluated checkpoints. No sampling temperature gives the base model both the validity and the distributional fit of either fine-tuned variant. Even if we pick the best temperature separately for each prompt after seeing the results, the base model beats both fine-tuned variants only on Qwen3-0.6B. The training-time baseline of \citet{zhang2024forcing} also improves on the base model for three of the four models, but its logit KL stays 2.7 to 5.7 times above ours despite seven times as many optimizer steps. On their own benchmark, when neither recipe has trained on the task being evaluated, theirs scores higher on Qwen3-1.7B and Gemma-3-12B-it, and ours on Qwen3-14B and GPT-OSS-20B (Appendix~\ref{app:zhang}).

We treat sample-level \(W_1\) as the primary structured-sampling metric because it is computed on parsed values and does not depend on how a model formats its output. Logit KL, computed against the five-decimal outputs used in training, is a complementary check. \(W_1\) has one known weakness. Many target families have unbounded support (e.g., Poisson, exponential, lognormal), so an extremely large literal is still a valid, in-support sample, and because \(W_1\) is unbounded a single such sample can dominate a prompt's score. For example, the GPT-OSS-20B base model emits one sample of order \(10^{31}\) on Poisson(\(\lambda=1\)), which yields a normalized \(W_1\) of \(1.4 \times 10^{28}\) on that prompt. These blowups occur mostly for base checkpoints, but also for the smallest fine-tuned model (Qwen3-0.6B, soft target) on several unseen-parameter prompts. The median across families absorbs most of them, but not all. For the Llama-3.2 base models, and for Qwen3-0.6B soft on unseen parameters, the aggregate \(W_1\) reflects a few extreme samples rather than the bulk of the distribution, and for the GPT-OSS-20B base model such samples roughly double it. We report all \(W_1\) values unmodified.

\paragraph{Open-ended random generation.}

\begin{table*}[htbp]
\centering
\scriptsize
\resizebox{\textwidth}{!}{%
\begin{tabular}{lcccccccccccc}
\toprule
 & \multicolumn{3}{c}{Top-90\% First-Token Support $\uparrow$} & \multicolumn{3}{c}{Unique Output Rate $\uparrow$} & \multicolumn{3}{c}{MCQ TV $\downarrow$} & \multicolumn{3}{c}{NoveltyBench Utility $\uparrow$} \\
\cmidrule(lr){2-4}\cmidrule(lr){5-7}\cmidrule(lr){8-10}\cmidrule(lr){11-13}
Model & Base & Soft & Hard & Base & Soft & Hard & Base & Soft & Hard & Base & Soft & Hard \\
\midrule
Qwen3-0.6B & 45 & \textbf{711.2} & 501.6 & \textbf{56.9\%} & 45.3\% & 45.9\% & \textbf{0.374} & -- & 0.398 & \textbf{2.639} & 1.654 & 2.04 \\
Qwen3-1.7B & 8.8 & \textbf{864.5} & 265 & 22.5\% & \textbf{39.1\%} & 30.3\% & 0.313 & \textbf{0.249} & 0.288 & 2.804 & \textbf{3.143} & 3.094 \\
Qwen3-4B & 4.4 & \textbf{299.6} & 127.8 & 11.9\% & \textbf{27.7\%} & 21.4\% & 0.192 & \textbf{0.116} & 0.156 & 2.915 & \textbf{3.559} & 3.322 \\
Qwen3-8B & 3.3 & \textbf{466.9} & 219.6 & 9.5\% & \textbf{37.9\%} & 27.8\% & 0.24 & 0.24 & \textbf{0.221} & 3.378 & \textbf{3.843} & 3.734 \\
Qwen3-14B & 3.5 & \textbf{536.2} & 194 & 10.9\% & \textbf{49.1\%} & 35.9\% & 0.284 & \textbf{0.184} & 0.19 & 3.392 & \textbf{4.096} & 3.849 \\
Gemma-3-1B-it & 5.5 & \textbf{52.6} & 36.6 & 13.3\% & 16.8\% & \textbf{17.3\%} & 0.209 & 0.205 & \textbf{0.203} & 2.303 & \textbf{2.555} & 2.527 \\
Gemma-3-4B-it & 2.6 & \textbf{115} & 22.9 & 6.9\% & \textbf{49.2\%} & 25.0\% & 0.24 & \textbf{0.178} & 0.256 & 2.332 & \textbf{3.034} & 2.776 \\
Gemma-3-12B-it & 3.3 & \textbf{44.8} & 23.6 & 7.4\% & \textbf{31.8\%} & 21.4\% & 0.335 & \textbf{0.289} & 0.313 & 2.173 & \textbf{2.863} & 2.678 \\
Gemma-3-27B-it & 2.3 & \textbf{63} & 39.4 & 5.8\% & \textbf{40.1\%} & 30.0\% & 0.284 & 0.315 & \textbf{0.238} & 2.27 & \textbf{2.858} & 2.623 \\
Llama-3.2-1B-it & 84.6 & \textbf{86.2} & 83.9 & \textbf{52.5\%} & 17.7\% & 18.4\% & \textbf{0.159} & -- & 0.199 & \textbf{2.56} & 2.388 & 2.466 \\
Llama-3.2-3B-it & 28.9 & \textbf{85.6} & 78.9 & \textbf{36.4\%} & 18.8\% & 17.9\% & \textbf{0.066} & 0.187 & 0.189 & \textbf{3.385} & 3.126 & 3.183 \\
GPT-OSS-20B & 28 & \textbf{289} & 212.4 & 36.2\% & 73.0\% & \textbf{77.1\%} & 0.166 & \textbf{0.061} & 0.072 & \textbf{3.786} & 2.179 & 2.615 \\
\bottomrule
\end{tabular}
}
\caption{\textbf{Generative-distribution summary.} Top-90\% first-token support and Unique Output Rate measure open-ended random-generation diversity; MCQ TV measures answer-position balance over parseable generations; NoveltyBench Utility is the benchmark's patience-discounted reward metric. MCQ TV is marked -- when no generation is parseable. Base is the original checkpoint; Soft and Hard are the two Calibration Fine-Tuning variants. Bold denotes the best value for each available metric.}
\label{tab:generative-distribution-summary}
\end{table*}

Calibration Fine-Tuning also transfers beyond explicit distribution-name prompts. Table~\ref{tab:generative-distribution-summary} summarizes the aggregate open-ended random-generation results. The broadest and most consistent effect is on first-token support: Calibration Fine-Tuning increases the top-$90\%$ first-token support size for all 12 models under soft-target and 11 of 12 under hard-target fine-tuning, often by one to two orders of magnitude.

Broader first-token support also translates into more diverse repeated generations: the unique-output rate rises significantly for 8 of the 12 models under both variants, typically by a factor of two to seven, and falls only for Qwen3-0.6B and the two Llama checkpoints. Appendix Figures~\ref{fig:appendix-open-multi-threshold} and~\ref{fig:appendix-open-qualitative} provide the fuller per-prompt breakdowns and qualitative repeated-sampling examples.

\paragraph{NoveltyBench.}
NoveltyBench tests whether broader first-token support translates into semantically useful open-ended generation under the benchmark's original partition-and-reward pipeline. Table~\ref{tab:generative-distribution-summary} shows that both variants improve overall utility over the base checkpoint for 8 of the 12 models, with the largest gains for medium and large Qwen and Gemma checkpoints. The gains are not uniform: GPT-OSS-20B is the clearest counterexample, where both fine-tuned variants increase distinctness but reduce utility relative to the base checkpoint, and Qwen3-0.6B shows a similar failure mode. Overall, NoveltyBench strengthens the transfer case, while also showing that higher semantic spread is only valuable when it remains aligned with utility. Appendix~\ref{sec:appendix-noveltybench} provides the split-level results and representative qualitative cases.

\paragraph{MCQ answer-position balance.}
MCQ transfer is mixed. Over parseable generations, answer-position balance improves for most models, significantly for four under soft-target fine-tuning, but soft-target fine-tuning also lowers the fraction of generations that parse as valid MCQs on seven models, sharply for GPT-OSS-20B and the smallest checkpoints. For the full breakdown, see Appendix~\ref{sec:appendix-mcq}.

\paragraph{Capability retention.}
Table~\ref{tab:retention-transfer-summary} reports TinyBenchmarks gp-IRT together with PALOMA perplexity, and Appendix Table~\ref{tab:retention-summary} gives the full TinyBenchmarks breakdown. The retention picture is mixed and less uniformly positive than the structured and transfer results above. On aggregate gp-IRT, the base checkpoint remains best for most models, especially among the smaller Qwen and Gemma checkpoints, both Llama models, and GPT-OSS-20B. A few medium and large models do improve under Calibration Fine-Tuning, notably Qwen3-8B, Qwen3-14B, Gemma-3-12B-it, and Gemma-3-27B-it. At the task level, MMLU, HellaSwag, and WinoGrande improve modestly on average, while TruthfulQA is nearly flat. The clearest systematic cost is GSM8K, where strict and flexible scoring regress substantially. Appendix~\ref{sec:appendix-capability-retention} provides a deeper task-level breakdown.

\begin{table}[t]
\centering
\begin{minipage}[c]{0.56\textwidth}
\centering
\scriptsize
\setlength{\tabcolsep}{3.5pt}
\resizebox{\linewidth}{!}{%
\begin{tabular}{lcccccc}
\toprule
 & \multicolumn{3}{c}{TinyBenchmarks gp-IRT $\uparrow$} & \multicolumn{3}{c}{PALOMA Perplexity $\downarrow$} \\
\cmidrule(lr){2-4}\cmidrule(lr){5-7}
Model & Base & Soft & Hard & Base & Soft & Hard \\
\midrule
Qwen3-0.6B & \textbf{0.36} & 0.302 & 0.31 & 19.73 & \textbf{19.27} & 19.64 \\
Qwen3-1.7B & \textbf{0.451} & 0.394 & 0.45 & 15.5 & \textbf{14.6} & 14.78 \\
Qwen3-4B & \textbf{0.619} & 0.584 & 0.616 & 13.4 & \textbf{12.52} & 12.71 \\
Qwen3-8B & 0.647 & \textbf{0.676} & 0.637 & 10.77 & \textbf{10.33} & 10.46 \\
Qwen3-14B & 0.682 & 0.719 & \textbf{0.722} & 9.77 & \textbf{9.54} & 9.58 \\
Gemma-3-1B-it & \textbf{0.389} & 0.33 & 0.349 & 29.48 & \textbf{28.92} & 29.54 \\
Gemma-3-4B-it & \textbf{0.584} & 0.537 & 0.572 & 24.68 & \textbf{21.51} & 22.16 \\
Gemma-3-12B-it & 0.684 & \textbf{0.709} & 0.698 & 56.56 & \textbf{36.34} & 47.99 \\
Gemma-3-27B-it & 0.716 & 0.677 & \textbf{0.743} & 25.87 & \textbf{23.53} & 24.26 \\
Llama-3.2-1B-it & \textbf{0.358} & 0.318 & 0.316 & \textbf{15.15} & 15.32 & 15.45 \\
Llama-3.2-3B-it & \textbf{0.48} & 0.463 & 0.461 & 12.04 & 12.04 & \textbf{12} \\
GPT-OSS-20B & \textbf{0.463} & 0.332 & 0.389 & 103.73 & 99.95 & \textbf{70.17} \\
\bottomrule
\end{tabular}
}
\end{minipage}\hfill
\begin{minipage}[c]{0.41\textwidth}
\captionof{table}{\textbf{Retention summary.} TinyBenchmarks gp-IRT measures downstream retention; PALOMA reports perplexity on text from diverse domains. Base is the original checkpoint; Soft and Hard are the two Calibration Fine-Tuning variants. Bold denotes the best value for each metric.}
\label{tab:retention-transfer-summary}
\end{minipage}
\end{table}

\paragraph{PALOMA perplexity.}
PALOMA gives a more favorable language-modeling view. Under the selected training budgets, Calibration Fine-Tuning preserves or improves aggregate perplexity for 11 of the 12 models under soft-target and 10 of 12 under hard-target fine-tuning, with degradations only on Llama-3.2-1B-it and, marginally, Gemma-3-1B-it under hard-target. This argues against the simple interpretation that Calibration Fine-Tuning merely flattens token probabilities across the vocabulary. For a deeper analysis of PALOMA, see Appendix~\ref{sec:appendix-paloma}.

\paragraph{Hyperparameter ablations.}
Appendix~\ref{sec:appendix-ablations} reports the ablations used to choose the final discretization and training budgets. The selected settings reflect the best overall tradeoff in held-out and unseen-parameter performance rather than monotonic preferences.

\section{Discussion and Limitations}
\label{sec:analysis}

\paragraph{What does Calibration Fine-Tuning seem to teach?}

Our main conclusion is that probabilistic calibration is trainable. Across 12 models, Calibration Fine-Tuning sharply improves structured distribution sampling on both held-out families and unseen parameter settings, reducing sample-level error and logit-level miscalibration. Both variants reach this level, showing that simple supervised training, even on sampled completions alone, can teach substantially better stochastic fidelity than the base checkpoint.

\paragraph{How does it transfer?}

The structured gains do transfer to natural language settings, but selectively. The clearest effect is that Calibration Fine-Tuning broadens first-token support and often improves semantically diverse generation, as reflected by open-ended random generation and NoveltyBench. MCQ answer-position balance improves among parseable generations, but soft-target fine-tuning often lowers parse rates, so transfer to targets only implicit in a task format remains partial. This is also where the method is useful: when a distribution is written down and exact draws are needed, a pseudo-random generator is cheaper and exact, while fine-tuning helps when the target is only implicit in the task and the valid outputs would otherwise have to be listed by hand.

\paragraph{What remains brittle?}
Capability retention is mixed, with the clearest systematic cost on GSM8K and aggregate gp-IRT often still favoring the base checkpoint, especially for smaller models. One plausible mechanism is that both Calibration Fine-Tuning variants train short direct completions without reasoning traces; after post-training regimes that often reward long reasoning, this may shift models toward shorter generations and hurt reasoning-heavy tasks, an effect we observe most strongly on the smaller checkpoints. Because hard-target supervision is sampled-path rather than dense prefix-level supervision, our final hard-target configuration uses more optimizer steps; retention differences between the two variants should therefore be read as configuration-level tradeoffs, not as isolated effects of the loss objective. A second practical limitation is that our structured targets are finite canonical approximations of the underlying SciPy laws: tail mass is assigned to edge bins to preserve total probability, which is tractable and consistent across training and evaluation but can distort heavy-tailed targets. The remaining challenge is to preserve general capabilities, arithmetic reasoning in particular, while keeping the calibration and transfer gains.

\section{Conclusion}
\label{sec:conclusion}

We studied Calibration Fine-Tuning as a simple way to improve stochastic generation behavior in language models. Across 12 models, fine-tuning on cheap synthetic mathematical distributions with either objective makes models behave more like controlled probabilistic samplers, and the gains transfer to unseen distributions and to natural language. Future work should study the mechanisms learned by each objective and develop retention-aware variants of Calibration Fine-Tuning.

\section*{Acknowledgements}

Sarath Chandar is supported by the Canada CIFAR AI Chairs program, the Canada Research Chair in Lifelong Machine Learning, and the NSERC Discovery Grant. Experiments were conducted using computational resources provided by Mila Quebec AI Institute.

\subsection*{AI use statement}
In this work, we used LLM-based coding assistants to help write and refactor experiment, evaluation, and analysis code, and to help edit and proofread the manuscript text. All AI-assisted code was reviewed and tested by the authors, and every number reported in the paper was produced by the authors' own pipeline and checked against an independent recomputation. We did not use generative AI tools to generate the research questions, to design the method, or to conduct the literature survey. We have reviewed all AI-assisted work and take responsibility for the final content of this work, including text, claims, and artifacts produced with the aid of generative AI.

\subsection*{Ethics statement}
This work fine-tunes publicly released open-weight language models on fully synthetic prompts that ask for samples from named mathematical distributions. No human subjects, personal data, or scraped user content are involved. The capability studied, following a user-specified sampling distribution, is a form of controllability, and we see no foreseeable misuse specific to it beyond that of the underlying models. We report the downstream capability costs of fine-tuning, which vary by model, so that practitioners can weigh them before deployment.

\subsection*{Reproducibility statement}
Both fine-tuning objectives are given as pseudocode in Algorithms~\ref{alg:calibrate-sft} and~\ref{alg:hard-label-sft}. Appendix~\ref{app:implementation} lists the training and data construction details, the hyperparameters, and the full reference table of benchmark distributions; Appendices~\ref{app:prompts-structured} through~\ref{app:mcq-prompt} give the exact prompt templates for every evaluation axis. All 12 models are publicly available checkpoints named in Section~\ref{sec:setup}. Code to build the training data, fine-tune, evaluate, and regenerate every table and figure is available at \url{https://github.com/chandar-lab/calibration-finetuning}.

\bibliography{references}
\bibliographystyle{iclr2027_conference}

\clearpage
\appendix

\begin{center}
\Large\textbf{Appendix}
\end{center}

\section{Calibration Fine-Tuning Algorithms}
\suppressfloats[t]  %

Algorithm~\ref{alg:calibrate-sft} formalizes the soft-target Calibration Fine-Tuning pipeline described in Section~\ref{sec:method}. The first phase constructs a prefix trie from the discretized target distribution and derives per-prefix next-token targets. The second phase samples a training path through the trie and computes the prompt-conditioned calibration loss. Algorithm~\ref{alg:hard-label-sft} describes the hard-target variant, which reuses the same prompt distributions and canonical output space, but replaces trie-derived soft targets with repeated sampled completions and standard completion cross-entropy.

\begin{algorithm}[H]
\caption{Soft-target Calibration Fine-Tuning}
\label{alg:calibrate-sft}
\begin{algorithmic}[1]
\REQUIRE Prompt text $u$, target law $P$, tokenizer $\mathcal{T}$, model $\pi_\theta$, temperature $\tau$
\STATE \textbf{// Build output space and prefix trie}
\STATE Discretize $P$ into canonical outputs $Y = \{y_1, \dots, y_M\}$ with induced masses $P_Y(y_i)$
\FOR{each $y_i \in Y$}
    \STATE Tokenize: $\mathbf{s}_i \leftarrow \mathcal{T}(y_i) \circ \langle\textsc{eos}\rangle$
    \FOR{each proper prefix $p$ of $\mathbf{s}_i$}
        \STATE $\texttt{prefix\_mass}[p] \mathrel{+}= P_Y(y_i)$
        \STATE $\texttt{next\_mass}[p][\mathbf{s}_i[|p|+1]] \mathrel{+}= P_Y(y_i)$
    \ENDFOR
\ENDFOR
\FOR{each prefix $p$ in trie}
    \STATE $q(v \mid p) \leftarrow \texttt{next\_mass}[p][v] \;/\; \texttt{prefix\_mass}[p]$ for each child $v$
\ENDFOR
\STATE \textbf{// Training step}
\STATE $\mathbf{x} \leftarrow \mathcal{T}(u)$
\STATE Sample $y \sim P_Y$; let $\mathbf{t} = (t_1, \dots, t_L, \langle\textsc{eos}\rangle) = \mathcal{T}(y) \circ \langle\textsc{eos}\rangle$
\FOR{$k = 0$ \TO $L$}
    \STATE $p \leftarrow (t_1, \dots, t_k)$
    \STATE Compute $\pi_\theta(\cdot \mid \mathbf{x}, p) \leftarrow \mathrm{softmax}(f_\theta(\mathbf{x}, p) / \tau)$
    \STATE $\ell_k \leftarrow \mathrm{KL}\!\left(q(\cdot \mid p)\,\|\,\pi_\theta(\cdot \mid \mathbf{x}, p)\right)$
\ENDFOR
\STATE $\mathcal{L}_{\mathrm{soft}} \leftarrow \frac{1}{L+1}\sum_{k=0}^{L} \ell_k$
\STATE Update LoRA parameters via $\nabla \mathcal{L}_{\mathrm{soft}}$
\end{algorithmic}
\end{algorithm}

\begin{algorithm}[H]
\caption{Hard-target Calibration Fine-Tuning}
\label{alg:hard-label-sft}
\begin{algorithmic}[1]
\REQUIRE Training prompts $\mathcal{P}_{\mathrm{train}}$, target laws $\{P_p\}_{p \in \mathcal{P}_{\mathrm{train}}}$, tokenizer $\mathcal{T}$, model $\pi_\theta$, samples per prompt per epoch $R$, epochs $E$
\FOR{$e = 1$ \TO $E$}
    \STATE Form epoch prompt multiset $\tilde{\mathcal{P}}_e \leftarrow \bigcup_{p \in \mathcal{P}_{\mathrm{train}}} \{p\}^{R}$
    \STATE Shuffle $\tilde{\mathcal{P}}_e$ with family-balanced ordering
    \FOR{each prompt $p \in \tilde{\mathcal{P}}_e$}
        \STATE Discretize $P_p$ into canonical outputs $Y_p$ with induced distribution $P_{Y,p}$
        \STATE Sample canonical output $y \sim P_{Y,p}$
        \STATE $\mathbf{x} \leftarrow \mathcal{T}(\texttt{prompt\_text}(p))$
        \STATE $\mathbf{t} \leftarrow \mathcal{T}(y) \circ \langle\textsc{eos}\rangle$
        \STATE Construct training sequence $\mathbf{z} \leftarrow \mathbf{x} \circ \mathbf{t}$
        \STATE Set labels $\mathbf{m} \leftarrow (-100)^{|\mathbf{x}|} \circ \mathbf{t}$ \COMMENT{mask prompt tokens}
    \ENDFOR
    \FOR{each minibatch $\{(\mathbf{z}^{(b)}, \mathbf{m}^{(b)})\}_{b=1}^{B}$}
        \STATE Compute next-token logits on each sequence $\mathbf{z}^{(b)}$
        \STATE $\mathcal{L}_{\mathrm{hard}} \leftarrow \mathrm{CE}(\text{shifted logits}, \text{shifted labels})$ over non-masked positions
        \STATE Update LoRA parameters via $\nabla \mathcal{L}_{\mathrm{hard}}$
    \ENDFOR
\ENDFOR
\end{algorithmic}
\end{algorithm}

\clearpage
\section{Implementation and Reproducibility Details}
\label{app:implementation}
\suppressfloats[t]  %

\subsection{Training and Data Construction}

Tables~\ref{tab:shared-hyperparams}--\ref{tab:method-hyperparams} consolidate the training and data-construction settings described in Section~\ref{sec:implementation-details}. Both Calibration Fine-Tuning variants share the same model loading procedure, frozen-base LoRA setup, optimizer, FSDP/bf16 distributed training setup, and family-balanced batching; method-specific settings differ only where required by the supervision format.

For data construction, both methods use the same 1988 training prompts. For continuous distributions, we first intersect the quantile interval $[Q(0.001), Q(0.999)]$ with any finite support bounds, then form the decimal grid at precision $d$. If this grid contains more than the configured maximum number of bins, we retain evenly spaced grid points including the endpoints. Bin boundaries are the midpoints between retained centers, and $P_Y$ is computed from CDF differences across these bins; probability mass below the lower endpoint and above the upper endpoint is assigned to the first and last bins. For integer-valued distributions, we enumerate the finite integer support when available; otherwise, we use quantile-derived integer bounds, cap the resulting support if needed, and assign any upper-tail mass to the last retained value. The final canonical precision is $d=5$ decimals for both variants, with method-specific caps of 1001 bins for soft-target fine-tuning and 16384 bins for hard-target fine-tuning.

Thus, the structured-sampling objectives are defined with respect to a finite canonical approximation of the underlying SciPy law; edge-bin tail assignment is a practical choice that preserves total probability mass but can distort tail behavior for heavy-tailed targets.

\begin{table}[htbp]
\centering
\small
\begin{tabular}{@{}ll@{}}
\toprule
Parameter & Value \\
\midrule
Fine-tuning method & LoRA \\
Training prompts & 1988 \\
Canonical decimals & 5 \\
Continuous quantile range & $[0.001, 0.999]$ \\
Tail mass outside range & Clipped to edge bins \\
LoRA rank / alpha / dropout & 16 / 32 / 0.05 \\
LoRA target modules & \texttt{q\_proj}, \texttt{k\_proj}, \texttt{v\_proj}, \texttt{o\_proj} \\
Optimizer & AdamW \\
Learning rate & $2 \times 10^{-4}$ \\
Weight decay & 0.01 \\
LR schedule & Cosine with 3\% linear warmup \\
Per-device batch size & 8 \\
Gradient accumulation & 1 \\
Training GPUs & 4 A100 GPUs with FSDP \\
Max sequence length & 256 tokens \\
Model/load precision & bf16 \\
Batching & Family-balanced \\
\bottomrule
\end{tabular}
\caption{\textbf{Shared training and data-construction settings for both Calibration Fine-Tuning variants.} These values are held fixed across model families with no per-model tuning.}
\label{tab:shared-hyperparams}
\end{table}

\begin{table}[htbp]
\centering
\small
\begin{tabular}{@{}lll@{}}
\toprule
Parameter & Soft & Hard \\
\midrule
Objective & Trie-induced KL & Completion CE \\
Maximum output-space bins & 1001 & 16384 \\
Epochs & 3 & 2 \\
Sampled completions per prompt per epoch & 1 & 16 \\
Optimizer steps per model & 189 & 1988 \\
Loss temperature & $\tau=1.0$ & N/A \\
\bottomrule
\end{tabular}
\caption{\textbf{Method-specific training hyperparameters.} Hard-target fine-tuning uses more optimizer steps because each completion provides sparse sampled-path supervision rather than dense prefix-level targets.}
\label{tab:method-hyperparams}
\end{table}

All final fine-tuning runs used one internal GPU-cluster node with 4 NVIDIA A100-SXM4-80GB GPUs, 24 CPU cores, and approximately 1TB system memory, using FSDP/bf16 training. W\&B-recorded optimizer-loop wall-clock times ranged from 0.7 to 14.9 minutes for soft-target fine-tuning (median 1.8 minutes; 2.6 A100 GPU-hours total across 12 models) and from 6.5 to 154.9 minutes for hard-target fine-tuning (median 18.3 minutes; 26.7 A100 GPU-hours total). Final evaluation jobs used the same cluster; summed per-run wall-clock estimates were 4.5h for structured sampling, 3.1h for open-ended random generation, 17.6h for MCQ generation, 4.8h for TinyBenchmarks, 11.5h for PALOMA, and 194.0h for NoveltyBench, with NoveltyBench dominated by generation, semantic partitioning, and reward-model scoring.

\paragraph{Assets.}
We use public model checkpoints and evaluation assets: Qwen3, Gemma-3-it, Llama-3.2-Instruct, GPT-OSS, NoveltyBench, TinyBenchmarks, PALOMA, the MCQ protocol of \citet{zhao2026dice}, and the evaluator checkpoints used by NoveltyBench. All are cited in the main text or appendix, and the released artifact lists the exact checkpoints, public sources, and upstream access or license requirements. Our training prompts and structured-sampling benchmark are synthetically generated and documented in Table~\ref{tab:benchmark-distributions} and Appendix~\ref{app:prompts-structured}.

\subsection{Benchmark Distribution Reference}

Table~\ref{tab:benchmark-distributions} lists all 36 distribution families used in the Calibration Fine-Tuning benchmark, organized by tier, along with the training parameter regions and paper-selected evaluation configurations. Shaded rows denote the 12 held-out families, which are excluded from training.

\begin{table*}[htbp]
\centering
\scriptsize
\setlength{\tabcolsep}{4pt}
\begin{tabular}{@{}p{0.06\linewidth}>{\raggedright\arraybackslash}p{0.15\linewidth}>{\raggedright\arraybackslash}p{0.14\linewidth}>{\raggedright\arraybackslash}p{0.27\linewidth}@{\hspace{0.02\linewidth}}>{\raggedright\arraybackslash}p{0.24\linewidth}@{}}
\toprule
Tier & Distribution & SciPy Name & Train & Test \\
\midrule
I & Uniform & \texttt{uniform} & $a \in [-5, 2]$, \allowbreak $w \in [1, 5]$ & $a=3.5$, \allowbreak $w=7$ \\
I & Gaussian & \texttt{norm} & $\mu \in [-2, 2]$, \allowbreak $\sigma \in [0.5, 2]$ & $\mu=3.5$, \allowbreak $\sigma=3$ \\
\rowcolor{oodgray}
I & Bernoulli & \texttt{bernoulli} & \multicolumn{2}{>{\raggedright\arraybackslash}p{0.53\linewidth}@{}}{$p=0.1$; \allowbreak $p=0.5$; \allowbreak $p=0.9$} \\
II & Beta & \texttt{beta} & $\alpha \in [0.5, 5]$, \allowbreak $\beta \in [0.5, 5]$ & $\alpha=7$, \allowbreak $\beta=7$ \\
II & Binomial & \texttt{binom} & $n \in \{5, \allowbreak 10, \allowbreak 15, \allowbreak 20\}$, \allowbreak $p \in \{0.2, \allowbreak 0.25, \allowbreak 0.3, \allowbreak 0.35, \allowbreak 0.4, \allowbreak 0.5, \allowbreak 0.6, \allowbreak 0.65, \allowbreak 0.7, \allowbreak 0.75, \allowbreak 0.8\}$ & $n=25$, \allowbreak $p=0.5$ \\
II & Exponential & \texttt{expon} & $\lambda \in [0.5, 5]$ & $\lambda=7$ \\
II & Geometric & \texttt{geom} & $p \in \{0.2, \allowbreak 0.225, \allowbreak 0.25, \allowbreak 0.3, \allowbreak 0.35, \allowbreak 0.4, \allowbreak 0.5, \allowbreak 0.6, \allowbreak 0.65, \allowbreak 0.7, \allowbreak 0.75, \allowbreak 0.775, \allowbreak 0.8\}$ & $p=0.125$ \\
II & Negative Binomial & \texttt{nbinom} & $r \in \{3, \allowbreak 5, \allowbreak 8, \allowbreak 12\}$, \allowbreak $p \in \{0.2, \allowbreak 0.25, \allowbreak 0.3, \allowbreak 0.35, \allowbreak 0.4, \allowbreak 0.5, \allowbreak 0.6, \allowbreak 0.65, \allowbreak 0.7, \allowbreak 0.75, \allowbreak 0.8\}$ & $r=15$, \allowbreak $p=0.15$ \\
II & LogNormal & \texttt{lognorm} & $\mu \in [-1, 1.5]$, \allowbreak $\sigma \in [0.25, 1.25]$ & $\mu=2.5$, \allowbreak $\sigma=2$ \\
II & Triangular & \texttt{triang} & $a \in [-3, 1]$, \allowbreak $w \in [1, 5]$, \allowbreak $f_{\mathrm{mode}} \in \{0.1, \allowbreak 0.3, \allowbreak 0.5, \allowbreak 0.7, \allowbreak 0.9\}$ & $a=2.5$, \allowbreak $w=7$, \allowbreak $f_{\mathrm{mode}}=0.5$ \\
II & Rayleigh & \texttt{rayleigh} & $\sigma \in [0.5, 2]$ & $\sigma=3$ \\
\rowcolor{oodgray}
II & Poisson & \texttt{poisson} & \multicolumn{2}{>{\raggedright\arraybackslash}p{0.53\linewidth}@{}}{$\lambda=1$; \allowbreak $\lambda=4$; \allowbreak $\lambda=12$} \\
\rowcolor{oodgray}
II & Maxwell & \texttt{maxwell} & \multicolumn{2}{>{\raggedright\arraybackslash}p{0.53\linewidth}@{}}{$\sigma=0.75$; \allowbreak $\sigma=1.5$; \allowbreak $\sigma=2.5$} \\
III & Cauchy & \texttt{cauchy} & $x_0 \in [-2, 2]$, \allowbreak $\gamma \in [0.5, 2]$ & $x_0=3.5$, \allowbreak $\gamma=3$ \\
III & Student's $t$ & \texttt{t} & $\nu \in \{2.5, \allowbreak 2.75, \allowbreak 3, \allowbreak 3.5, \allowbreak 4, \allowbreak 4.5, \allowbreak 5, \allowbreak 6, \allowbreak 7, \allowbreak 8, \allowbreak 9, \allowbreak 10\}$ & $\nu=16$ \\
\rowcolor{oodgray}
III & Chi & \texttt{chi} & \multicolumn{2}{>{\raggedright\arraybackslash}p{0.53\linewidth}@{}}{$\nu=2$; \allowbreak $\nu=5$; \allowbreak $\nu=10$} \\
III & Chi-Square & \texttt{chi2} & $\nu \in \{2, \allowbreak 3, \allowbreak 4, \allowbreak 5, \allowbreak 6, \allowbreak 7, \allowbreak 8, \allowbreak 10, \allowbreak 12, \allowbreak 14, \allowbreak 16, \allowbreak 18, \allowbreak 20\}$ & $\nu=32$ \\
III & F-Distribution & \texttt{f} & $d_1 \in \{3, \allowbreak 5, \allowbreak 7, \allowbreak 10\}$, \allowbreak $d_2 \in \{5, \allowbreak 10, \allowbreak 15, \allowbreak 20\}$ & $d_1=12$, \allowbreak $d_2=24$ \\
III & Gamma & \texttt{gamma} & $\alpha \in [1, 5]$, \allowbreak $\beta \in [1, 5]$ & $\alpha=7$, \allowbreak $\beta=7$ \\
\rowcolor{oodgray}
III & Weibull & \texttt{weibull\_min} & \multicolumn{2}{>{\raggedright\arraybackslash}p{0.53\linewidth}@{}}{$k=0.5$, $\lambda=0.5$; \allowbreak $k=1.5$, $\lambda=1.5$; \allowbreak $k=3$, $\lambda=3$} \\
\rowcolor{oodgray}
II & TruncNorm & \texttt{truncnorm} & \multicolumn{2}{>{\raggedright\arraybackslash}p{0.53\linewidth}@{}}{$\mu=0$, $\sigma=1$, $a=-1$, $b=1$; \allowbreak $\mu=0$, $\sigma=1$, $a=-2$, $b=2$; \allowbreak $\mu=1$, $\sigma=1.5$, $a=-1$, $b=2$} \\
III & Laplace & \texttt{laplace} & $\mu \in [-2, 2]$, \allowbreak $b \in [0.5, 2]$ & $\mu=3.5$, \allowbreak $b=3$ \\
III & Logistic & \texttt{logistic} & $\mu \in [-2, 2]$, \allowbreak $s \in [0.5, 2]$ & $\mu=3.5$, \allowbreak $s=3$ \\
III & Pareto & \texttt{pareto} & $\alpha \in \{2, \allowbreak 2.25, \allowbreak 2.5, \allowbreak 2.75, \allowbreak 3, \allowbreak 3.5, \allowbreak 4, \allowbreak 4.5, \allowbreak 5\}$, \allowbreak $x_m \in [0.5, 2]$ & $\alpha=6.5$, \allowbreak $x_m=3.5$ \\
III & Hypergeometric & \texttt{hypergeom} & $M \in \{30, \allowbreak 50, \allowbreak 80\}$, \allowbreak $N \in \{5, \allowbreak 10, \allowbreak 15\}$, \allowbreak $K/M \in \{0.2, \allowbreak 0.35, \allowbreak 0.5, \allowbreak 0.65, \allowbreak 0.8\}$ & $M=100$, \allowbreak $N=20$, \allowbreak $K/M=0.5$ \\
III & Gumbel & \texttt{gumbel\_r} & $\mu \in [-2, 2]$, \allowbreak $\beta \in [0.5, 2]$ & $\mu=3.5$, \allowbreak $\beta=3$ \\
III & Skellam & \texttt{skellam} & $\mu_1 \in [1, 8]$, \allowbreak $\mu_2 \in [1, 8]$ & $\mu_1=10.5$, \allowbreak $\mu_2=10.5$ \\
III & Beta-Binomial & \texttt{betabinom} & $n \in \{10, \allowbreak 20, \allowbreak 30\}$, \allowbreak $\alpha \in [0.5, 5]$, \allowbreak $\beta \in [0.5, 5]$ & $n=40$, \allowbreak $\alpha=6.5$, \allowbreak $\beta=6.5$ \\
III & Lomax & \texttt{lomax} & $\alpha \in [1.5, 4]$, \allowbreak $\lambda \in [0.5, 3]$ & $\alpha=6$, \allowbreak $\lambda=4.5$ \\
III & Inverse Gaussian & \texttt{invgauss} & $\mu \in [0.5, 3]$, \allowbreak $\lambda \in [0.5, 5]$ & $\mu=5$, \allowbreak $\lambda=7$ \\
\rowcolor{oodgray}
III & Bimodal Gaussian Mixture & mixture of \texttt{norm} & \multicolumn{2}{>{\raggedright\arraybackslash}p{0.53\linewidth}@{}}{$0.5\,\mathcal{N}(-4,1)+0.5\,\mathcal{N}(4,1)$; \allowbreak $0.75\,\mathcal{N}(-3,1)+0.25\,\mathcal{N}(3,1)$; \allowbreak $0.5\,\mathcal{N}(0,0.3)+0.5\,\mathcal{N}(3,2)$} \\
\rowcolor{oodgray}
III & Trimodal Gaussian Mixture & mixture of \texttt{norm} & \multicolumn{2}{>{\raggedright\arraybackslash}p{0.53\linewidth}@{}}{$0.33\,\mathcal{N}(-4,1)+0.33\,\mathcal{N}(0,1)+0.34\,\mathcal{N}(4,1)$; \allowbreak $0.5\,\mathcal{N}(-3,0.8)+0.3\,\mathcal{N}(0,1.2)+0.2\,\mathcal{N}(5,1)$; \allowbreak $0.2\,\mathcal{N}(-2,1)+0.2\,\mathcal{N}(2,1)+0.6\,\mathcal{N}(6,0.8)$} \\
\rowcolor{oodgray}
III & Quadmodal Gaussian Mixture & mixture of \texttt{norm} & \multicolumn{2}{>{\raggedright\arraybackslash}p{0.53\linewidth}@{}}{$0.25\,\mathcal{N}(-6,1)+0.25\,\mathcal{N}(-2,1)+0.25\,\mathcal{N}(2,1)+0.25\,\mathcal{N}(6,1)$; \allowbreak $0.3\,\mathcal{N}(-5,1)+0.2\,\mathcal{N}(-1,0.8)+0.3\,\mathcal{N}(3,1.2)+0.2\,\mathcal{N}(7,1)$; \allowbreak $0.15\,\mathcal{N}(-4,1)+0.15\,\mathcal{N}(0,1)+0.15\,\mathcal{N}(4,1)+0.55\,\mathcal{N}(8,0.8)$} \\
\rowcolor{oodgray}
III & Bimodal Poisson Mixture & mixture of \texttt{poisson} & \multicolumn{2}{>{\raggedright\arraybackslash}p{0.53\linewidth}@{}}{$0.5\,\mathrm{Poi}(2)+0.5\,\mathrm{Poi}(15)$; \allowbreak $0.5\,\mathrm{Poi}(1)+0.5\,\mathrm{Poi}(8)$; \allowbreak $0.7\,\mathrm{Poi}(3)+0.3\,\mathrm{Poi}(20)$} \\
\rowcolor{oodgray}
III & Zipf & \texttt{zipfian} & \multicolumn{2}{>{\raggedright\arraybackslash}p{0.53\linewidth}@{}}{$a=1$, $n=100$; \allowbreak $a=1.5$, $n=100$; \allowbreak $a=2$, $n=100$} \\
\rowcolor{oodgray}
III & Categorical & pmf over $\{1, \dots, 6\}$ & \multicolumn{2}{>{\raggedright\arraybackslash}p{0.53\linewidth}@{}}{$p=(0.40, 0.25, 0.15, 0.10, 0.05, 0.05)$; \allowbreak $p=(0.35, 0.05, 0.05, 0.05, 0.15, 0.35)$; \allowbreak $p=(0.10, 0.10, 0.20, 0.20, 0.25, 0.15)$} \\
\bottomrule
\end{tabular}
\caption{\textbf{Calibration Fine-Tuning benchmark distributions.} The SciPy Name column lists the underlying \texttt{scipy.stats} distribution used to instantiate each target family \cite{2020SciPy-NMeth}; samples are then drawn from the corresponding frozen SciPy distribution. The mixture families and the categorical family have no single \texttt{scipy.stats} counterpart; for these the column names the component distribution or the object sampled. For seen families, the Train column summarizes the parameter region used for supervised calibration, and the Test column gives the paper-selected evaluation configuration used for unseen-parameter generalization. Shaded rows denote the 12 held-out families, which are excluded from training; having no training region, their entry spans both columns and lists the three evaluation settings. Mixtures are written as weighted sums of components, $w\,\mathcal{N}(\mu,\sigma)$ and $w\,\mathrm{Poi}(\lambda)$, with weights to two decimals; for Categorical, $p$ lists the probabilities of outcomes $1$ to $6$.}
\label{tab:benchmark-distributions}
\end{table*}
\normalsize

\clearpage
\section{Prompt Templates}
\suppressfloats[t]  %

\subsection{Structured Sampling}
\label{app:prompts-structured}

All structured sampling training and evaluation prompts are generated from the following template:

\begin{quote}
\small
\texttt{Generate exactly ONE random number from a [Distribution] distribution with parameters [params]. Output ONLY the number.}
\end{quote}

\noindent The \texttt{[Distribution]} field is the display name from Table~\ref{tab:benchmark-distributions}, and \texttt{[params]} is rendered as comma-separated key-value pairs in the canonical parameterization used by the corresponding distribution family. When applicable, the prompt is wrapped in the model's native chat template; for reasoning-capable models, we use the same reasoning-suppression settings as in training and evaluation. Generations are counted as valid numeric outputs if, after stripping any reasoning trace, they parse as a finite decimal or scientific-notation literal.

\subsection{String Seed of Thought Baseline}
\label{app:prompts-seed-string-thought}

For the String Seed of Thought baseline \citep{misaki2025stringseed}, we keep the original task prompt unchanged as the user message and add a benchmark-specific system or developer instruction. In the structured-sampling setting, this instruction asks the model to first generate a complex random string, use it as the seed for the requested probabilistic decision, place the seed and reasoning in intermediate tags, and emit the final sampled value in an \texttt{<answer>} tag. For GPT-OSS, the same idea is implemented through its native reasoning protocol: the seed and reasoning are kept in analysis, and the final channel is instructed to contain only the sampled answer. SSOT generations use \texttt{max\_new\_tokens}$\,{=}\,$2048 to accommodate the seed and reasoning trace. We then extract the final answer from the requested answer channel or tag before applying the same parsing and evaluation code used for direct sampling. Results for this baseline are reported in Appendix~\ref{sec:appendix-ssot}.

\subsection{Open-Ended Random Generation}
\label{app:prompts-open}

All prompts follow the template:

\begin{quote}
\small
\texttt{[Verb]} a random \texttt{[category]}. Output ONLY the answer.
\end{quote}

\noindent where \texttt{[Verb]} cycles through ``Think of,'' ``Choose,'' ``Name,'' and ``Pick.'' Representative examples include:

\begin{quote}
\small
Choose a random first name. Output ONLY the answer.\\
Name a random city. Output ONLY the answer.\\
Think of a random animal. Output ONLY the answer.\\
Pick a random musical instrument. Output ONLY the answer.\\
\end{quote}

\noindent The full set of 102 prompts spans categories including words, names, cities, countries, animals, foods, jobs, hobbies, sports, colors, flowers, trees, diseases, body parts, musical instruments, vehicles, gemstones, programming languages, and more.

\subsection{MCQ Generation}

\label{app:mcq-prompt}

\noindent The exact prompt used for MCQ generation evaluation is:

\begin{quote}
\small
\ttfamily
\raggedright
You are a medical education expert who creates high-quality multiple-choice questions for medical students and professionals.\par
\medskip
Please generate a medical multiple-choice question (single answer, 4 options). The question should cover medical knowledge and be of moderate difficulty.\par
\medskip
Please strictly follow this format:\par
Question: [Question content]\par
A. [Option A content]\par
B. [Option B content]\par
C. [Option C content]\par
D. [Option D content]\par
Correct Answer: [A/B/C/D]\par
Explanation: [Brief explanation]\par
\medskip
Requirements:\par
(1) The question should have practical medical value.\par
(2) All four options should be plausible with reasonable distractors.\par
(3) Only one correct answer.\par
(4) Output directly without any additional content.\par
(5) Cover different medical knowledge areas (e.g., internal medicine, surgery, pharmacology, pathology, diagnostics).\par
(6) The correct answer should be evenly distributed among A, B, C, D options to avoid bias toward any particular option.\par
\end{quote}

For MCQ generation, we use \texttt{max\_new\_tokens}$\,{=}\,$256. A regex-based strict parser validates each generation; any generation with a missing, duplicated, or malformed field is counted as a parse failure. No constrained decoding is applied.

\clearpage
\section{Additional Experimental Results}
\suppressfloats[t]  %

\subsection{Structured Distribution Sampling: Held-Out Families}

\begin{figure}[htbp]
\centering
\includegraphics[width=\linewidth]{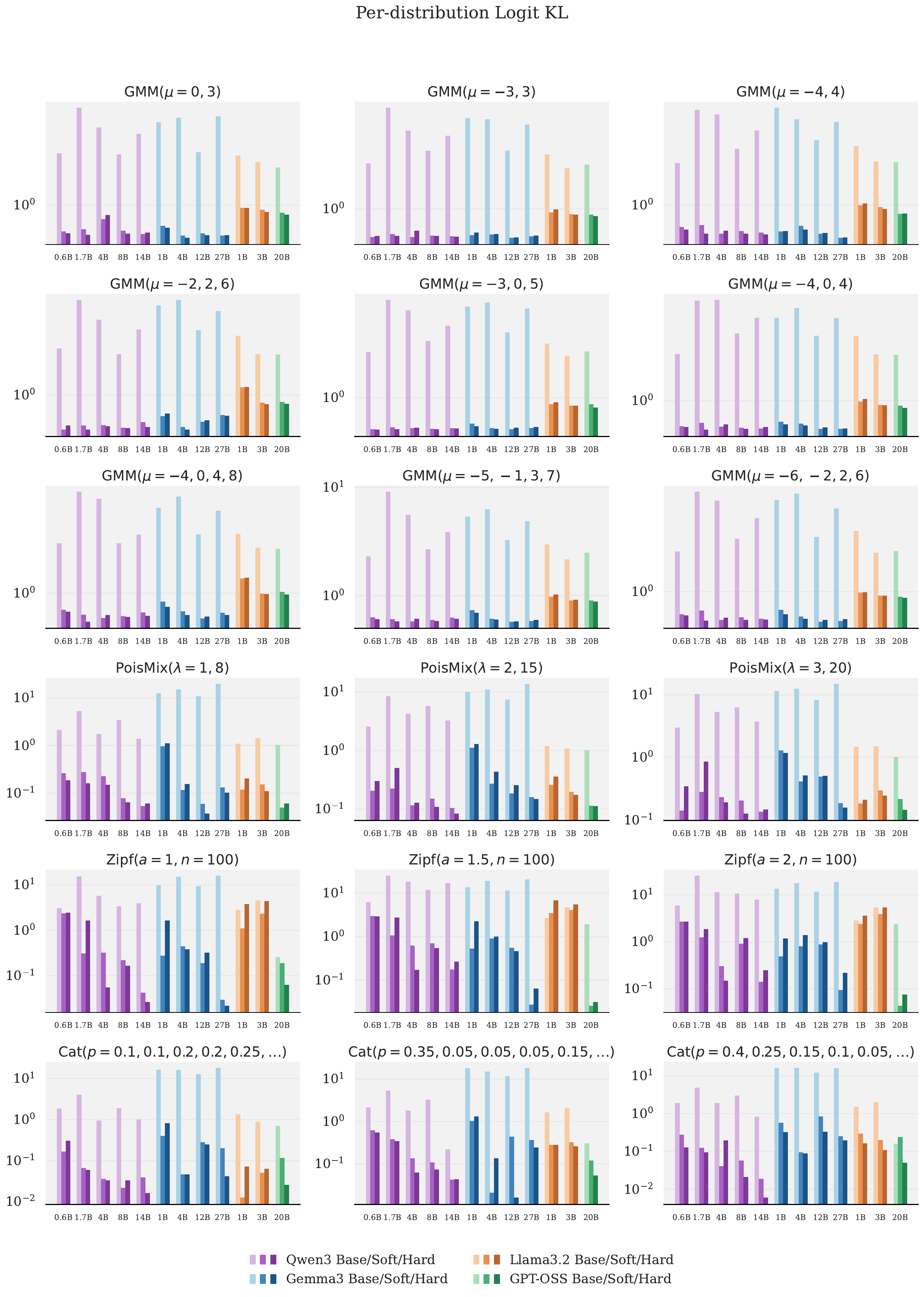}
\caption{\textbf{Per-distribution logit-level KL on the held-out families, part 2 of 2} (Gaussian mixtures with two, three, and four modes, bimodal Poisson mixture, Zipf, Categorical), computed against the shared five-decimal, max-bins-\(16384\) trie target. Each panel isolates one held-out prompt and compares token-level calibration for the base, soft-target, and hard-target models.}
\label{fig:appendix-novel-kl-per-dist}
\end{figure}

Table~\ref{tab:main-ood} reports the full held-out-family structured-sampling results across all 12 models. Figures~\ref{fig:appendix-ood-w1-per-dist}--\ref{fig:appendix-novel-kl-per-dist} give per-distribution breakdowns across the 12 held-out families. Hatched bars in the \(W_1\) plots indicate distribution/model pairs for which no finite \(W_1\) estimate is available because no valid parsed samples were obtained. Unless otherwise stated, all appendix logit-KL results use the same trie-derived evaluation target as the main text, built from five-decimal canonical outputs with max bins \(=16384\). Because this evaluation target is shared across methods, logit KL is directly comparable across Base, Soft, and Hard; however, it is intentionally higher-resolution than the 1001-bin target used during soft-target training.

\begin{table*}[t]
\centering
\footnotesize
\begin{tabular}{ll rrr}
\toprule
Model & & Logit KL $\downarrow$ & $W_1$ $\downarrow$ & Valid (\%) $\uparrow$ \\
\midrule
Qwen3-0.6B & Base & \ci{2.688}{2.218}{3.360} & \ci{0.313}{0.195}{0.583} & \ci{9.5}{3.2}{18.1} \\
 & Soft & \ci{0.675}{0.391}{1.090} & \ci{0.237}{0.159}{86.121} & \ci{74.9}{53.9}{94.3} \\
 & Hard & \ci{0.653}{0.380}{1.053} & \ci{0.209}{0.134}{0.298} & \ci{75.7}{53.5}{93.7} \\
\midrule
Qwen3-1.7B & Base & \ci{6.524}{4.205}{9.657} & \ci{0.356}{0.260}{0.477} & \ci{84.8}{65.8}{98.7} \\
 & Soft & \ci{0.433}{0.292}{0.589} & \ci{0.177}{0.107}{0.313} & \ci{92.9}{85.1}{98.6} \\
 & Hard & \ci{0.536}{0.320}{0.870} & \ci{0.166}{0.117}{0.314} & \ci{90.3}{74.1}{99.8} \\
\midrule
Qwen3-4B & Base & \ci{4.713}{3.210}{6.515} & \ci{0.332}{0.271}{0.398} & \ci{92.9}{77.6}{100.0} \\
 & Soft & \ci{0.377}{0.259}{0.490} & \ci{0.154}{0.119}{0.218} & \ci{99.1}{97.7}{99.8} \\
 & Hard & \ci{0.356}{0.214}{0.494} & \ci{0.163}{0.115}{0.224} & \ci{99.8}{99.5}{100.0} \\
\midrule
Qwen3-8B & Base & \ci{3.330}{2.368}{4.537} & \ci{0.322}{0.247}{0.570} & \ci{93.3}{81.9}{99.8} \\
 & Soft & \ci{0.365}{0.224}{0.499} & \ci{0.140}{0.085}{0.189} & \ci{99.4}{98.4}{99.9} \\
 & Hard & \ci{0.352}{0.215}{0.487} & \ci{0.120}{0.071}{0.159} & \ci{99.7}{99.1}{100.0} \\
\midrule
Qwen3-14B & Base & \ci{3.272}{2.132}{4.862} & \ci{0.246}{0.225}{0.316} & \ci{95.5}{84.5}{100.0} \\
 & Soft & \ci{0.314}{0.182}{0.442} & \ci{0.097}{0.071}{0.139} & \ci{99.8}{99.5}{100.0} \\
 & Hard & \ci{0.314}{0.189}{0.435} & \ci{0.116}{0.067}{0.145} & \ci{99.9}{99.7}{100.0} \\
\midrule
Gemma-3-1B-it & Base & \ci{8.851}{6.072}{11.658} & \ci{0.590}{0.328}{59.737} & \ci{69.1}{45.5}{91.2} \\
 & Soft & \ci{0.715}{0.553}{0.924} & \ci{0.242}{0.165}{0.360} & \ci{67.0}{47.8}{86.7} \\
 & Hard & \ci{1.008}{0.667}{1.460} & \ci{0.287}{0.186}{0.445} & \ci{60.1}{35.7}{84.1} \\
\midrule
Gemma-3-4B-it & Base & \ci{10.115}{7.223}{13.178} & \ci{0.370}{0.286}{0.562} & \ci{95.0}{89.1}{99.0} \\
 & Soft & \ci{0.477}{0.323}{0.660} & \ci{0.244}{0.138}{0.398} & \ci{89.1}{76.9}{97.8} \\
 & Hard & \ci{0.421}{0.278}{0.570} & \ci{0.183}{0.122}{0.292} & \ci{95.9}{89.0}{99.7} \\
\midrule
Gemma-3-12B-it & Base & \ci{6.236}{4.394}{8.244} & \ci{0.247}{0.206}{0.287} & \ci{94.9}{87.7}{99.4} \\
 & Soft & \ci{0.407}{0.282}{0.514} & \ci{0.135}{0.078}{0.218} & \ci{99.6}{98.8}{100.0} \\
 & Hard & \ci{0.385}{0.256}{0.496} & \ci{0.156}{0.077}{0.237} & \ci{99.5}{99.0}{99.9} \\
\midrule
Gemma-3-27B-it & Base & \ci{10.265}{7.166}{13.658} & \ci{0.241}{0.200}{0.301} & \ci{87.2}{68.7}{99.6} \\
 & Soft & \ci{0.342}{0.209}{0.465} & \ci{0.104}{0.081}{0.277} & \ci{99.9}{99.8}{100.0} \\
 & Hard & \ci{0.338}{0.213}{0.464} & \ci{0.120}{0.072}{0.258} & \ci{99.8}{99.5}{100.0} \\
\midrule
Llama-3.2-1B-it & Base & \ci{2.666}{2.263}{3.017} & \ci{63.516}{1.528}{134.4} & \ci{52.7}{31.2}{72.3} \\
 & Soft & \ci{0.782}{0.485}{1.166} & \ci{0.339}{0.168}{18.097} & \ci{82.8}{66.6}{95.8} \\
 & Hard & \ci{0.963}{0.478}{1.749} & \ci{0.260}{0.169}{0.408} & \ci{82.4}{60.9}{98.2} \\
\midrule
Llama-3.2-3B-it & Base & \ci{2.530}{1.900}{3.302} & \ci{5.168}{0.432}{16.182} & \ci{62.6}{38.7}{85.8} \\
 & Soft & \ci{0.806}{0.428}{1.349} & \ci{0.196}{0.113}{0.333} & \ci{89.2}{72.7}{98.8} \\
 & Hard & \ci{0.920}{0.381}{1.741} & \ci{0.186}{0.105}{0.295} & \ci{90.8}{73.9}{99.8} \\
\midrule
GPT-OSS-20B & Base & \ci{1.498}{0.967}{1.948} & \ci{0.635}{0.200}{$3.3{\times}10^{3}$} & \ci{99.0}{98.0}{99.4} \\
 & Soft & \ci{0.497}{0.305}{0.704} & \ci{0.149}{0.083}{0.191} & \ci{99.8}{99.3}{100.0} \\
 & Hard & \ci{0.468}{0.262}{0.674} & \ci{0.112}{0.053}{0.139} & \ci{99.8}{99.4}{100.0} \\
\bottomrule
\end{tabular}
\caption{\textbf{Structured-sampling results on the 12 held-out distribution families (36 prompts; Table~\ref{tab:benchmark-distributions}).} Base is the original checkpoint, Soft and Hard the soft-target and hard-target Calibration Fine-Tuning checkpoints. Logit KL is the trie-based logit-evaluation average against the shared five-decimal, max-bins-\(16384\) target, averaged within family and then across families. $W_1$ is the $Q_{95}$-$Q_{05}$ normalized Wasserstein-1 distance, averaged within family and aggregated across families with a median. Valid is the percentage of generations that parse to in-support numeric outputs. Point estimate with its 95\% hierarchical bootstrap interval ($B=2000$; families, then distributions within family) in smaller type. Intervals resample evaluation prompts, not training runs.}
\label{tab:main-ood}
\end{table*}

\begin{figure*}[t]
\centering
\includegraphics[width=\textwidth]{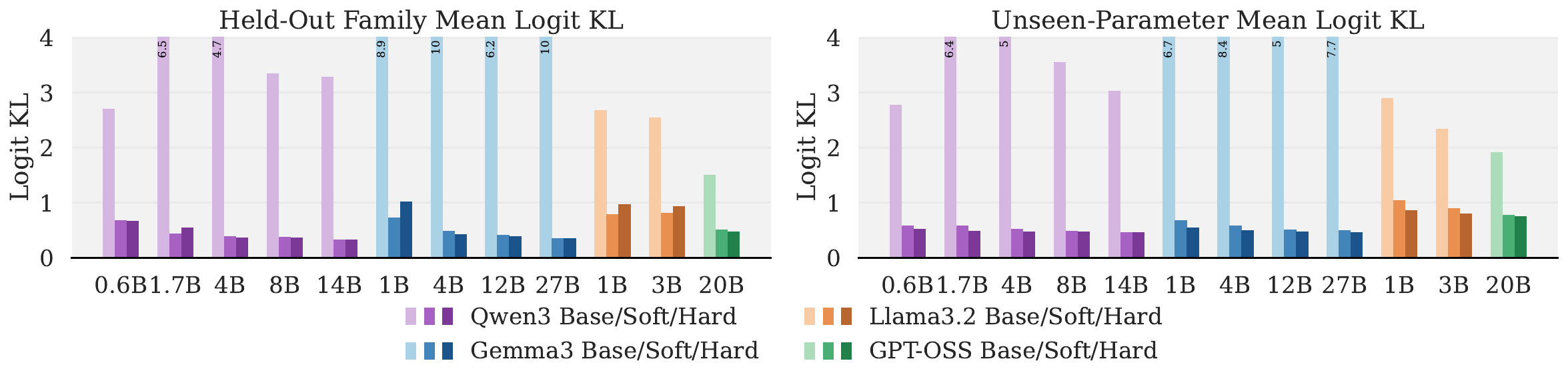}
\caption{\textbf{Logit-level structured-sampling result.} Trie-based logit KL between the model's next-token distribution and the target, averaged within family and then across families. \textbf{Left}: the 12 held-out families; \textbf{right}: unseen parameter settings of the 24 training families. Bars that exceed the axis are clipped and labeled with their value.}
\label{fig:appendix-structured-kl}
\end{figure*}

\clearpage

\begin{center}
\begin{minipage}{\linewidth}
\centering
\includegraphics[width=\textwidth]{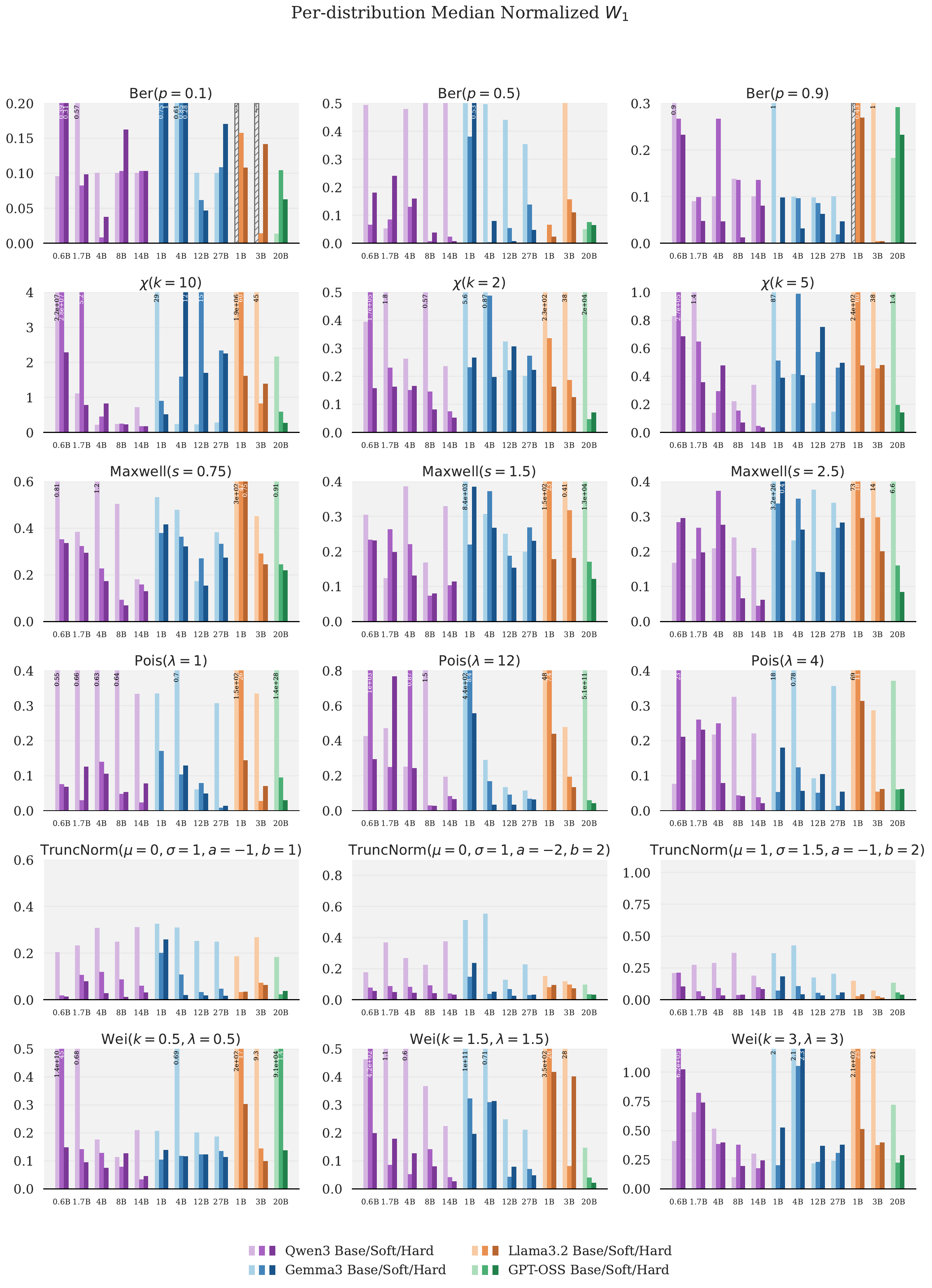}
\captionof{figure}{\textbf{Per-distribution median normalized \(W_1\) on the held-out families, part 1 of 2} (Bernoulli, Poisson, Maxwell, TruncNorm, Chi, Weibull). Each panel corresponds to one held-out prompt and compares sample fidelity for the base, soft-target, and hard-target models.}
\label{fig:appendix-ood-w1-per-dist}
\end{minipage}
\end{center}

\begin{figure}[htbp]
\centering
\includegraphics[width=\linewidth]{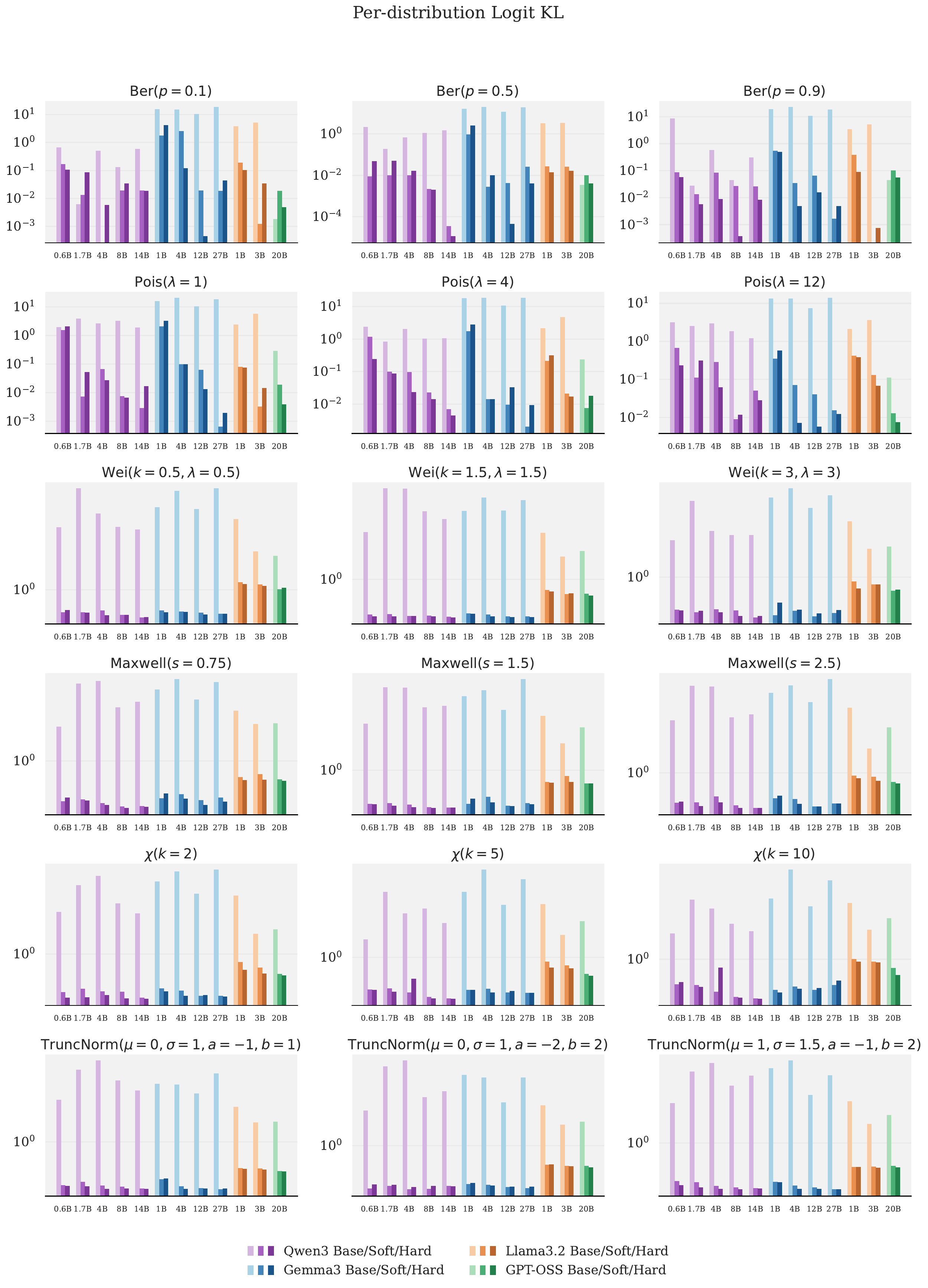}
\caption{\textbf{Per-distribution logit-level KL on the held-out families, part 1 of 2} (Bernoulli, Poisson, Maxwell, TruncNorm, Chi, Weibull), computed against the shared five-decimal, max-bins-\(16384\) trie target. Each panel isolates one held-out prompt and compares token-level calibration for the base, soft-target, and hard-target models.}
\label{fig:appendix-ood-kl-per-dist}
\end{figure}

\begin{figure}[htbp]
\centering
\includegraphics[width=\linewidth]{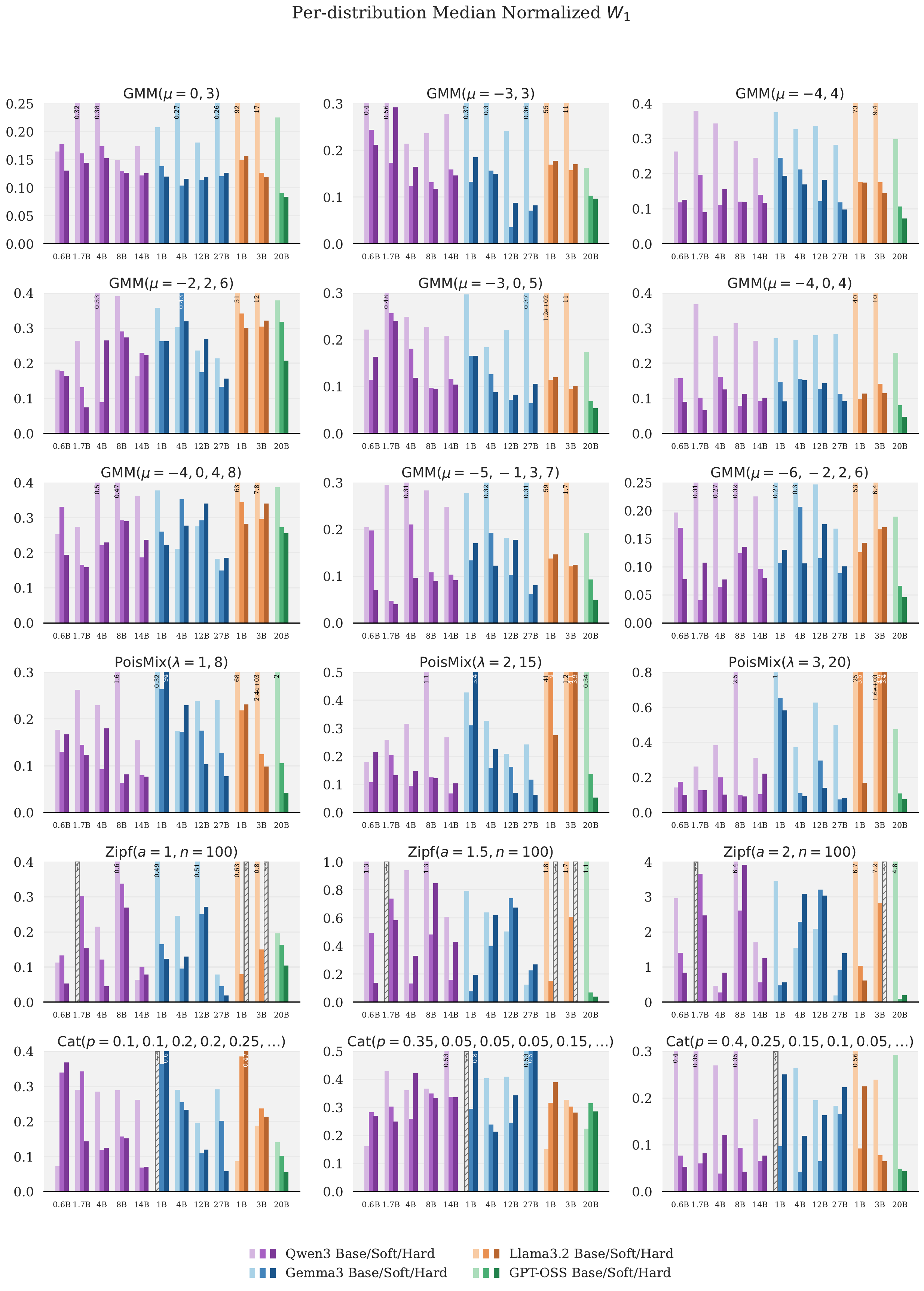}
\caption{\textbf{Per-distribution median normalized \(W_1\) on the held-out families, part 2 of 2} (Gaussian mixtures with two, three, and four modes, bimodal Poisson mixture, Zipf, Categorical). Each panel corresponds to one held-out prompt and compares sample fidelity for the base, soft-target, and hard-target models.}
\label{fig:appendix-novel-w1-per-dist}
\end{figure}

\clearpage
\subsection{Structured Distribution Sampling: Unseen Parameters}

\begin{figure}[htbp]
\centering
\includegraphics[width=\linewidth,height=0.78\textheight,keepaspectratio]{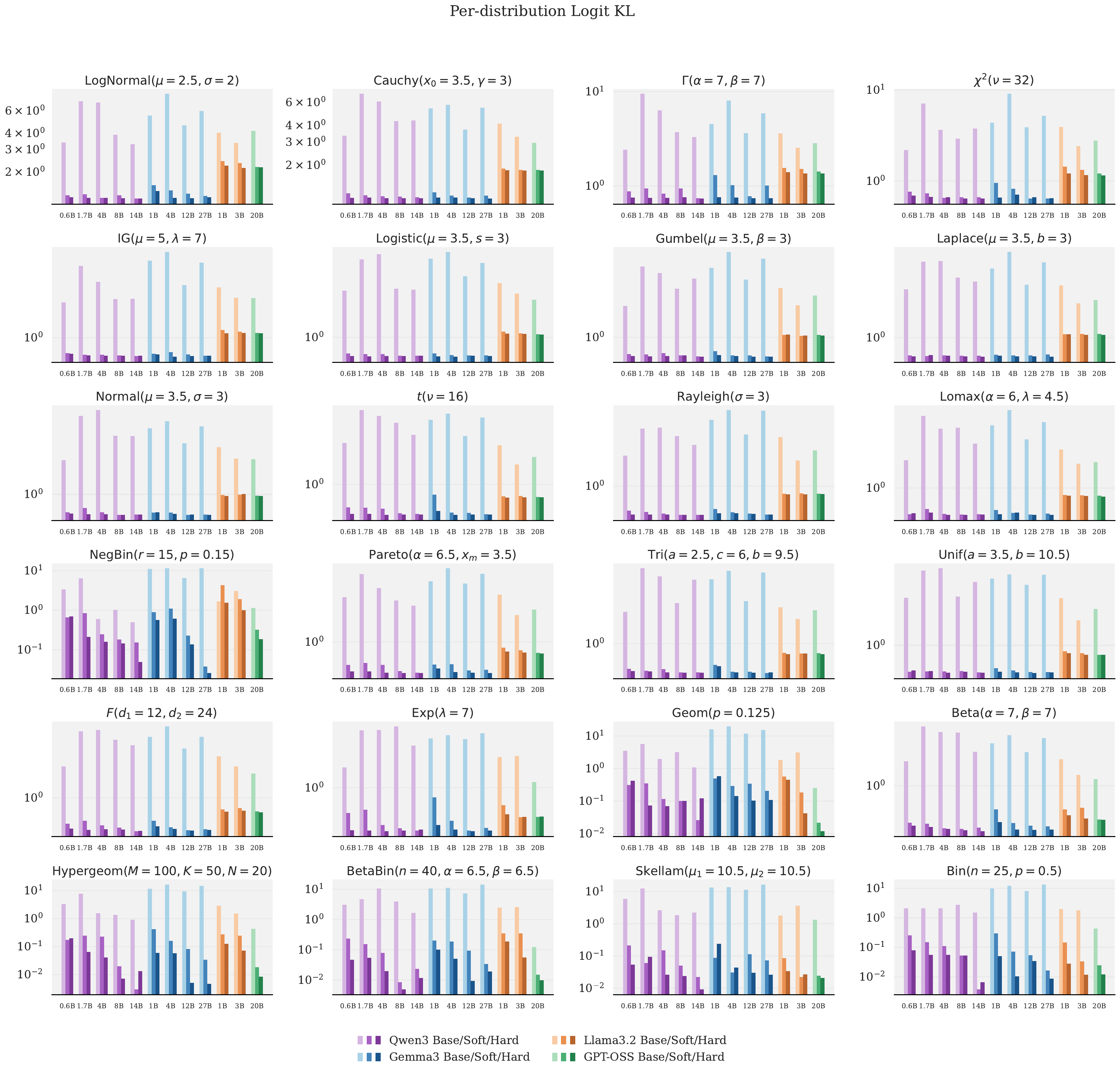}
\caption{\textbf{Per-distribution logit-level KL on the unseen-parameter structured-sampling benchmark, computed against the shared five-decimal, max-bins-\(16384\) trie target.} Each panel shows calibration quality for one unseen parameter setting under the base, soft-target, and hard-target models.}
\label{fig:appendix-unseen-kl-per-dist}
\end{figure}

\begin{figure}[htbp]
\centering
\includegraphics[width=\linewidth,height=0.78\textheight,keepaspectratio]{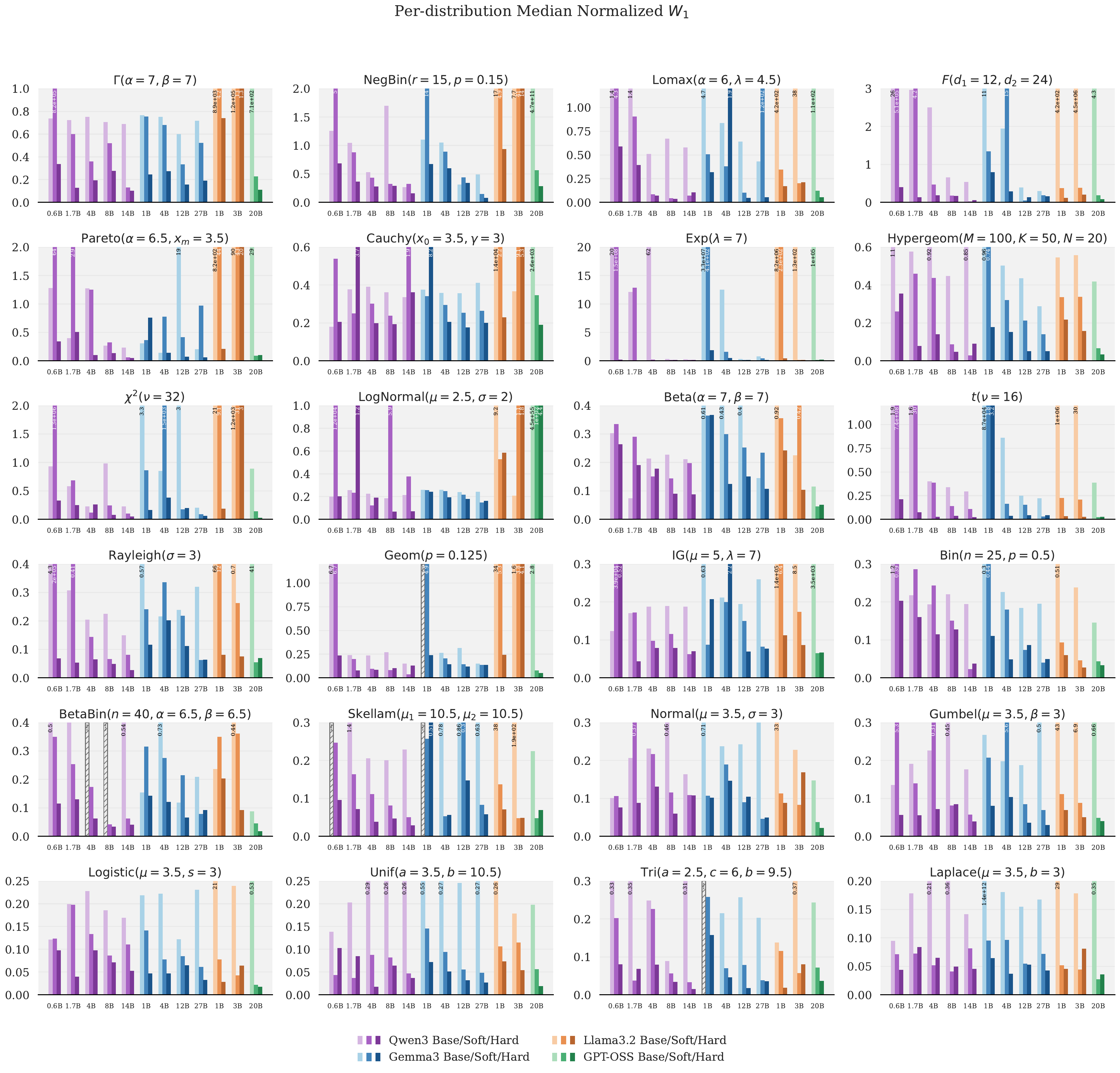}
\caption{\textbf{Per-distribution median normalized \(W_1\) on the unseen-parameter structured-sampling benchmark.} Each panel corresponds to one unseen parameter setting and compares sample fidelity for the base, soft-target, and hard-target models.}
\label{fig:appendix-unseen-w1-per-dist}
\end{figure}

Table~\ref{tab:main-unseen} reports the unseen-parameter evaluation, where models are tested on parameter settings outside the training grid for seen distribution families. Figures~\ref{fig:appendix-unseen-kl-per-dist}--\ref{fig:appendix-unseen-w1-per-dist} give per-distribution breakdowns of logit-level KL and sample-level \(W_1\). Entries marked ``--'' in Table~\ref{tab:main-unseen} indicate aggregate \(W_1\) values that are undefined because at least one unseen family has no finite \(W_1\) estimate.

\begin{table*}[t]
\centering
\footnotesize
\begin{tabular}{ll rrr}
\toprule
Model & & Logit KL $\downarrow$ & $W_1$ $\downarrow$ & Valid (\%) $\uparrow$ \\
\midrule
Qwen3-0.6B & Base & \ci{2.761}{2.468}{3.106} & \ci{0.734}{0.197}{1.252} & \ci{3.1}{2.2}{4.1} \\
 & Soft & \ci{0.572}{0.470}{0.691} & \ci{4.640}{0.334}{$1.9{\times}10^{4}$} & \ci{96.1}{91.9}{98.7} \\
 & Hard & \ci{0.519}{0.403}{0.636} & \ci{0.203}{0.106}{0.325} & \ci{99.6}{99.3}{99.8} \\
\midrule
Qwen3-1.7B & Base & \ci{6.351}{5.622}{7.168} & \ci{0.362}{0.217}{0.646} & \ci{82.9}{67.2}{95.5} \\
 & Soft & \ci{0.578}{0.462}{0.699} & \ci{0.287}{0.197}{0.569} & \ci{96.8}{92.2}{99.5} \\
 & Hard & \ci{0.476}{0.352}{0.597} & \ci{0.086}{0.071}{0.159} & \ci{99.8}{99.7}{99.9} \\
\midrule
Qwen3-4B & Base & \ci{4.978}{4.041}{5.921} & \ci{0.232}{0.220}{0.396} & \ci{94.2}{84.8}{100.0} \\
 & Soft & \ci{0.511}{0.396}{0.639} & \ci{0.175}{0.120}{0.300} & \ci{99.0}{97.6}{99.8} \\
 & Hard & \ci{0.464}{0.341}{0.603} & \ci{0.089}{0.070}{0.138} & \ci{99.9}{99.8}{99.9} \\
\midrule
Qwen3-8B & Base & \ci{3.543}{2.994}{4.141} & \ci{0.268}{0.224}{0.452} & \ci{88.1}{75.7}{97.0} \\
 & Soft & \ci{0.483}{0.348}{0.624} & \ci{0.115}{0.081}{0.152} & \ci{99.8}{99.6}{99.9} \\
 & Hard & \ci{0.462}{0.337}{0.596} & \ci{0.071}{0.049}{0.089} & \ci{100.0}{100.0}{100.0} \\
\midrule
Qwen3-14B & Base & \ci{3.022}{2.519}{3.541} & \ci{0.227}{0.194}{0.292} & \ci{98.0}{94.4}{100.0} \\
 & Soft & \ci{0.459}{0.331}{0.600} & \ci{0.074}{0.055}{0.104} & \ci{99.9}{99.9}{99.9} \\
 & Hard & \ci{0.454}{0.324}{0.589} & \ci{0.053}{0.039}{0.088} & \ci{100.0}{100.0}{100.0} \\
\midrule
Gemma-3-1B-it & Base & \ci{6.676}{5.504}{8.013} & \ci{0.631}{0.374}{1.096} & \ci{66.3}{48.2}{82.6} \\
 & Soft & \ci{0.670}{0.548}{0.797} & \ci{0.350}{0.255}{0.740} & \ci{86.7}{78.8}{93.4} \\
 & Hard & \ci{0.539}{0.421}{0.663} & \ci{0.222}{0.142}{0.375} & \ci{99.1}{98.3}{99.7} \\
\midrule
Gemma-3-4B-it & Base & \ci{8.405}{7.135}{9.897} & \ci{0.312}{0.225}{0.749} & \ci{75.3}{61.6}{86.7} \\
 & Soft & \ci{0.572}{0.440}{0.716} & \ci{0.283}{0.189}{0.375} & \ci{98.3}{97.6}{98.9} \\
 & Hard & \ci{0.491}{0.371}{0.623} & \ci{0.142}{0.103}{0.236} & \ci{99.8}{99.7}{99.8} \\
\midrule
Gemma-3-12B-it & Base & \ci{5.001}{4.100}{6.027} & \ci{0.269}{0.239}{0.381} & \ci{97.6}{95.7}{99.2} \\
 & Soft & \ci{0.496}{0.374}{0.617} & \ci{0.149}{0.089}{0.215} & \ci{99.7}{99.3}{99.9} \\
 & Hard & \ci{0.459}{0.334}{0.588} & \ci{0.086}{0.064}{0.127} & \ci{99.9}{99.8}{100.0} \\
\midrule
Gemma-3-27B-it & Base & \ci{7.668}{6.263}{9.235} & \ci{0.265}{0.208}{0.319} & \ci{98.4}{96.9}{99.6} \\
 & Soft & \ci{0.485}{0.358}{0.621} & \ci{0.083}{0.068}{0.146} & \ci{99.9}{99.8}{99.9} \\
 & Hard & \ci{0.455}{0.330}{0.587} & \ci{0.057}{0.049}{0.091} & \ci{99.9}{99.9}{100.0} \\
\midrule
Llama-3.2-1B-it & Base & \ci{2.885}{2.612}{3.147} & \ci{33.654}{16.999}{416.4} & \ci{63.9}{49.2}{78.2} \\
 & Soft & \ci{1.034}{0.742}{1.385} & \ci{0.351}{0.136}{4.171} & \ci{97.8}{96.1}{99.1} \\
 & Hard & \ci{0.847}{0.633}{1.060} & \ci{0.140}{0.073}{0.216} & \ci{99.2}{98.5}{99.8} \\
\midrule
Llama-3.2-3B-it & Base & \ci{2.326}{2.091}{2.558} & \ci{1.166}{0.366}{23.413} & \ci{68.3}{52.8}{82.6} \\
 & Soft & \ci{0.888}{0.659}{1.143} & \ci{0.233}{0.114}{0.737} & \ci{98.5}{97.3}{99.4} \\
 & Hard & \ci{0.788}{0.574}{0.998} & \ci{0.114}{0.079}{0.207} & \ci{99.6}{99.2}{99.9} \\
\midrule
GPT-OSS-20B & Base & \ci{1.910}{1.525}{2.317} & \ci{0.773}{0.348}{40.623} & \ci{96.6}{92.0}{99.0} \\
 & Soft & \ci{0.762}{0.533}{0.995} & \ci{0.064}{0.047}{0.101} & \ci{99.9}{99.7}{100.0} \\
 & Hard & \ci{0.743}{0.524}{0.968} & \ci{0.048}{0.033}{0.073} & \ci{99.9}{99.8}{100.0} \\
\bottomrule
\end{tabular}
\caption{\textbf{Structured-sampling results on unseen parameter settings of the 24 training families.} Columns as in Table~\ref{tab:main-ood}. Point estimate with its 95\% hierarchical bootstrap interval ($B=2000$; families, then distributions within family) in smaller type. Intervals resample evaluation prompts, not training runs.}
\label{tab:main-unseen}
\end{table*}


\clearpage
\clearpage
\subsection{Open-Ended Random Generation}

Table~\ref{tab:opengen-summary} reports the aggregate open-generation results for all 12 models with bootstrap intervals.

\begin{table*}[t]
\centering
\footnotesize
\begin{tabular}{llrr}
\toprule
Model & & Top-90\% first-token support $\uparrow$ & Unique rate (\%) $\uparrow$ \\
\midrule
Qwen3-0.6B & Base & \ci{45.0}{30.7}{61.9} & \ci{56.9}{52.7}{61.0} \\
 & Soft & \ci{711.2}{573.8}{860.2} & \ci{45.3}{40.3}{50.2} \\
 & Hard & \ci{501.6}{392.3}{628.7} & \ci{45.9}{40.5}{51.0} \\
\midrule
Qwen3-1.7B & Base & \ci{8.8}{7.0}{11.1} & \ci{22.5}{19.6}{25.7} \\
 & Soft & \ci{864.5}{650.1}{$1.1{\times}10^{3}$} & \ci{39.1}{34.5}{43.9} \\
 & Hard & \ci{265.0}{193.4}{350.7} & \ci{30.3}{26.4}{34.6} \\
\midrule
Qwen3-4B & Base & \ci{4.4}{3.5}{5.5} & \ci{11.9}{10.3}{13.8} \\
 & Soft & \ci{299.6}{190.0}{440.3} & \ci{27.7}{24.3}{31.4} \\
 & Hard & \ci{127.8}{65.6}{220.7} & \ci{21.4}{18.1}{24.8} \\
\midrule
Qwen3-8B & Base & \ci{3.3}{2.7}{4.0} & \ci{9.5}{7.7}{11.4} \\
 & Soft & \ci{466.9}{256.2}{787.0} & \ci{37.9}{33.8}{42.3} \\
 & Hard & \ci{219.6}{101.4}{398.4} & \ci{27.8}{23.9}{32.0} \\
\midrule
Qwen3-14B & Base & \ci{3.5}{2.9}{4.1} & \ci{10.9}{9.2}{12.6} \\
 & Soft & \ci{536.2}{238.3}{$1.0{\times}10^{3}$} & \ci{49.1}{44.4}{53.6} \\
 & Hard & \ci{194.0}{111.7}{309.8} & \ci{35.9}{31.9}{40.1} \\
\midrule
Gemma-3-1B-it & Base & \ci{5.5}{3.7}{8.0} & \ci{13.3}{11.2}{15.5} \\
 & Soft & \ci{52.6}{8.9}{130.3} & \ci{16.8}{13.4}{20.6} \\
 & Hard & \ci{36.6}{7.2}{90.3} & \ci{17.3}{13.9}{21.1} \\
\midrule
Gemma-3-4B-it & Base & \ci{2.6}{2.3}{3.0} & \ci{6.9}{5.7}{8.2} \\
 & Soft & \ci{115.0}{93.0}{138.9} & \ci{49.2}{45.0}{53.3} \\
 & Hard & \ci{22.9}{18.4}{28.0} & \ci{25.0}{21.8}{28.5} \\
\midrule
Gemma-3-12B-it & Base & \ci{3.3}{2.8}{3.9} & \ci{7.4}{6.4}{8.5} \\
 & Soft & \ci{44.8}{32.8}{59.0} & \ci{31.8}{27.8}{35.7} \\
 & Hard & \ci{23.6}{16.7}{33.7} & \ci{21.4}{18.2}{24.5} \\
\midrule
Gemma-3-27B-it & Base & \ci{2.3}{1.9}{2.7} & \ci{5.8}{4.9}{6.6} \\
 & Soft & \ci{63.0}{46.9}{83.0} & \ci{40.1}{36.3}{44.1} \\
 & Hard & \ci{39.4}{28.3}{54.3} & \ci{30.0}{26.7}{33.4} \\
\midrule
Llama-3.2-1B-it & Base & \ci{84.6}{70.1}{101.9} & \ci{52.5}{49.2}{55.9} \\
 & Soft & \ci{86.2}{68.3}{107.2} & \ci{17.7}{15.2}{20.1} \\
 & Hard & \ci{83.9}{68.1}{100.9} & \ci{18.4}{16.0}{21.0} \\
\midrule
Llama-3.2-3B-it & Base & \ci{28.9}{23.8}{34.7} & \ci{36.4}{33.1}{39.7} \\
 & Soft & \ci{85.6}{69.6}{103.0} & \ci{18.8}{16.4}{21.5} \\
 & Hard & \ci{78.9}{62.9}{97.7} & \ci{17.9}{15.5}{20.7} \\
\midrule
GPT-OSS-20B & Base & \ci{28.0}{21.2}{36.9} & \ci{36.2}{32.8}{39.8} \\
 & Soft & \ci{289.0}{221.5}{374.0} & \ci{73.0}{68.7}{77.4} \\
 & Hard & \ci{212.4}{156.1}{284.0} & \ci{77.1}{73.2}{80.7} \\
\bottomrule
\end{tabular}
\caption{\textbf{Open-ended random generation across all evaluated models.} Top-90\% support is the number of first-step next tokens needed to cover 90\% of the model's probability mass, averaged over the 102 prompts; unique rate is the fraction of distinct outputs among 100 samples per prompt. Intervals are 95\% bootstrap intervals ($B=2000$), shown in smaller type after the point estimate. Intervals resample prompts.}
\label{tab:opengen-summary}
\end{table*}

\begin{figure}[htbp]
\centering
\includegraphics[width=\linewidth]{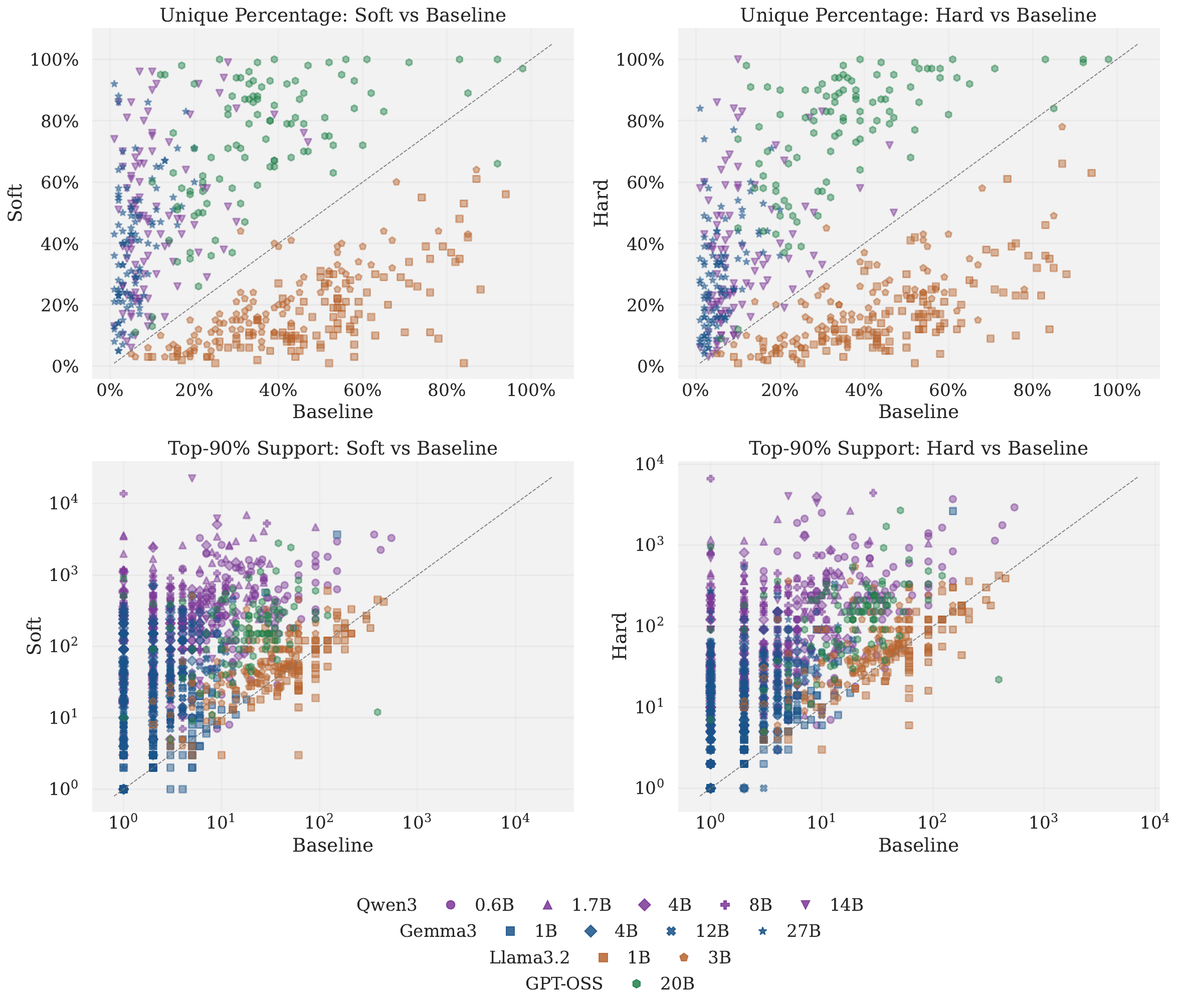}
\caption{\textbf{Per-prompt open-generation comparisons at the 90\% probability-mass threshold.} Each point is one prompt/model pair and compares first-step next-token support breadth or unique-output percentage between the base model and one fine-tuned variant (soft-target or hard-target Calibration Fine-Tuning). Points are colored by model family and use marker shape to denote model size.}
\label{fig:appendix-open-multi-threshold}
\end{figure}

Figure~\ref{fig:appendix-open-multi-threshold} gives the prompt-level view underlying the aggregate open-generation results. Each point corresponds to one prompt/model pair, comparing the base checkpoint to either soft-target or hard-target Calibration Fine-Tuning. The clearest effect is on first-token support: soft-target fine-tuning increases top-90\% next-token support on 1101/1224 prompt/model pairs, across 102 prompts and 12 models, while hard-target fine-tuning does so on 1089/1224 pairs. The effect on realized output diversity is more model-dependent. Unique-output rate increases for most Qwen, Gemma, and GPT-OSS prompts, but decreases for Llama prompts, showing that broader first-token support does not always translate into more diverse full completions.

\begin{figure}[htbp]
\centering
\includegraphics[width=\linewidth]{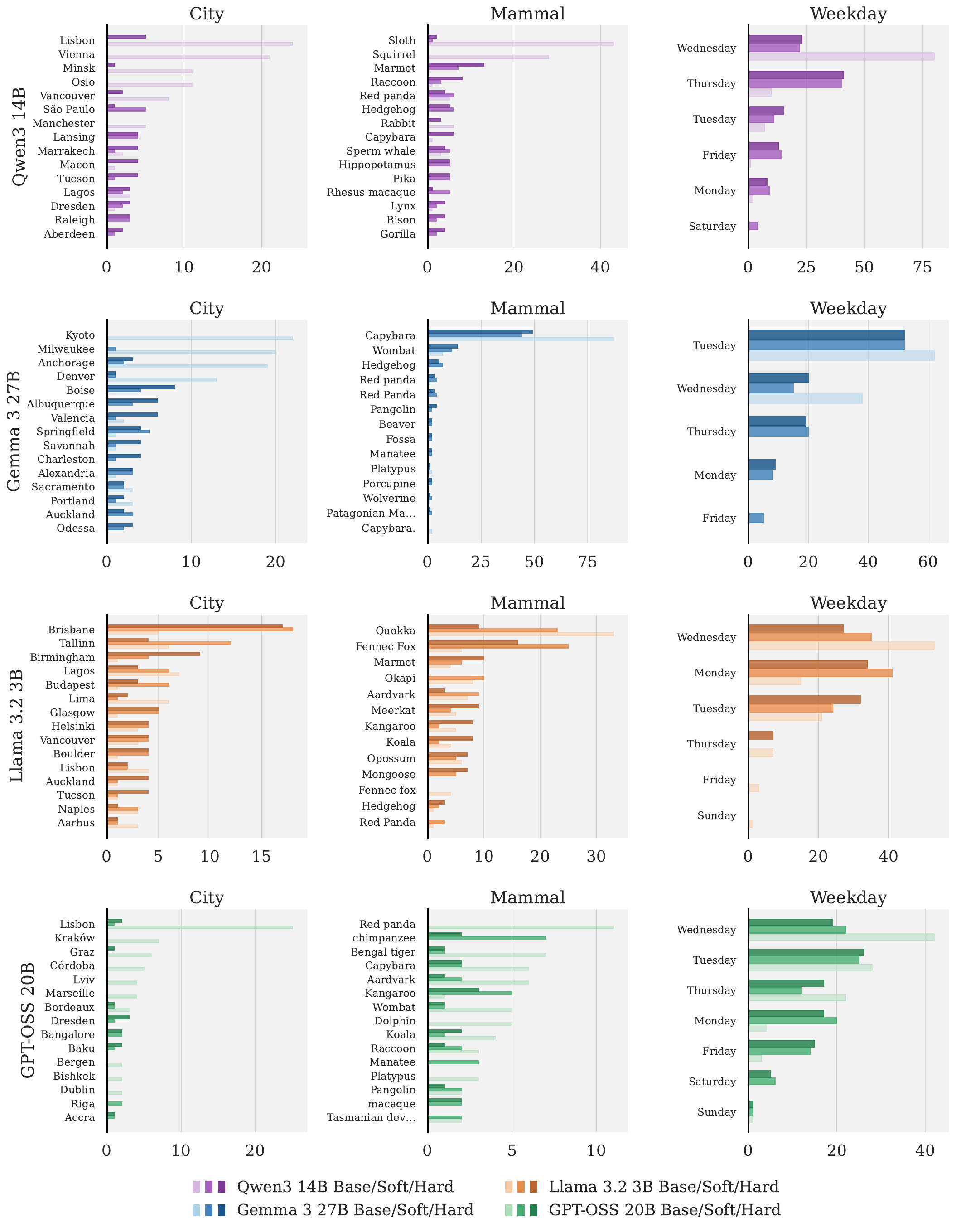}
\caption{\textbf{Qualitative open-generation examples from repeated sampling.} For selected prompts (city, mammal, weekday) and one strong model from each family, we compare empirical output frequencies under the base, soft-target, and hard-target models. To choose which outputs to display, we take the top 10 outputs from each state, union them, rank the union by peak then total empirical frequency across the three states, and show the top 15 resulting labels.}
\label{fig:appendix-open-qualitative}
\end{figure}

Figure~\ref{fig:appendix-open-qualitative} shows that these aggregate gains correspond to visibly less concentrated empirical output distributions. For Qwen3-14B on the weekday prompt, the base model assigns 80\% of samples to Wednesday, whereas the soft-target and hard-target checkpoints spread mass across several weekdays, with the top answer falling to roughly 40\%. The same pattern appears in larger open supports: for Gemma-3-27B-it on the city prompt, the base model places 74\% of samples on its top four cities, while the soft-target and hard-target checkpoints reduce this concentration to 15\% and 24\%, respectively. GPT-OSS-20B exhibits an even stronger version of this behavior on cities and mammals, where the calibrated checkpoints make the empirical distribution close to flat over the displayed labels. Overall, the qualitative examples support the main open-generation conclusion: Calibration Fine-Tuning often spreads probability mass over plausible alternatives, although the effect is weaker for the Llama checkpoints.

\clearpage
\subsection{NoveltyBench}
\label{sec:appendix-noveltybench}

Table~\ref{tab:noveltybench-summary} reports the full NoveltyBench results across all evaluated models, split by benchmark partition. We show both \emph{mean distinct} and \emph{mean utility} on the curated and WildChat splits separately to distinguish raw semantic breadth from the benchmark's reward-weighted overall utility.

\begin{table*}[t]
\centering
\footnotesize
\setlength{\tabcolsep}{4pt}
\begin{tabular}{llrrrr}
\toprule
Model & & Curated distinct $\uparrow$ & Curated utility $\uparrow$ & WildChat distinct $\uparrow$ & WildChat utility $\uparrow$ \\
\midrule
Qwen3-0.6B & Base & \ci{8.85}{8.44}{9.23} & \ci{2.753}{2.36}{3.18} & \ci{7.22}{7.05}{7.40} & \ci{2.627}{2.50}{2.75} \\
 & Soft & \ci{8.53}{8.13}{8.89} & \ci{2.143}{1.83}{2.49} & \ci{8.87}{8.73}{8.98} & \ci{1.605}{1.53}{1.68} \\
 & Hard & \ci{8.69}{8.27}{9.08} & \ci{2.433}{2.10}{2.79} & \ci{8.26}{8.10}{8.41} & \ci{2.000}{1.90}{2.09} \\
\midrule
Qwen3-1.7B & Base & \ci{5.50}{4.98}{6.02} & \ci{3.393}{3.05}{3.75} & \ci{3.93}{3.75}{4.12} & \ci{2.745}{2.63}{2.87} \\
 & Soft & \ci{7.82}{7.39}{8.24} & \ci{3.730}{3.36}{4.08} & \ci{6.54}{6.35}{6.74} & \ci{3.084}{2.96}{3.21} \\
 & Hard & \ci{7.04}{6.53}{7.56} & \ci{4.101}{3.72}{4.51} & \ci{5.18}{4.99}{5.36} & \ci{2.993}{2.86}{3.12} \\
\midrule
Qwen3-4B & Base & \ci{4.07}{3.59}{4.57} & \ci{3.738}{3.39}{4.10} & \ci{3.15}{3.00}{3.32} & \ci{2.833}{2.73}{2.95} \\
 & Soft & \ci{7.07}{6.55}{7.56} & \ci{5.176}{4.76}{5.59} & \ci{4.94}{4.74}{5.13} & \ci{3.398}{3.27}{3.53} \\
 & Hard & \ci{5.93}{5.37}{6.50} & \ci{4.631}{4.20}{5.05} & \ci{4.18}{3.99}{4.36} & \ci{3.191}{3.07}{3.31} \\
\midrule
Qwen3-8B & Base & \ci{4.20}{3.70}{4.67} & \ci{4.214}{3.84}{4.57} & \ci{3.59}{3.42}{3.77} & \ci{3.294}{3.17}{3.41} \\
 & Soft & \ci{6.89}{6.34}{7.42} & \ci{5.246}{4.88}{5.64} & \ci{5.04}{4.85}{5.23} & \ci{3.703}{3.58}{3.83} \\
 & Hard & \ci{5.97}{5.45}{6.49} & \ci{5.008}{4.61}{5.41} & \ci{4.44}{4.25}{4.62} & \ci{3.607}{3.47}{3.74} \\
\midrule
Qwen3-14B & Base & \ci{3.29}{2.91}{3.72} & \ci{3.861}{3.51}{4.25} & \ci{3.35}{3.19}{3.52} & \ci{3.345}{3.23}{3.46} \\
 & Soft & \ci{6.47}{5.97}{6.97} & \ci{5.400}{5.01}{5.77} & \ci{5.20}{5.01}{5.39} & \ci{3.965}{3.83}{4.11} \\
 & Hard & \ci{5.39}{4.81}{6.01} & \ci{4.995}{4.57}{5.42} & \ci{4.35}{4.16}{4.55} & \ci{3.735}{3.61}{3.86} \\
\midrule
Gemma-3-1B-it & Base & \ci{3.78}{3.34}{4.26} & \ci{3.464}{3.12}{3.84} & \ci{2.49}{2.37}{2.61} & \ci{2.187}{2.09}{2.29} \\
 & Soft & \ci{4.89}{4.41}{5.40} & \ci{3.753}{3.36}{4.17} & \ci{3.45}{3.29}{3.61} & \ci{2.435}{2.32}{2.55} \\
 & Hard & \ci{4.54}{4.02}{5.08} & \ci{3.789}{3.41}{4.20} & \ci{3.11}{2.96}{3.27} & \ci{2.400}{2.29}{2.51} \\
\midrule
Gemma-3-4B-it & Base & \ci{2.52}{2.16}{2.91} & \ci{3.109}{2.81}{3.43} & \ci{1.97}{1.87}{2.06} & \ci{2.254}{2.17}{2.33} \\
 & Soft & \ci{4.57}{4.06}{5.10} & \ci{4.535}{4.12}{4.95} & \ci{3.02}{2.88}{3.16} & \ci{2.884}{2.78}{3.00} \\
 & Hard & \ci{3.66}{3.19}{4.15} & \ci{3.999}{3.65}{4.37} & \ci{2.52}{2.40}{2.65} & \ci{2.654}{2.55}{2.76} \\
\midrule
Gemma-3-12B-it & Base & \ci{2.44}{2.17}{2.73} & \ci{3.221}{2.93}{3.50} & \ci{1.92}{1.84}{2.01} & \ci{2.069}{2.00}{2.14} \\
 & Soft & \ci{3.98}{3.46}{4.50} & \ci{4.318}{3.88}{4.77} & \ci{2.72}{2.60}{2.84} & \ci{2.717}{2.62}{2.82} \\
 & Hard & \ci{3.51}{3.06}{3.99} & \ci{4.082}{3.70}{4.47} & \ci{2.40}{2.29}{2.51} & \ci{2.537}{2.45}{2.63} \\
\midrule
Gemma-3-27B-it & Base & \ci{2.03}{1.79}{2.28} & \ci{2.845}{2.59}{3.10} & \ci{1.83}{1.75}{1.91} & \ci{2.212}{2.14}{2.29} \\
 & Soft & \ci{4.01}{3.47}{4.54} & \ci{4.418}{3.94}{4.87} & \ci{2.49}{2.37}{2.60} & \ci{2.702}{2.61}{2.80} \\
 & Hard & \ci{2.87}{2.47}{3.29} & \ci{3.475}{3.12}{3.83} & \ci{2.23}{2.13}{2.34} & \ci{2.538}{2.45}{2.63} \\
\midrule
Llama-3.2-1B-it & Base & \ci{7.52}{7.01}{7.99} & \ci{3.252}{2.88}{3.64} & \ci{7.35}{7.17}{7.52} & \ci{2.491}{2.38}{2.60} \\
 & Soft & \ci{7.77}{7.32}{8.22} & \ci{3.410}{3.01}{3.80} & \ci{7.97}{7.81}{8.13} & \ci{2.286}{2.19}{2.39} \\
 & Hard & \ci{7.76}{7.29}{8.20} & \ci{3.537}{3.15}{3.95} & \ci{7.66}{7.48}{7.84} & \ci{2.359}{2.25}{2.47} \\
\midrule
Llama-3.2-3B-it & Base & \ci{5.53}{4.97}{6.10} & \ci{4.185}{3.79}{4.57} & \ci{5.27}{5.07}{5.47} & \ci{3.305}{3.18}{3.44} \\
 & Soft & \ci{6.34}{5.78}{6.90} & \ci{4.428}{4.04}{4.81} & \ci{6.18}{6.00}{6.38} & \ci{2.995}{2.86}{3.11} \\
 & Hard & \ci{6.26}{5.71}{6.85} & \ci{4.385}{3.99}{4.78} & \ci{6.15}{5.95}{6.34} & \ci{3.063}{2.94}{3.18} \\
\midrule
GPT-OSS-20B & Base & \ci{6.43}{5.94}{6.92} & \ci{5.232}{4.81}{5.64} & \ci{4.94}{4.78}{5.11} & \ci{3.642}{3.51}{3.77} \\
 & Soft & \ci{7.75}{7.28}{8.22} & \ci{4.180}{3.71}{4.65} & \ci{8.64}{8.51}{8.77} & \ci{1.979}{1.89}{2.08} \\
 & Hard & \ci{7.66}{7.20}{8.13} & \ci{4.371}{3.90}{4.85} & \ci{7.68}{7.52}{7.84} & \ci{2.440}{2.33}{2.55} \\
\bottomrule
\end{tabular}
\caption{\textbf{NoveltyBench split-level summary across all evaluated models.} Distinct is the mean number of semantic partitions per prompt, and Utility is the benchmark's patience-discounted reward-based score, reported separately on the curated and WildChat splits. Base denotes the original checkpoint, Soft the soft-target and Hard the hard-target Calibration Fine-Tuning checkpoint. Intervals are 95\% bootstrap intervals ($B=2000$), shown in smaller type after the point estimate. Intervals resample prompts within each split.}
\label{tab:noveltybench-summary}
\end{table*}

The following tables provide qualitative prompt-level examples from the curated split. We sort prompts by how much their utility changes, to show three kinds of case: cases where soft-target Calibration Fine-Tuning clearly helps, cases where the hard-target variant is competitive or best, and failure cases where Calibration Fine-Tuning hurts reward-weighted utility. For each prompt and method, we display the top five semantic-partition representatives ranked by partition score together with that method's prompt-level Distinct/Utility values.

These examples also illustrate a limitation of the NoveltyBench scoring pipeline. Since semantic partitions and utilities are computed by the benchmark evaluator rather than by human preference, the reported Distinct/Utility values do not always perfectly match qualitative answer quality. For example, in Table~\ref{tab:noveltybench-qualitative-base-gpt_oss_20b-curated-28}, the base model is assigned high utility despite repeatedly concentrating on a small set of answers (mostly Tesla Model 3 and Toyota RAV4 variants), while the calibrated checkpoints produce visibly more distinct recommendations.

\begin{table}[htbp]
\centering
\scriptsize
\setlength{\tabcolsep}{3pt}
\renewcommand{\arraystretch}{1.15}
\caption{\textbf{Representative NoveltyBench cases where soft-target Calibration Fine-Tuning clearly improves reward-weighted utility.} Model: Qwen3-14B (Product \& Purchase Recommendations). Prompt: I am doing online shopping for a new leather wallet. What's the first store/website I should check out? Please just give me one suggestion; I don't need to hear your whole thought process.}
\label{tab:noveltybench-qualitative-soft-qwen3_14b-curated-31}
\begin{tabular}{@{}p{0.32\textwidth}p{0.32\textwidth}p{0.32\textwidth}@{}}
\toprule
Base & Soft & Hard \\
\midrule
\parbox[t]{\linewidth}{D: 3, U: 3.56\newline 1. Leatherman.com} & \cellcolor{green!12} \parbox[t]{\linewidth}{D: 10, U: 8.00\newline 1. Herschel} & \parbox[t]{\linewidth}{D: 6, U: 5.43\newline 1. Zappos} \\
\parbox[t]{\linewidth}{2. Levi's} & \cellcolor{green!12} \parbox[t]{\linewidth}{2. Tumi} & \parbox[t]{\linewidth}{2. M. Gemi} \\
\parbox[t]{\linewidth}{3. Levi's.} & \cellcolor{green!12} \parbox[t]{\linewidth}{3. Ssense} & \parbox[t]{\linewidth}{3. Wynterleeleather.com} \\
\parbox[t]{\linewidth}{4. LeatherMan.com} & \cellcolor{green!12} \parbox[t]{\linewidth}{4. ebay.com} & \parbox[t]{\linewidth}{4. LeatherMan.com} \\
\parbox[t]{\linewidth}{5. LeatherMan.com} & \cellcolor{green!12} \parbox[t]{\linewidth}{5. M. Gemi} & \parbox[t]{\linewidth}{5. Tumi} \\
\bottomrule
\end{tabular}
\end{table}

\begin{table}[htbp]
\centering
\scriptsize
\setlength{\tabcolsep}{3pt}
\renewcommand{\arraystretch}{1.15}
\caption{\textbf{Representative NoveltyBench cases where soft-target Calibration Fine-Tuning clearly improves reward-weighted utility.} Model: Gemma-3-27B-it (Subjective Rankings \& Opinions). Prompt: What is the coolest Pokémon from the second generation? Just give me one.}
\label{tab:noveltybench-qualitative-soft-gemma3_27b_it-curated-54}
\begin{tabular}{@{}p{0.32\textwidth}p{0.32\textwidth}p{0.32\textwidth}@{}}
\toprule
Base & Soft & Hard \\
\midrule
\parbox[t]{\linewidth}{D: 1, U: 1.57\newline 1. Typhlosion. It's a really solid pick! A powerful Fire-type with a cool, mature design - a big, powerful tiger wreathed in flames. It feels like a ...} & \cellcolor{green!12} \parbox[t]{\linewidth}{D: 10, U: 8.78\newline 1. Umbreon. It's just a sleek, mysterious, and powerful Dark-type with glowing markings. Everything about it is awesome.} & \parbox[t]{\linewidth}{D: 2, U: 2.82\newline 1. Tyranitar. It's just massive, imposing, and has a really cool design that feels genuinely threatening. It's a pseudo-legendary that's still ...} \\
\parbox[t]{\linewidth}{2. Typhlosion. It's a fire-type with a really awesome, somewhat melancholic design - a big, powerful tiger wreathed in flames. Plus, its backstory (a ...} & \cellcolor{green!12} \parbox[t]{\linewidth}{2. Typhlosion.} & \parbox[t]{\linewidth}{2. Typhlosion. Its design is sleek and powerful, it's a Fire-type starter with a lot of nostalgia, and it just looks cool breathing those blue flames. ...} \\
\parbox[t]{\linewidth}{3. Typhlosion. It's a powerful Fire-type with a really cool, mature design - a big, strong tiger with flames constantly swirling. It feels like a truly ...} & \cellcolor{green!12} \parbox[t]{\linewidth}{3. Houndoom.} & \parbox[t]{\linewidth}{3. Typhlosion. Its design is sleek and powerful, and it has a cool, fiery aesthetic without being too over-the-top. Plus, it's a powerful Fire-type with ...} \\
\parbox[t]{\linewidth}{4. Typhlosion. It's a fiery, powerful, and surprisingly graceful final evolution with a really cool design. Plus, that fiery mane is just awesome! It ...} & \cellcolor{green!12} \parbox[t]{\linewidth}{4. Tyranitar is the coolest Pokémon from the second generation. It's a massive, imposing pseudo-legendary with a fantastic design and a powerful ...} & \parbox[t]{\linewidth}{4. Typhlosion. Its design is sleek and powerful, it's a Fire-type starter that doesn't just look like a bigger version of itself (looking at you, ...} \\
\parbox[t]{\linewidth}{5. Typhlosion. It's a fully evolved Fire-type with a really cool, almost melancholy design. The idea of a Pokémon constantly emitting flames from its ...} & \cellcolor{green!12} \parbox[t]{\linewidth}{5. Typhlosion} & \parbox[t]{\linewidth}{5. Tyranitar. It's a massive, intimidating pseudo-legendary with a fantastic design and a powerful presence. It just feels cool.} \\
\bottomrule
\end{tabular}
\end{table}

\begin{table}[htbp]
\centering
\scriptsize
\setlength{\tabcolsep}{3pt}
\renewcommand{\arraystretch}{1.15}
\caption{\textbf{Representative NoveltyBench cases where soft-target Calibration Fine-Tuning clearly improves reward-weighted utility.} Model: GPT-OSS-20B (Creativity). Prompt: Write a haiku about a whale and a walnut tree.}
\label{tab:noveltybench-qualitative-soft-gpt_oss_20b-curated-3}
\begin{tabular}{@{}p{0.32\textwidth}p{0.32\textwidth}p{0.32\textwidth}@{}}
\toprule
Base & Soft & Hard \\
\midrule
\parbox[t]{\linewidth}{D: 2, U: 2.93\newline 1. Whale roars the sea- a walnut tree echoes back, roots and waves entwine.} & \cellcolor{green!12} \parbox[t]{\linewidth}{D: 5, U: 5.94\newline 1. Ripples whisper soft, Distance spun by the whale's steps- Nuts scent, wind hears it.} & \parbox[t]{\linewidth}{D: 6, U: 5.07\newline 1. Sea wind whispers- whales and walnut branches sway in sync- blossoms scent the tide.} \\
\parbox[t]{\linewidth}{2. Silent whale glides- Its song weaves through the orchard, Walnut leaves blush, hush.} & \cellcolor{green!12} \parbox[t]{\linewidth}{2. The wind whips low leaves Below, a whale sings in waves Sunlight paints their stories} & \parbox[t]{\linewidth}{2. Waves hush softly- whale's breath stirs the brook below, nuts wait for spring's kiss.} \\
\parbox[t]{\linewidth}{3. Silent whale rises, Wind whispers through walnut bark- Sea and earth converse.} & \cellcolor{green!12} \parbox[t]{\linewidth}{3. Beneath waves gleams an eye, Tree bones grow, rain becomes rust, Silent bond of earth.} & \parbox[t]{\linewidth}{3. Beneath waves gleam, Whale bones drape like sea silk,- Walnut traces fall.} \\
\parbox[t]{\linewidth}{4. Waves hush the shore- a whale drifts beneath the sky, nuts sway in the wind.} & \cellcolor{green!12} \parbox[t]{\linewidth}{4. Waves hush softly Below, roots, palm branches tremble- Sky, sea, nutmaiden.} & \parbox[t]{\linewidth}{4. The sea song swells low Below, a whale bends near. Cedar tree sighs deep.} \\
\parbox[t]{\linewidth}{5. Silent whale glides, under the walnut's green shadow- seas and woods converse.} & \cellcolor{green!12} \parbox[t]{\linewidth}{5. Rippling moonlight Finds the walnut's tall shade- Whale stirs the stream.} & \parbox[t]{\linewidth}{5. Rippling moonlight Finds the shell, a distant glow Swayed by tree, it listens.} \\
\bottomrule
\end{tabular}
\end{table}

\begin{table}[htbp]
\centering
\scriptsize
\setlength{\tabcolsep}{3pt}
\renewcommand{\arraystretch}{1.15}
\caption{\textbf{Representative cases where the hard-target variant is competitive or best on prompt-level NoveltyBench utility.} Model: Qwen3-14B (Factual Knowledge). Prompt: List a capital city in Africa.}
\label{tab:noveltybench-qualitative-hard-qwen3_14b-curated-58}
\begin{tabular}{@{}p{0.32\textwidth}p{0.32\textwidth}p{0.32\textwidth}@{}}
\toprule
Base & Soft & Hard \\
\midrule
\parbox[t]{\linewidth}{D: 4, U: 4.23\newline 1. A capital city in Africa is Abuja, which is the capital of Nigeria.} & \parbox[t]{\linewidth}{D: 4, U: 4.24\newline 1. Abuja} & \cellcolor{green!12} \parbox[t]{\linewidth}{D: 7, U: 6.78\newline 1. Addis Ababa} \\
\parbox[t]{\linewidth}{2. Cairo is a capital city in Africa.} & \parbox[t]{\linewidth}{2. Cairo} & \cellcolor{green!12} \parbox[t]{\linewidth}{2. Cairo is the capital city of Egypt, which is located in Africa.} \\
\parbox[t]{\linewidth}{3. Abuja is the capital city of Nigeria.} & \parbox[t]{\linewidth}{3. Lome is a capital city in Africa.} & \cellcolor{green!12} \parbox[t]{\linewidth}{3. An example of a capital city in Africa is Cairo, the capital of Egypt.} \\
\parbox[t]{\linewidth}{4. A capital city in Africa is Cairo, the capital of Egypt.} & \parbox[t]{\linewidth}{4. Lagos is the capital city of Nigeria.} & \cellcolor{green!12} \parbox[t]{\linewidth}{4. Abuja is the capital city of Nigeria, an African country.} \\
\parbox[t]{\linewidth}{5. A capital city in Africa is Cairo, the capital of Egypt.} & \parbox[t]{\linewidth}{5. The capital of Nigeria is Abuja.} & \cellcolor{green!12} \parbox[t]{\linewidth}{5. Nairobi is a capital city in Africa. It is the capital of Kenya and serves as a major cultural and economic hub on the continent.} \\
\bottomrule
\end{tabular}
\end{table}

\begin{table}[htbp]
\centering
\scriptsize
\setlength{\tabcolsep}{3pt}
\renewcommand{\arraystretch}{1.15}
\caption{\textbf{Representative cases where the hard-target variant is competitive or best on prompt-level NoveltyBench utility.} Model: Gemma-3-27B-it (Creativity). Prompt: One thing you might find in a magical forest.}
\label{tab:noveltybench-qualitative-hard-gemma3_27b_it-curated-16}
\begin{tabular}{@{}p{0.32\textwidth}p{0.32\textwidth}p{0.32\textwidth}@{}}
\toprule
Base & Soft & Hard \\
\midrule
\parbox[t]{\linewidth}{D: 4, U: 4.42\newline 1. A whispering tree! It wouldn't just look ancient and magical, but its leaves would rustle and murmur secrets on the breeze - stories of the forest, ...} & \parbox[t]{\linewidth}{D: 4, U: 4.58\newline 1. Definitely a talking animal. It could be a wise old owl offering cryptic advice, a mischievous squirrel trading riddles for nuts, or a regal stag ...} & \cellcolor{green!12} \parbox[t]{\linewidth}{D: 6, U: 7.40\newline 1. A tree with leaves made of shimmering amethyst crystals. They chime softly in the breeze, and catching the light creates a dazzling, otherworldly ...} \\
\parbox[t]{\linewidth}{2. A talking animal! Perhaps a wise old owl offering cryptic advice, a mischievous squirrel bartering for shiny objects, or a regal stag who is actually ...} & \parbox[t]{\linewidth}{2. Glowing mushrooms. They come in all sizes and colours - sapphire blue, emerald green, ruby red - and illuminate the forest floor with a soft, ...} & \cellcolor{green!12} \parbox[t]{\linewidth}{2. A mushroom ring that, when stepped into, momentarily swaps your senses - you hear colors, see sounds, smell textures, etc. It's disorienting, but ...} \\
\parbox[t]{\linewidth}{3. A mushroom ring that, when stepped into, briefly shows you a possible future. It's classic magical forest, a little mysterious, and opens up ...} & \parbox[t]{\linewidth}{3. A mushroom ring that, if you step inside, temporarily grants you the ability to understand the language of animals. Or maybe...a stream that flows ...} & \cellcolor{green!12} \parbox[t]{\linewidth}{3. A glowing mushroom circle, pulsing with soft, ethereal light and rumored to be a doorway to the realm of the fae. Maybe if you step inside, you'll ...} \\
\parbox[t]{\linewidth}{4. A glowing mushroom ring! It's a classic magical forest element - a circle of fungi emitting a soft, ethereal light, perhaps marking a portal to ...} & \parbox[t]{\linewidth}{4. A mushroom ring that, when stepped into, momentarily swaps your memories with those of the oldest tree in the forest.} & \cellcolor{green!12} \parbox[t]{\linewidth}{4. A whispering tree whose leaves murmur forgotten stories to those who listen closely. Each leaf holds a snippet of a tale - a lost love, a heroic ...} \\
\parbox[t]{\linewidth}{5. A glowing mushroom circle! It's a classic magical forest trope - a ring of mushrooms emitting a soft, ethereal light. Maybe fairies dance within it, ...} & \parbox[t]{\linewidth}{5. A mushroom ring that, when stepped within, briefly shows you a vision of your greatest desire. But be warned - lingering too long steals your ...} & \cellcolor{green!12} \parbox[t]{\linewidth}{5. A tree whose leaves whisper secrets to those who listen closely. The secrets aren't always useful - sometimes it's just the recipe for a forgotten ...} \\
\bottomrule
\end{tabular}
\end{table}

\begin{table}[htbp]
\centering
\scriptsize
\setlength{\tabcolsep}{3pt}
\renewcommand{\arraystretch}{1.15}
\caption{\textbf{Representative cases where the hard-target variant is competitive or best on prompt-level NoveltyBench utility.} Model: GPT-OSS-20B (Factual Knowledge). Prompt: Name one person who won the Fields Medal.}
\label{tab:noveltybench-qualitative-hard-gpt_oss_20b-curated-71}
\begin{tabular}{@{}p{0.32\textwidth}p{0.32\textwidth}p{0.32\textwidth}@{}}
\toprule
Base & Soft & Hard \\
\midrule
\parbox[t]{\linewidth}{D: 6, U: 4.37\newline 1. Alexander Grothendieck.} & \parbox[t]{\linewidth}{D: 6, U: 4.71\newline 1. Sir Michael Atiyah} & \cellcolor{green!12} \parbox[t]{\linewidth}{D: 10, U: 6.61\newline 1. Pierre Deligne} \\
\parbox[t]{\linewidth}{2. One person who has won the Fields Medal is John G. Thompson (awarded in 1974).} & \parbox[t]{\linewidth}{2. Solomon Lefschetz} & \cellcolor{green!12} \parbox[t]{\linewidth}{2. Terence Tao} \\
\parbox[t]{\linewidth}{3. One Fields Medal laureate is Grigori Perelman (1998).} & \parbox[t]{\linewidth}{3. Laurence J. Lafforgue} & \cellcolor{green!12} \parbox[t]{\linewidth}{3. Maryna Viazovska} \\
\parbox[t]{\linewidth}{4. One notable Fields Medalist is Grigori Perelman, who received the award in 2006 for his proof of the Poincar conjecture.} & \parbox[t]{\linewidth}{4. Andrew Wiles} & \cellcolor{green!12} \parbox[t]{\linewidth}{4. Andrew Wiles} \\
\parbox[t]{\linewidth}{5. Ada Levine (2024)} & \parbox[t]{\linewidth}{5. John Milnor} & \cellcolor{green!12} \parbox[t]{\linewidth}{5. Laurent Lafforgue} \\
\bottomrule
\end{tabular}
\end{table}

\begin{table}[htbp]
\centering
\scriptsize
\setlength{\tabcolsep}{3pt}
\renewcommand{\arraystretch}{1.15}
\caption{\textbf{Representative failure cases where Calibration Fine-Tuning hurts prompt-level utility.} Model: GPT-OSS-20B (Product \& Purchase Recommendations). Prompt: Suggest a birthday gift for a friend's 25th birthday. Just one}
\label{tab:noveltybench-qualitative-base-gpt_oss_20b-curated-29}
\begin{tabular}{@{}p{0.32\textwidth}p{0.32\textwidth}p{0.32\textwidth}@{}}
\toprule
Base & Soft & Hard \\
\midrule
\cellcolor{green!12} \parbox[t]{\linewidth}{D: 8, U: 8.40\newline 1. A personalized leather passport holder-engraved with their initials and a short birthday message. It's stylish, practical for future travels, and ...} & \parbox[t]{\linewidth}{D: 10, U: 5.34\newline 1. Luxury leather wallet} & \parbox[t]{\linewidth}{D: 10, U: 3.95\newline 1. A personalized leather journal with a custom engraved name or monogram.} \\
\cellcolor{green!12} \parbox[t]{\linewidth}{2. A handcrafted leather journal with a personalized engraved initial on the cover. It's a timeless, practical gift that encourages creativity, ...} & \parbox[t]{\linewidth}{2. Bluetooth speaker with WiFi connectivity} & \parbox[t]{\linewidth}{2. A personalized leather journal with a custom monogram and a luxurious fountain pen set.} \\
\cellcolor{green!12} \parbox[t]{\linewidth}{3. A highquality leather messenger bag. It's practical, stylish, and gets better with age-perfect for a 25yearold looking to upgrade their everyday ...} & \parbox[t]{\linewidth}{3. Custom portrait in the style of a retro comic book} & \parbox[t]{\linewidth}{3. Portable Bluetooth speaker with a sleek, waterproof design for music on-the-go.} \\
\cellcolor{green!12} \parbox[t]{\linewidth}{4. How about a highquality, personalized leather wallet? It's classy, practical, and the added personal touch makes it special for a 25th birthday.} & \parbox[t]{\linewidth}{4. Personalized leather guitar strap with high-quality engraving} & \parbox[t]{\linewidth}{4. A premium yoga retreat vacation voucher.} \\
\cellcolor{green!12} \parbox[t]{\linewidth}{5. A personalized leather passport holder with their initials and a small map of their favorite travel destinations.} & \parbox[t]{\linewidth}{5. Roku Streaming Stick+} & \parbox[t]{\linewidth}{5. A personalized wood cutting board with a lifetime woodworm seal. It's both functional and stylish for a cooking enthusiast.} \\
\bottomrule
\end{tabular}
\end{table}

\begin{table}[htbp]
\centering
\scriptsize
\setlength{\tabcolsep}{3pt}
\renewcommand{\arraystretch}{1.15}
\caption{\textbf{Representative failure cases where Calibration Fine-Tuning hurts prompt-level utility.} Model: Qwen3-14B (Character \& Entity Naming). Prompt: Suggest a name for a dappled-gray filly living in the mountains.}
\label{tab:noveltybench-qualitative-base-qwen3_14b-curated-23}
\begin{tabular}{@{}p{0.32\textwidth}p{0.32\textwidth}p{0.32\textwidth}@{}}
\toprule
Base & Soft & Hard \\
\midrule
\cellcolor{green!12} \parbox[t]{\linewidth}{D: 9, U: 9.00\newline 1. A dappled-gray filly living in the mountains could have a name that reflects both her color and her wild, elegant nature.} & \parbox[t]{\linewidth}{D: 6, U: 6.02\newline 1. Shadowmist A dappled-gray filly living in the mountains would be cloaked in an ever-changing play of light and shadow. Shadowmist evokes the ...} & \parbox[t]{\linewidth}{D: 3, U: 4.46\newline 1. Liora Mistveil ("Liora" evokes light and silver, while "Mistveil" suggests the ethereal, shifting light of a dappled coat and a mountainous landscape ...} \\
\cellcolor{green!12} \parbox[t]{\linewidth}{2. How about Ashmere? Meaning \& Origin: - Ash refers to the ashen-gray color, which matches the filly's dappled-gray coat. - Mere can mean a small lake ...} & \parbox[t]{\linewidth}{2. Liora Elowen (LIO-rah AH-loh-uhn) Meaning \& Inspiration: - Liora stems from Hebrew, meaning "light that shines" or "my light," nodding to the way ...} & \parbox[t]{\linewidth}{2. Glimmerrest This name evokes the shimmering dapples of her coat and the serene, restful quality of mountain life. It also hints at a mystical or ...} \\
\cellcolor{green!12} \parbox[t]{\linewidth}{3. How about Ashmoor? Meaning \& Inspiration: - Ash refers to the gray or ashen coloration of the filly's coat, and also evokes the imagery of ash trees, ...} & \parbox[t]{\linewidth}{3. Cindermoss This name evokes the soft, dappled appearance of the filly ("cin" suggesting a light or gray tone, like "cin" in "cinereous") and a mossy ...} & \parbox[t]{\linewidth}{3. Shadowspire This name evokes the interplay of light and shadow in the mountains, reflecting the dappled-gray coat of the filly. "Shadow" hints at the ...} \\
\cellcolor{green!12} \parbox[t]{\linewidth}{4. How about Shadowpaw? This name evokes the dappled-gray coat and the mystical, wild nature of a mountain-dwelling filly. "Shadow" reflects the gray ...} & \parbox[t]{\linewidth}{4. Shadowveil This name evokes the impression of a filly whose coat shimmers with a dappled silvery-gray hue, as if the mountain mist and twilight weave ...} & \parbox[t]{\linewidth}{4. Mistspire This name blends the ethereal quality of "mist" with the towering, spired imagery of the mountains, evoking the filly's dappled-gray coat ...} \\
\cellcolor{green!12} \parbox[t]{\linewidth}{5. How about Shadowflake? This name blends the dappled-gray coloring ("shadow") with the delicate, sparkling appearance of the dapples ("flake"), and it ...} & \parbox[t]{\linewidth}{5. Shadowspire This name evokes a mystical, elegant, and mountainous presence. "Shadow" reflects the dappled-gray coat, while "spire" suggests the ...} & \parbox[t]{\linewidth}{5. Ashmere This name evokes the image of a dappled-gray filly with subtle, ghostly markings reminiscent of mist or sunlight filtering through mountain ...} \\
\bottomrule
\end{tabular}
\end{table}

\begin{table}[htbp]
\centering
\scriptsize
\setlength{\tabcolsep}{3pt}
\renewcommand{\arraystretch}{1.15}
\caption{\textbf{Representative failure cases where Calibration Fine-Tuning hurts prompt-level utility.} Model: GPT-OSS-20B (Product \& Purchase Recommendations). Prompt: What's the best car to get in 2023? Just give me one single suggestion. Otherwise I'll have decision paralysis.}
\label{tab:noveltybench-qualitative-base-gpt_oss_20b-curated-28}
\begin{tabular}{@{}p{0.32\textwidth}p{0.32\textwidth}p{0.32\textwidth}@{}}
\toprule
Base & Soft & Hard \\
\midrule
\cellcolor{green!12} \parbox[t]{\linewidth}{D: 9, U: 8.37\newline 1. Toyota RAV4 Prime (plugin hybrid) It gives you the best mix of rangeability, realworld fuel economy, advanced safety tech, and practicality for a ...} & \parbox[t]{\linewidth}{D: 7, U: 4.34\newline 1. Tesla Model 3} & \parbox[t]{\linewidth}{D: 6, U: 2.73\newline 1. Hyundai Ioniq 5.} \\
\cellcolor{green!12} \parbox[t]{\linewidth}{2. 2023 Tesla Model 3 (LongRange) - It delivers unbeatable electric performance, a strong resale value, and highly praised safety and tech features, ...} & \parbox[t]{\linewidth}{2. Toyota Camry} & \parbox[t]{\linewidth}{2. Lexus UX300e (Hybrid)} \\
\cellcolor{green!12} \parbox[t]{\linewidth}{3. 2023 Toyota RAV4 Prime - a plugin hybrid SUV that balances great fuel economy, strong power, clever tech, and earned safety awards for a confident, ...} & \parbox[t]{\linewidth}{3. Honda Civic} & \parbox[t]{\linewidth}{3. Toyota RAV4 Prime} \\
\cellcolor{green!12} \parbox[t]{\linewidth}{4. Toyota RAV4 Prime - a plugin hybrid SUV that delivers great fuel economy, instant electric torque, a solid driving range, and the reliability you ...} & \parbox[t]{\linewidth}{4. MercedesBenz GLE 2023} & \parbox[t]{\linewidth}{4. 2023 Honda Escape Hybrid Convertible} \\
\cellcolor{green!12} \parbox[t]{\linewidth}{5. 2023 Tesla Model 3 - a highly capable, wellbalanced electric sedan that offers excellent range, performance, and tech.} & \parbox[t]{\linewidth}{5. Toyota RAV4 TRD Pro} & \parbox[t]{\linewidth}{5. Chevrolet Ford Explorer](} \\
\bottomrule
\end{tabular}
\end{table}

\clearpage
\subsection{MCQ Answer-Position Balance}
\label{sec:appendix-mcq}

Table~\ref{tab:mcq-summary} reports the full MCQ answer-position results under the prompt in Appendix~\ref{app:mcq-prompt}, following the protocol of \citet{zhao2026dice}. This evaluation differs from structured sampling because the target distribution is only implicit: the prompt asks the model to generate valid medical multiple-choice questions while distributing the correct answer approximately uniformly across A/B/C/D. We therefore report both MCQ parse rate, the fraction of generations that can be parsed as MCQs with a question, four A/B/C/D options, and a correct-answer field in A/B/C/D, and TV distance from the uniform answer-position distribution, computed both over parseable generations and over all attempts, with each unparseable generation counted as a fifth outcome with zero target mass (joint TV).

Over parseable generations, balance improves on most models, significantly for four under soft-target and two under hard-target fine-tuning, and worsens only for Llama-3.2-3B-it. Soft-target fine-tuning also lowers the parse rate on seven models, sharply for GPT-OSS-20B (100\% to 21\%) and the smallest Qwen and Llama checkpoints, while hard-target fine-tuning lowers it on three and raises it on three. Joint TV combines both effects: it improves where parse rates stay high, as for Qwen3-4B and Qwen3-14B. Overall, the balance gain is partly offset by a format cost.

\begin{table*}[t]
\centering
\footnotesize
\begin{tabular}{llrrr}
\toprule
Model & & Parse rate (\%) $\uparrow$ & TV (parseable) $\downarrow$ & Joint TV $\downarrow$ \\
\midrule
Qwen3-0.6B & Base & \ci{47.6}{44.4}{50.9} & \ci{0.374}{0.33}{0.42} & \ci{0.571}{0.55}{0.59} \\
 & Soft & \ci{0.0}{0.0}{0.0} & -- & \ci{1.000}{1.00}{1.00} \\
 & Hard & \ci{12.8}{10.8}{14.9} & \ci{0.398}{0.33}{0.48} & \ci{0.872}{0.85}{0.89} \\
\midrule
Qwen3-1.7B & Base & \ci{84.7}{82.3}{87.0} & \ci{0.313}{0.28}{0.35} & \ci{0.380}{0.35}{0.41} \\
 & Soft & \ci{98.7}{98.0}{99.3} & \ci{0.249}{0.22}{0.28} & \ci{0.256}{0.23}{0.29} \\
 & Hard & \ci{99.8}{99.5}{100.0} & \ci{0.288}{0.26}{0.32} & \ci{0.289}{0.26}{0.32} \\
\midrule
Qwen3-4B & Base & \ci{99.9}{99.7}{100.0} & \ci{0.192}{0.16}{0.22} & \ci{0.192}{0.16}{0.22} \\
 & Soft & \ci{100.0}{100.0}{100.0} & \ci{0.116}{0.10}{0.14} & \ci{0.116}{0.10}{0.14} \\
 & Hard & \ci{100.0}{100.0}{100.0} & \ci{0.156}{0.13}{0.19} & \ci{0.156}{0.13}{0.19} \\
\midrule
Qwen3-8B & Base & \ci{99.9}{99.7}{100.0} & \ci{0.240}{0.21}{0.27} & \ci{0.240}{0.21}{0.27} \\
 & Soft & \ci{97.9}{97.0}{98.7} & \ci{0.240}{0.21}{0.27} & \ci{0.245}{0.22}{0.27} \\
 & Hard & \ci{99.9}{99.7}{100.0} & \ci{0.221}{0.19}{0.25} & \ci{0.221}{0.19}{0.25} \\
\midrule
Qwen3-14B & Base & \ci{100.0}{100.0}{100.0} & \ci{0.284}{0.26}{0.31} & \ci{0.284}{0.26}{0.31} \\
 & Soft & \ci{100.0}{100.0}{100.0} & \ci{0.184}{0.16}{0.21} & \ci{0.184}{0.16}{0.21} \\
 & Hard & \ci{100.0}{100.0}{100.0} & \ci{0.190}{0.16}{0.22} & \ci{0.190}{0.16}{0.22} \\
\midrule
Gemma-3-1B-it & Base & \ci{19.7}{17.3}{22.3} & \ci{0.209}{0.18}{0.24} & \ci{0.803}{0.78}{0.83} \\
 & Soft & \ci{61.6}{58.4}{64.3} & \ci{0.205}{0.19}{0.22} & \ci{0.384}{0.36}{0.42} \\
 & Hard & \ci{94.0}{92.6}{95.5} & \ci{0.203}{0.19}{0.22} & \ci{0.212}{0.20}{0.24} \\
\midrule
Gemma-3-4B-it & Base & \ci{99.9}{99.7}{100.0} & \ci{0.240}{0.21}{0.27} & \ci{0.240}{0.21}{0.27} \\
 & Soft & \ci{76.7}{74.0}{79.2} & \ci{0.178}{0.16}{0.21} & \ci{0.262}{0.24}{0.29} \\
 & Hard & \ci{99.8}{99.5}{100.0} & \ci{0.256}{0.23}{0.28} & \ci{0.256}{0.23}{0.28} \\
\midrule
Gemma-3-12B-it & Base & \ci{100.0}{100.0}{100.0} & \ci{0.335}{0.31}{0.36} & \ci{0.335}{0.31}{0.36} \\
 & Soft & \ci{97.7}{96.7}{98.6} & \ci{0.289}{0.26}{0.31} & \ci{0.294}{0.27}{0.32} \\
 & Hard & \ci{100.0}{100.0}{100.0} & \ci{0.313}{0.29}{0.34} & \ci{0.313}{0.29}{0.34} \\
\midrule
Gemma-3-27B-it & Base & \ci{100.0}{100.0}{100.0} & \ci{0.284}{0.26}{0.31} & \ci{0.284}{0.26}{0.31} \\
 & Soft & \ci{64.3}{61.3}{67.3} & \ci{0.315}{0.29}{0.34} & \ci{0.381}{0.36}{0.41} \\
 & Hard & \ci{100.0}{100.0}{100.0} & \ci{0.238}{0.21}{0.27} & \ci{0.238}{0.21}{0.27} \\
\midrule
Llama-3.2-1B-it & Base & \ci{13.8}{11.6}{16.0} & \ci{0.159}{0.09}{0.24} & \ci{0.862}{0.84}{0.88} \\
 & Soft & \ci{0.0}{0.0}{0.0} & -- & \ci{1.000}{1.00}{1.00} \\
 & Hard & \ci{3.9}{2.7}{5.1} & \ci{0.199}{0.14}{0.34} & \ci{0.961}{0.95}{0.97} \\
\midrule
Llama-3.2-3B-it & Base & \ci{95.0}{93.6}{96.4} & \ci{0.066}{0.04}{0.10} & \ci{0.100}{0.07}{0.13} \\
 & Soft & \ci{99.8}{99.5}{100.0} & \ci{0.187}{0.16}{0.22} & \ci{0.188}{0.16}{0.22} \\
 & Hard & \ci{99.2}{98.6}{99.7} & \ci{0.189}{0.16}{0.22} & \ci{0.191}{0.16}{0.22} \\
\midrule
GPT-OSS-20B & Base & \ci{100.0}{100.0}{100.0} & \ci{0.166}{0.14}{0.19} & \ci{0.166}{0.14}{0.19} \\
 & Soft & \ci{21.2}{18.7}{23.7} & \ci{0.061}{0.02}{0.13} & \ci{0.788}{0.76}{0.81} \\
 & Hard & \ci{35.1}{32.1}{38.0} & \ci{0.072}{0.04}{0.12} & \ci{0.649}{0.62}{0.68} \\
\bottomrule
\end{tabular}
\caption{\textbf{Multiple-choice generation results across all evaluated models.} Parse rate is the fraction of 1{,}000 generations that parse as MCQs with a question, four A/B/C/D options, and a correct-answer field in A/B/C/D. TV (parseable) is the total variation distance between answer-position frequencies and the uniform distribution over parseable generations, omitted when none parses. Joint TV is computed over all attempts, counting each unparseable generation as a fifth outcome with zero target mass. Intervals are 95\% bootstrap intervals ($B=2000$), shown in smaller type after the point estimate. Intervals resample generations.}
\label{tab:mcq-summary}
\end{table*}

\clearpage
\subsection{Capability Retention}
\label{sec:appendix-capability-retention}

\begin{figure}[htbp]
\centering
\includegraphics[width=\linewidth]{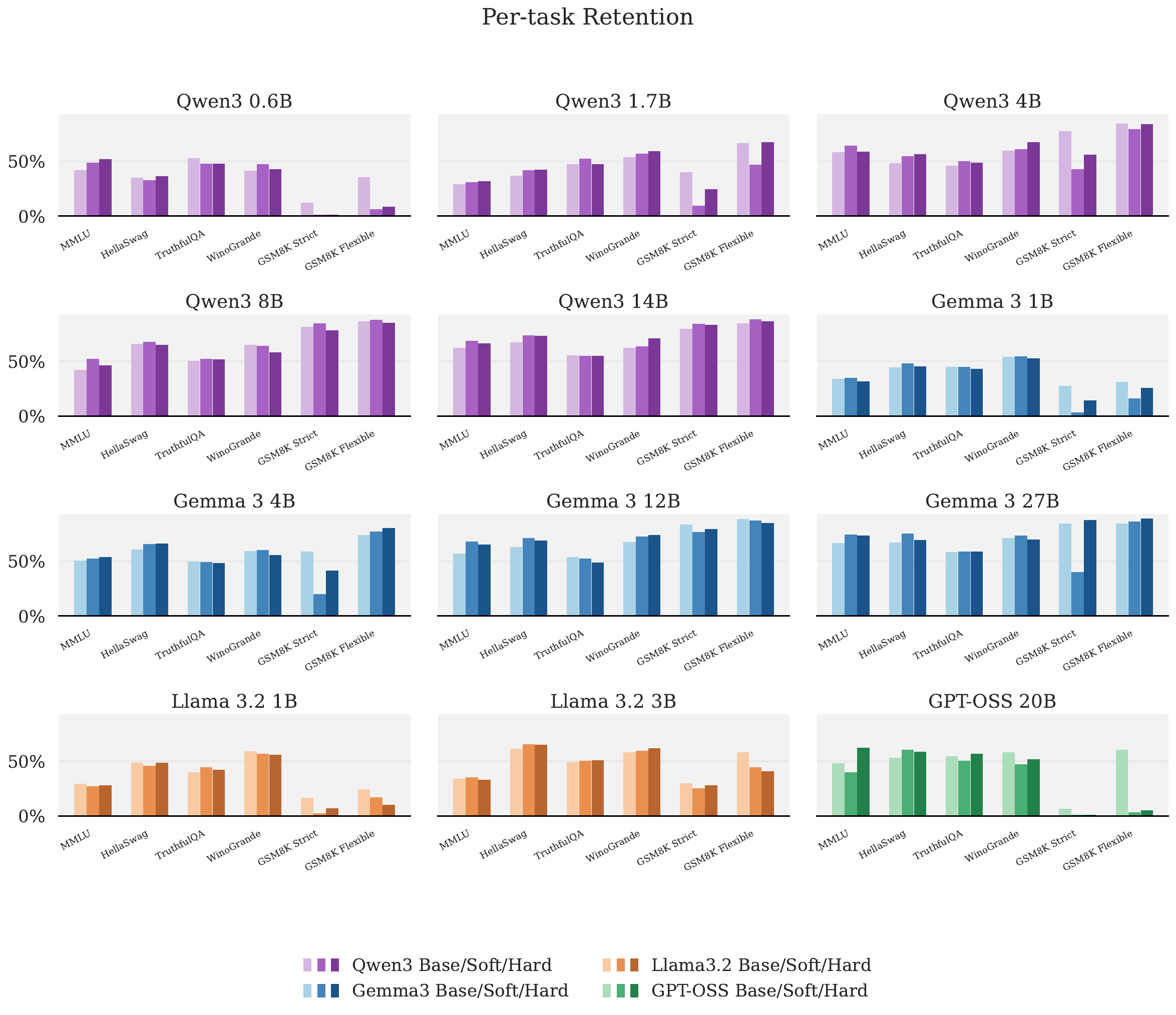}
\caption{\textbf{Per-task retention gp-IRT for base, soft-target, and hard-target models across the TinyBenchmarks suite.} This figure breaks the aggregate retention summary down by benchmark to show where Calibration Fine-Tuning preserves or changes capability.}
\label{fig:appendix-retention-gpirt-per-task}
\end{figure}

Table~\ref{tab:retention-summary} and Figure~\ref{fig:appendix-retention-gpirt-per-task} report the full TinyBenchmarks retention breakdown \citep{polo2024tinybenchmarksevaluatingllmsfewer}. TinyBenchmarks is useful in this setting because it estimates broad downstream capability from small 100-example benchmark slices, allowing us to compare many fine-tuned checkpoints under a fixed greedy-decoding protocol. The aggregate gp-IRT results are model-dependent rather than uniformly preserved: the base checkpoint remains strongest for many smaller models and for both Llama checkpoints, while Calibration Fine-Tuning improves aggregate retention for several medium and large Qwen and Gemma checkpoints.

The task-level view (Table~\ref{tab:retention-per-task}) clarifies this mixed picture. MMLU, HellaSwag, and WinoGrande improve on average after fine-tuning, and TruthfulQA remains nearly unchanged, suggesting that Calibration Fine-Tuning does not simply degrade downstream behavior uniformly. The clearest systematic cost is arithmetic reasoning: both strict and flexible GSM8K gp-IRT decrease under both variants. Overall, these results suggest that Calibration Fine-Tuning can improve stochastic generation behavior while preserving part of the original capability profile, but retention-aware variants are needed if calibration gains must be obtained without arithmetic degradation.

\begin{table*}[t]
\centering
\begin{minipage}[t]{0.64\textwidth}
\centering
\footnotesize
\renewcommand{\cifont}{\scriptsize}
\setlength{\tabcolsep}{2.5pt}
\begin{tabular}[t]{l rrr}
\toprule
Model & Base & Soft & Hard \\
\midrule
Qwen3-0.6B & \ci{0.360}{0.33}{0.40} & \ci{0.302}{0.27}{0.33} & \ci{0.310}{0.28}{0.34} \\
Qwen3-1.7B & \ci{0.451}{0.41}{0.49} & \ci{0.394}{0.36}{0.43} & \ci{0.450}{0.41}{0.49} \\
Qwen3-4B & \ci{0.619}{0.58}{0.66} & \ci{0.584}{0.55}{0.62} & \ci{0.616}{0.58}{0.65} \\
Qwen3-8B & \ci{0.647}{0.61}{0.68} & \ci{0.676}{0.64}{0.71} & \ci{0.637}{0.60}{0.67} \\
Qwen3-14B & \ci{0.682}{0.65}{0.72} & \ci{0.719}{0.68}{0.75} & \ci{0.722}{0.68}{0.75} \\
Gemma-3-1B-it & \ci{0.389}{0.35}{0.43} & \ci{0.330}{0.30}{0.37} & \ci{0.349}{0.32}{0.39} \\
Gemma-3-4B-it & \ci{0.584}{0.54}{0.62} & \ci{0.537}{0.50}{0.57} & \ci{0.572}{0.54}{0.61} \\
Gemma-3-12B-it & \ci{0.684}{0.65}{0.72} & \ci{0.709}{0.67}{0.74} & \ci{0.698}{0.66}{0.73} \\
Gemma-3-27B-it & \ci{0.716}{0.67}{0.74} & \ci{0.677}{0.64}{0.71} & \ci{0.743}{0.70}{0.77} \\
Llama-3.2-1B-it & \ci{0.358}{0.33}{0.40} & \ci{0.318}{0.29}{0.35} & \ci{0.316}{0.29}{0.35} \\
Llama-3.2-3B-it & \ci{0.480}{0.44}{0.52} & \ci{0.463}{0.43}{0.50} & \ci{0.461}{0.42}{0.50} \\
GPT-OSS-20B & \ci{0.463}{0.43}{0.50} & \ci{0.332}{0.30}{0.36} & \ci{0.389}{0.36}{0.42} \\
\bottomrule
\end{tabular}
\captionof{table}{\textbf{TinyBenchmarks aggregate gp-IRT per model}, averaged across the six tasks. Higher is better. Intervals are 95\% bootstrap intervals ($B=2000$), shown in smaller type after the point estimate. Intervals resample the 100 evaluated items of each task and re-estimate gp-IRT on each resample.}
\label{tab:retention-summary}
\end{minipage}
\hfill
\begin{minipage}[t]{0.33\textwidth}
\centering
\scriptsize
\renewcommand{\cifont}{\scriptsize}
\setlength{\tabcolsep}{2.5pt}
\begin{tabular}[t]{l rrr}
\toprule
Task & Base & Soft & Hard \\
\midrule
MMLU & 0.456 & 0.493 & 0.499 \\
HellaSwag & 0.538 & 0.582 & 0.576 \\
TruthfulQA & 0.498 & 0.503 & 0.495 \\
WinoGrande & 0.587 & 0.593 & 0.596 \\
GSM8K Strict & 0.494 & 0.321 & 0.414 \\
GSM8K Flex & 0.645 & 0.529 & 0.551 \\
\bottomrule
\end{tabular}
\captionof{table}{\textbf{gp-IRT per task}, averaged across all 12 models. Higher is better.}
\label{tab:retention-per-task}
\end{minipage}
\end{table*}

\clearpage
\subsection{PALOMA}
\label{sec:appendix-paloma}

To measure perplexity on text from diverse domains, we run teacher-forced next-token evaluation on seven PALOMA test slices: WikiText-103, C4, Dolma, mC4-English, Penn Treebank, RedPajama, and Falcon RefinedWeb \citep{magnusson2024paloma}. We score fixed 2{,}048-token windows with stride 1{,}024 using the base model tokenizer and aggregate token-level negative log-likelihood across all slices. From these totals we report both perplexity and bits-per-byte, with the latter providing a tokenizer-robust comparison across model families.

\begin{figure}[htbp]
\centering
\includegraphics[width=\linewidth]{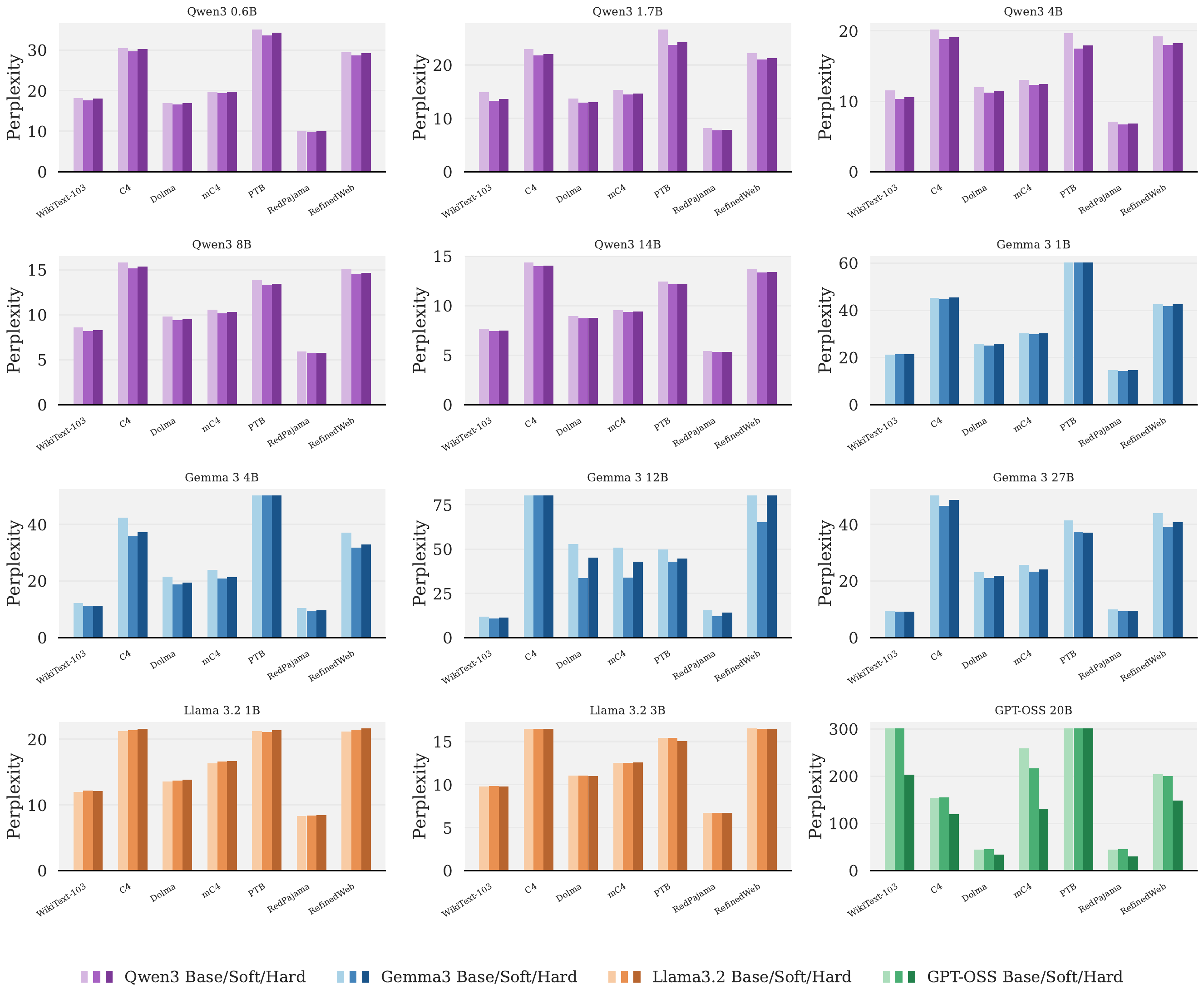}
\caption{\textbf{PALOMA perplexity by evaluation slice for the base, soft-target, and hard-target models.} Lower is better. The figure shows that perplexity is often preserved or improved after Calibration Fine-Tuning, with the strongest variant depending on model family and slice.}
\label{fig:appendix-paloma-perplexity}
\end{figure}

\begin{figure}[htbp]
\centering
\includegraphics[width=\linewidth]{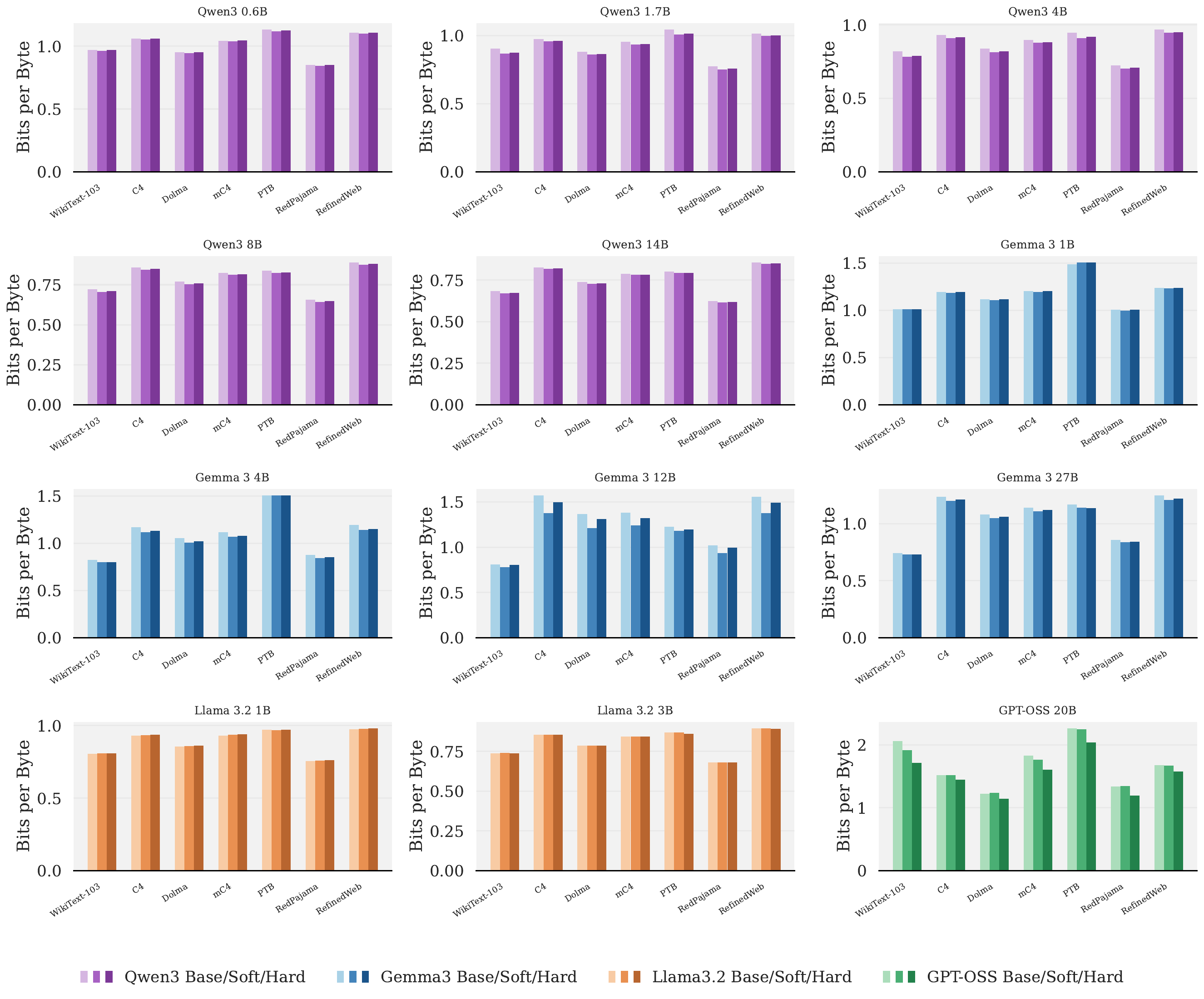}
\caption{\textbf{PALOMA bits-per-byte by evaluation slice for the base, soft-target, and hard-target models.} Lower is better. This tokenizer-robust view follows the same broad pattern as perplexity: fine-tuning often improves held-out text likelihood rather than uniformly degrading it.}
\label{fig:appendix-paloma-bpb}
\end{figure}

Tables~\ref{tab:paloma-model-summary} and~\ref{tab:paloma-slice-summary} report PALOMA language-modeling retention \citep{magnusson2024paloma} by model and by evaluation slice, while Figures~\ref{fig:appendix-paloma-perplexity} and~\ref{fig:appendix-paloma-bpb} show the full per-slice breakdown for each checkpoint. We report both perplexity and bits-per-byte: perplexity measures next-token fit under each model tokenizer, while bits-per-byte gives a more tokenizer-robust view across model families.

The model-level results are more favorable than the TinyBenchmarks retention results. At least one fine-tuned variant preserves or improves aggregate PALOMA for 11 of the 12 checkpoints, often with nontrivial perplexity reductions. The only clear aggregate degradation is Llama-3.2-1B-it, where both variants slightly worsen PALOMA. Calibration Fine-Tuning therefore does not generally degrade perplexity, and in many cases it lowers it.

The per-slice averages show the same trend. For every PALOMA slice, at least one fine-tuned variant improves over the base model in both perplexity and bits-per-byte. Thus, the PALOMA gains are not driven by a single corpus, although their magnitude varies substantially by model family and slice.

\begin{table*}[t]
\centering
\footnotesize
\begin{tabular}{llrr}
\toprule
Model & & Perplexity $\downarrow$ & Bits per byte $\downarrow$ \\
\midrule
Qwen3-0.6B & Base & \ci{19.73}{13.61}{27.86} & \ci{0.998}{0.899}{1.079} \\
 & Soft & \ci{19.27}{13.19}{26.64} & \ci{0.990}{0.894}{1.067} \\
 & Hard & \ci{19.64}{13.18}{27.61} & \ci{0.997}{0.895}{1.076} \\
\midrule
Qwen3-1.7B & Base & \ci{15.50}{10.74}{21.17} & \ci{0.918}{0.823}{0.989} \\
 & Soft & \ci{14.60}{10.62}{19.86} & \ci{0.898}{0.812}{0.972} \\
 & Hard & \ci{14.78}{10.23}{20.23} & \ci{0.902}{0.807}{0.974} \\
\midrule
Qwen3-4B & Base & \ci{13.40}{9.21}{18.10} & \ci{0.869}{0.773}{0.942} \\
 & Soft & \ci{12.52}{8.76}{17.11} & \ci{0.846}{0.754}{0.920} \\
 & Hard & \ci{12.71}{8.91}{17.27} & \ci{0.851}{0.756}{0.927} \\
\midrule
Qwen3-8B & Base & \ci{10.77}{7.63}{14.61} & \ci{0.796}{0.706}{0.869} \\
 & Soft & \ci{10.33}{7.31}{13.80} & \ci{0.782}{0.692}{0.852} \\
 & Hard & \ci{10.46}{7.42}{13.99} & \ci{0.786}{0.694}{0.856} \\
\midrule
Qwen3-14B & Base & \ci{9.77}{6.83}{13.01} & \ci{0.763}{0.668}{0.831} \\
 & Soft & \ci{9.54}{7.00}{12.57} & \ci{0.755}{0.674}{0.823} \\
 & Hard & \ci{9.58}{6.77}{12.73} & \ci{0.757}{0.666}{0.823} \\
\midrule
Gemma-3-1B-it & Base & \ci{29.48}{19.95}{41.28} & \ci{1.146}{1.053}{1.225} \\
 & Soft & \ci{28.92}{19.66}{41.53} & \ci{1.140}{1.048}{1.225} \\
 & Hard & \ci{29.54}{19.84}{41.90} & \ci{1.147}{1.057}{1.228} \\
\midrule
Gemma-3-4B-it & Base & \ci{24.68}{14.69}{39.11} & \ci{1.086}{0.944}{1.202} \\
 & Soft & \ci{21.51}{13.27}{33.39} & \ci{1.039}{0.915}{1.154} \\
 & Hard & \ci{22.16}{13.58}{34.67} & \ci{1.050}{0.920}{1.161} \\
\midrule
Gemma-3-12B-it & Base & \ci{56.56}{26.52}{114.6} & \ci{1.367}{1.137}{1.569} \\
 & Soft & \ci{36.34}{18.41}{63.60} & \ci{1.217}{1.032}{1.365} \\
 & Hard & \ci{47.99}{21.94}{93.75} & \ci{1.311}{1.079}{1.504} \\
\midrule
Gemma-3-27B-it & Base & \ci{25.87}{14.11}{42.00} & \ci{1.102}{0.935}{1.233} \\
 & Soft & \ci{23.53}{13.09}{38.12} & \ci{1.070}{0.909}{1.192} \\
 & Hard & \ci{24.26}{13.29}{40.09} & \ci{1.080}{0.905}{1.210} \\
\midrule
Llama-3.2-1B-it & Base & \ci{15.15}{10.88}{19.77} & \ci{0.884}{0.799}{0.948} \\
 & Soft & \ci{15.32}{11.26}{20.18} & \ci{0.888}{0.810}{0.951} \\
 & Hard & \ci{15.45}{11.10}{20.25} & \ci{0.890}{0.808}{0.954} \\
\midrule
Llama-3.2-3B-it & Base & \ci{12.04}{8.87}{15.32} & \ci{0.809}{0.733}{0.869} \\
 & Soft & \ci{12.04}{8.74}{15.52} & \ci{0.809}{0.726}{0.869} \\
 & Hard & \ci{12.00}{8.91}{15.53} & \ci{0.808}{0.732}{0.871} \\
\midrule
GPT-OSS-20B & Base & \ci{103.7}{47.18}{274.3} & \ci{1.491}{1.263}{1.789} \\
 & Soft & \ci{99.95}{47.78}{229.8} & \ci{1.479}{1.265}{1.734} \\
 & Hard & \ci{70.17}{33.24}{155.9} & \ci{1.366}{1.161}{1.599} \\
\bottomrule
\end{tabular}
\caption{\textbf{PALOMA aggregate results by model, over the seven PALOMA evaluation slices.} Lower is better for both metrics. Base denotes the original checkpoint, Soft the soft-target and Hard the hard-target Calibration Fine-Tuning checkpoint. Intervals are 95\% bootstrap intervals ($B=2000$), shown in smaller type after the point estimate. Intervals resample slices, then files within slice.}
\label{tab:paloma-model-summary}
\end{table*}

\begin{table*}[htbp]
\centering
\scriptsize
\begin{minipage}[t]{0.48\textwidth}
\centering
\resizebox{0.97\linewidth}{!}{%
\begin{tabular}{lccc}
\toprule
Slice & Base & Soft & Hard \\
\midrule
WikiText-103 & 60.63 & 42.31 & \textbf{27.84} \\
C4 & 48.74 & \textbf{41.62} & 42.36 \\
Dolma & 20.95 & \textbf{18.77} & 19.05 \\
mC4 & 40.38 & 34.77 & \textbf{28.63} \\
PTB & 171.22 & 164.07 & \textbf{111.6} \\
RedPajama & 12.07 & 11.62 & \textbf{10.56} \\
RefinedWeb & 47.87 & 42.46 & \textbf{40.78} \\
\bottomrule
\end{tabular}
}
\vspace{3pt}
\caption*{(a) PALOMA Perplexity $\downarrow$}
\end{minipage}
\hfill
\begin{minipage}[t]{0.48\textwidth}
\centering
\resizebox{0.97\linewidth}{!}{%
\begin{tabular}{lccc}
\toprule
Slice & Base & Soft & Hard \\
\midrule
WikiText-103 & 0.92 & 0.893 & \textbf{0.881} \\
C4 & 1.089 & \textbf{1.06} & 1.068 \\
Dolma & 0.968 & \textbf{0.942} & 0.947 \\
mC4 & 1.075 & 1.046 & \textbf{1.043} \\
PTB & 1.215 & 1.199 & \textbf{1.183} \\
RedPajama & 0.843 & 0.825 & \textbf{0.821} \\
RefinedWeb & 1.13 & \textbf{1.101} & 1.107 \\
\bottomrule
\end{tabular}
}
\vspace{3pt}
\caption*{(b) PALOMA Bits per Byte $\downarrow$}
\end{minipage}
\captionsetup{skip=5pt}
\caption{\textbf{PALOMA slice-level summary averaged across all evaluated models.} Each row corresponds to one PALOMA evaluation slice, and values are the mean across the 12 models for each condition. Lower is better for both metrics.}
\label{tab:paloma-slice-summary}
\end{table*}

\clearpage
\section{Hyperparameter Ablations}
\label{sec:appendix-ablations}
\suppressfloats[t]  %

This appendix reports the ablations used to choose the output-space construction and training budgets for Calibration Fine-Tuning. We run these sweeps on two representative large checkpoints, Qwen3-14B and Gemma-3-12B-it. They use separate training runs and are evaluated on a subset of the held-out families (Bernoulli, Poisson, Maxwell, TruncNorm, Chi, and Weibull), so their values differ from the main tables. All ablation results are evaluated against the same high-resolution reference distribution used in the main structured-sampling evaluation, with five-decimal canonical outputs and max bins \(=16384\). We therefore compare training configurations under a fixed evaluation target rather than changing the metric with the training discretization.

Table~\ref{tab:grid-granularity} first shows the construction cost of the discretized output space and prefix trie. Increasing the decimal precision mainly increases output-space memory, while increasing max bins mainly increases trie construction time and trie memory. This makes the discretization choice a practical tradeoff: finer output spaces provide a more faithful canonical approximation, but very high precision quickly becomes expensive before training even begins.

\begin{figure*}[htbp]
\centering
\includegraphics[width=0.9\textwidth]{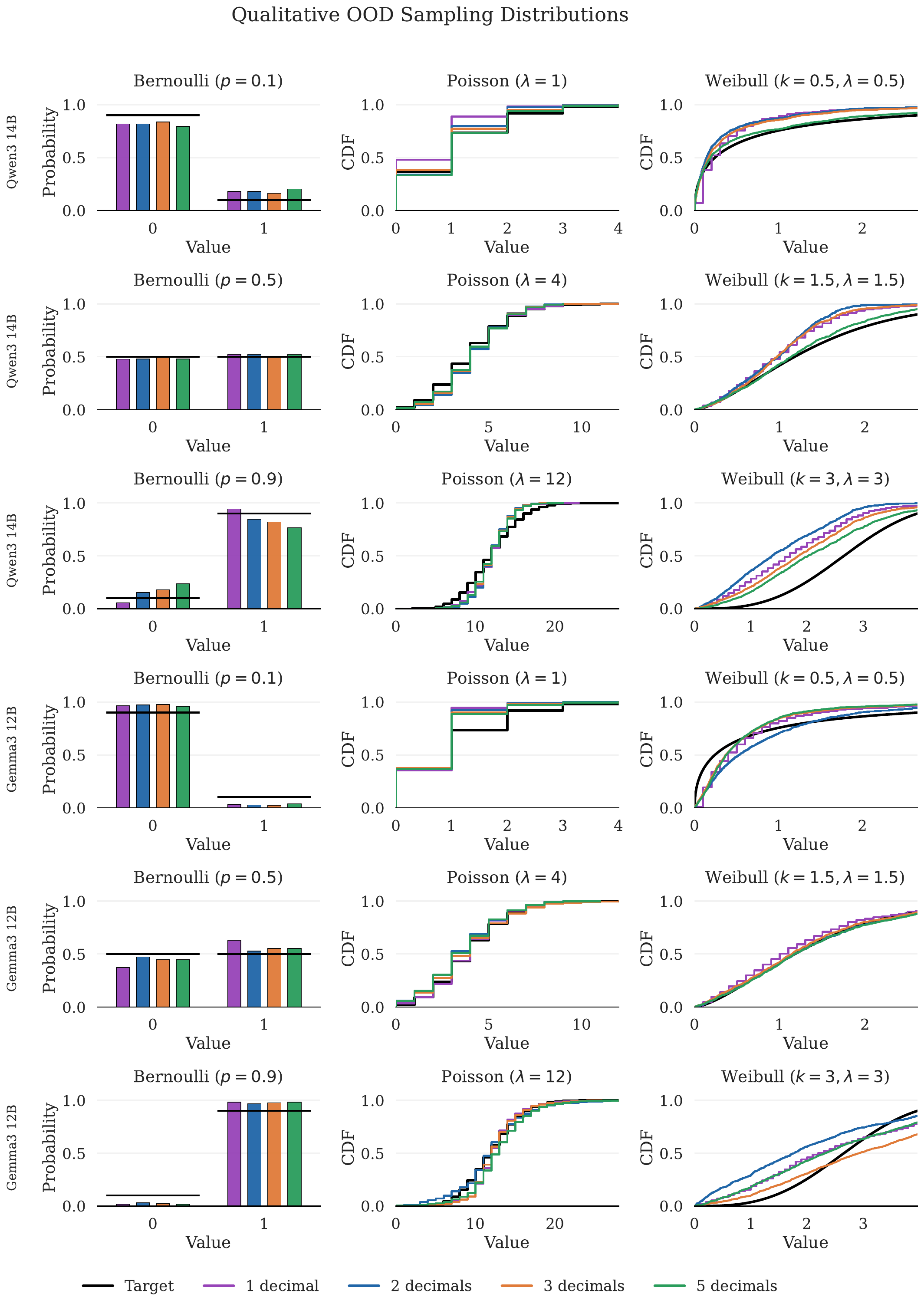}
\caption{\textbf{Qualitative OOD sampling distributions for the output-discretization ablation.} Colors denote the discretization precision used during soft-target Calibration Fine-Tuning. Finer discretizations change behavior in a model-dependent way, consistent with the quantitative ablation tables.}
\label{fig:appendix-num-decimals-qualitative}
\end{figure*}

Table~\ref{tab:grid-ablation} reports the corresponding soft-target ablation. The selected setting, \(d=5\) with max bins \(=1001\), is not chosen because finer or larger output spaces are monotonically better. Instead, it gives the strongest or near-strongest held-out-family performance for both ablated models while keeping the soft-target trie compact. Figure~\ref{fig:appendix-num-decimals-qualitative} gives the same conclusion qualitatively: changing decimal precision affects the sampled distribution, but the effect is model-dependent and smaller than the overall gap between calibrated and base behavior.

Table~\ref{tab:hard-label-ablation} reports the hard-target ablation. Because hard-target supervision observes only sampled completions rather than dense trie-derived token targets, we vary both the output-space construction and the sparse-supervision budget through samples per prompt and epochs. The final setting, \(d=5\), max bins \(=16384\), 16 samples per prompt per epoch, and 2 epochs, keeps the total sampled draws at 32 per prompt and gives a stable compromise between held-out-family and unseen-parameter performance. Other settings are occasionally best on a single model or split, but the selected configuration is the most consistent across the two ablated large-model checkpoints.

\begin{table*}[htbp]
\centering
\scriptsize
\resizebox{0.98\textwidth}{!}{%
\begin{tabular}{rrrrrr}
\toprule
$d$ & max bins & \shortstack{mean\\output-space time (s)} & \shortstack{mean\\trie time (s)} & \shortstack{worst\\output-space peak (MB)} & \shortstack{worst\\trie peak (MB)} \\
\midrule
2 & 1001 & 0.007 & 0.073 & 2.5 & 2.6 \\
2 & 2048 & 0.010 & 0.131 & 2.5 & 5.1 \\
2 & 4096 & 0.014 & 0.178 & 2.5 & 9.8 \\
2 & 8192 & 0.019 & 0.270 & 2.6 & 17.3 \\
2 & 16384 & 0.031 & 0.431 & 3.2 & 27.5 \\
3 & 1001 & 0.009 & 0.118 & 24.3 & 3.3 \\
3 & 2048 & 0.015 & 0.205 & 24.3 & 6.1 \\
3 & 4096 & 0.026 & 0.384 & 24.4 & 11.9 \\
3 & 8192 & 0.042 & 0.641 & 24.4 & 21.5 \\
3 & 16384 & 0.070 & 1.077 & 24.5 & 40.7 \\
4 & 1001 & 0.012 & 0.136 & 242.9 & 4.0 \\
4 & 2048 & 0.019 & 0.263 & 242.9 & 8.0 \\
4 & 4096 & 0.033 & 0.505 & 242.9 & 16.5 \\
4 & 8192 & 0.062 & 1.014 & 243.0 & 26.0 \\
4 & 16384 & 0.112 & 1.835 & 243.1 & 49.8 \\
5 & 1001 & 0.041 & 0.156 & 2428.5 & 4.4 \\
5 & 2048 & 0.047 & 0.306 & 2428.5 & 9.4 \\
5 & 4096 & 0.062 & 0.607 & 2428.6 & 18.5 \\
5 & 8192 & 0.091 & 1.191 & 2428.6 & 35.2 \\
\rowcolor{green!12}
5 & 16384 & 0.150 & 2.303 & 2428.8 & 69.6 \\
6 & 1001 & 0.321 & 0.189 & 24285.1 & 4.9 \\
6 & 2048 & 0.328 & 0.350 & 24285.1 & 10.1 \\
6 & 4096 & 0.342 & 0.698 & 24285.1 & 20.8 \\
6 & 8192 & 0.373 & 1.377 & 24285.2 & 39.3 \\
6 & 16384 & 0.432 & 2.689 & 24285.3 & 76.7 \\
\bottomrule
\end{tabular}
}
\caption{\textbf{Construction cost for the discretization-precision grid profiled in the granularity sweep.} We report mean output-space and trie construction times across the prompt suite, together with the worst per-prompt tracemalloc peak for output-space construction and trie construction. The green row marks the shared high-resolution evaluation reference selected for the ablations.}
\label{tab:grid-granularity}
\end{table*}

\begin{table*}[htbp]
\centering
\begin{minipage}[t]{0.36\textwidth}
\centering
\scriptsize
\resizebox{\linewidth}{!}{%
\begin{tabular}{cccc}
\toprule
$d$ & max bins & \shortstack{Held-out\\$W_1 \downarrow$} & \shortstack{Unseen\\$W_1 \downarrow$} \\
\midrule
\multicolumn{4}{c}{Qwen3-14B} \\
\midrule
2 & 1001 & 0.119 & 0.0936 \\
2 & 4096 & 0.1183 & 0.0887 \\
2 & 16384 & 0.1193 & 0.0817 \\
3 & 1001 & 0.0975 & 0.0792 \\
3 & 4096 & 0.0999 & 0.0881 \\
3 & 16384 & 0.1008 & 0.0813 \\
\rowcolor{green!12}
5 & 1001 & \textbf{0.0847} & \textbf{0.074} \\
5 & 4096 & 0.0927 & 0.0847 \\
5 & 16384 & 0.0917 & 0.0801 \\
\midrule
\multicolumn{4}{c}{Gemma-3-12B-it} \\
\midrule
2 & 1001 & 0.1133 & 0.1283 \\
2 & 4096 & 0.126 & \textbf{0.1255} \\
2 & 16384 & 0.1238 & 0.1326 \\
3 & 1001 & 0.1099 & 0.1349 \\
3 & 4096 & 0.1145 & 0.1551 \\
3 & 16384 & 0.122 & 0.1436 \\
\rowcolor{green!12}
5 & 1001 & \textbf{0.1034} & 0.1489 \\
5 & 4096 & 0.1124 & 0.1602 \\
5 & 16384 & 0.1046 & 0.1696 \\
\bottomrule
\end{tabular}
}
\captionof{table}{\textbf{Full ablation over the granularity of the canonical numeric output space.} Rows vary the number of decimal places $d$ and the cap on the number of support bins used during training. All checkpoints are evaluated against a shared high-resolution reference with $d=5$ and max bins $=16384$. Results are reported as $Q_{95}$-$Q_{05}$ normalized Wasserstein-1, averaged within family and then aggregated across families with a median. The green row denotes the final selected setting. Values in bold denote the best setting for each model and split.}
\label{tab:grid-ablation}

\end{minipage}
\hfill
\begin{minipage}[t]{0.60\textwidth}
\centering
\scriptsize
\resizebox{\linewidth}{!}{%
\begin{tabular}{ccccccc}
\toprule
$d$ & \shortstack{max\\bins} & spp & epochs & \shortstack{total\\draws} & \shortstack{Held-out\\$W_1 \downarrow$} & \shortstack{Unseen\\$W_1 \downarrow$} \\
\midrule
\multicolumn{7}{c}{Qwen3-14B} \\
\midrule
3 & 1001 & 16 & 2 & 32 & 0.0759 & 0.0631 \\
2 & 1001 & 16 & 2 & 32 & 0.0781 & 0.07 \\
5 & 1001 & 16 & 2 & 32 & 0.0757 & 0.0625 \\
3 & 4096 & 16 & 2 & 32 & 0.0753 & 0.0637 \\
3 & 16384 & 16 & 2 & 32 & \textbf{0.0627} & 0.0613 \\
3 & 1001 & 8 & 4 & 32 & 0.0797 & 0.061 \\
3 & 1001 & 32 & 1 & 32 & 0.0768 & \textbf{0.05} \\
5 & 16384 & 8 & 4 & 32 & 0.085 & 0.0566 \\
\rowcolor{green!12}
5 & 16384 & 16 & 2 & 32 & 0.0736 & 0.0529 \\
5 & 16384 & 32 & 1 & 32 & 0.0701 & 0.0597 \\
\midrule
\multicolumn{7}{c}{Gemma-3-12B-it} \\
\midrule
3 & 1001 & 16 & 2 & 32 & 0.1149 & 0.0965 \\
2 & 1001 & 16 & 2 & 32 & 0.1166 & 0.1031 \\
5 & 1001 & 16 & 2 & 32 & 0.1116 & 0.0941 \\
3 & 4096 & 16 & 2 & 32 & 0.1236 & 0.0922 \\
3 & 16384 & 16 & 2 & 32 & 0.1136 & 0.0814 \\
3 & 1001 & 8 & 4 & 32 & 0.1184 & 0.0823 \\
3 & 1001 & 32 & 1 & 32 & 0.1138 & 0.0777 \\
5 & 16384 & 8 & 4 & 32 & 0.1179 & 0.0759 \\
\rowcolor{green!12}
5 & 16384 & 16 & 2 & 32 & \textbf{0.105} & 0.0811 \\
5 & 16384 & 32 & 1 & 32 & 0.1218 & \textbf{0.0708} \\
\bottomrule
\end{tabular}
}
\captionof{table}{\textbf{Hard-target fine-tuning ablation on the two large models.} The anchor setting uses $d=3$, max bins $=1001$, and a matched training budget of 16 samples per prompt per epoch for 2 epochs. Rows vary the output-space construction and the split between samples per prompt and epochs, keeping the total number of sampled draws fixed at 32 per prompt. All checkpoints are evaluated against the shared reference with $d=5$ and max bins $=16384$. Results are reported as $Q_{95}$-$Q_{05}$ normalized Wasserstein-1, averaged within family and then aggregated across families with a median. The green row denotes the final selected setting. Values in bold denote the best setting for each model and split.}
\label{tab:hard-label-ablation}

\end{minipage}
\end{table*}

\clearpage
\section{Soft-Target and Hard-Target at a Matched Optimizer Budget}
\label{app:matched-budget}
\suppressfloats[t]  %

The selected soft-target and hard-target configurations differ by roughly a factor of ten in optimizer steps (189 and 1988), so the main-text results do not isolate the objective. We therefore retrain soft-target fine-tuning for 32 epochs, 2016 steps, on four models spanning three families, keeping data, LoRA configuration, and optimizer unchanged, and compare the three fine-tuned conditions on every evaluation axis. Equal steps do not mean equal information: at 2016 steps the soft-target model sees the same exact targets many times over. The run also checks the soft-target budget itself, which was not part of the ablations in Appendix~\ref{sec:appendix-ablations}.

\begin{table*}[t]
\centering
\footnotesize
\setlength{\tabcolsep}{4pt}
\begin{tabular}{ll rr rr}
\toprule
& & \multicolumn{2}{c}{Logit KL $\downarrow$} & \multicolumn{2}{c}{$W_1$ $\downarrow$} \\
\cmidrule(lr){3-4}\cmidrule(lr){5-6}
Model & Condition & \shortstack{held-out\\parameters} & \shortstack{held-out\\families} & \shortstack{held-out\\parameters} & \shortstack{held-out\\families} \\
\midrule
Qwen3-1.7B & base & \ci{6.351}{5.62}{7.17} & \ci{6.524}{4.21}{9.66} & \ci{0.362}{0.22}{0.65} & \ci{0.356}{0.26}{0.48} \\
 & soft@189 & \ci{0.578}{0.46}{0.70} & \ci{0.433}{0.29}{0.59} & \ci{0.287}{0.20}{0.57} & \ci{0.177}{0.11}{0.31} \\
 & soft@2016 & \ci{0.469}{0.34}{0.61} & \ci{0.536}{0.35}{0.81} & \ci{0.110}{0.07}{0.14} & \ci{0.172}{0.12}{0.30} \\
 & hard@1988 & \ci{0.476}{0.35}{0.60} & \ci{0.536}{0.32}{0.87} & \ci{0.086}{0.07}{0.16} & \ci{0.166}{0.12}{0.31} \\
\midrule
Qwen3-14B & base & \ci{3.022}{2.52}{3.54} & \ci{3.272}{2.13}{4.86} & \ci{0.227}{0.19}{0.29} & \ci{0.246}{0.23}{0.32} \\
 & soft@189 & \ci{0.459}{0.33}{0.60} & \ci{0.314}{0.18}{0.44} & \ci{0.074}{0.06}{0.10} & \ci{0.097}{0.07}{0.14} \\
 & soft@2016 & \ci{0.446}{0.32}{0.58} & \ci{0.323}{0.19}{0.45} & \ci{0.044}{0.03}{0.08} & \ci{0.129}{0.06}{0.15} \\
 & hard@1988 & \ci{0.454}{0.32}{0.59} & \ci{0.314}{0.19}{0.43} & \ci{0.053}{0.04}{0.09} & \ci{0.116}{0.07}{0.14} \\
\midrule
Gemma-3-12B-it & base & \ci{5.001}{4.10}{6.03} & \ci{6.236}{4.39}{8.24} & \ci{0.269}{0.24}{0.38} & \ci{0.247}{0.21}{0.29} \\
 & soft@189 & \ci{0.496}{0.37}{0.62} & \ci{0.407}{0.28}{0.51} & \ci{0.149}{0.09}{0.21} & \ci{0.135}{0.08}{0.22} \\
 & soft@2016 & \ci{0.457}{0.33}{0.60} & \ci{0.351}{0.23}{0.46} & \ci{0.070}{0.04}{0.11} & \ci{0.147}{0.08}{0.19} \\
 & hard@1988 & \ci{0.459}{0.33}{0.59} & \ci{0.385}{0.26}{0.50} & \ci{0.086}{0.06}{0.13} & \ci{0.156}{0.08}{0.24} \\
\midrule
GPT-OSS-20B & base & \ci{1.910}{1.52}{2.32} & \ci{1.498}{0.97}{1.95} & \ci{0.773}{0.35}{40.62} & \ci{0.635}{0.20}{$3.3{\times}10^{3}$} \\
 & soft@189 & \ci{0.762}{0.53}{0.99} & \ci{0.497}{0.31}{0.70} & \ci{0.064}{0.05}{0.10} & \ci{0.149}{0.08}{0.19} \\
 & soft@2016 & \ci{0.766}{0.53}{1.02} & \ci{0.484}{0.28}{0.69} & \ci{0.034}{0.03}{0.05} & \ci{0.099}{0.07}{0.19} \\
 & hard@1988 & \ci{0.743}{0.52}{0.97} & \ci{0.468}{0.26}{0.67} & \ci{0.048}{0.03}{0.07} & \ci{0.112}{0.05}{0.14} \\
\bottomrule
\end{tabular}
\caption{\textbf{Structured sampling at a matched optimizer budget.} soft@2016 is soft-target fine-tuning retrained on the same data at a budget close to the 1988 steps of hard-target. Point estimate with its 95\% hierarchical bootstrap interval ($B=2000$; families, then distributions within family) in smaller type. Intervals resample evaluation prompts, not training runs. The suffix of each condition is its number of optimizer steps.}
\label{tab:matched-budget-structured}
\end{table*}

On structured sampling the three conditions perform similarly (Table~\ref{tab:matched-budget-structured}). On all four models every matched-budget estimate lies inside the hard-target interval, on held-out parameters and held-out families alike, and the longer soft-target run closes the gap to hard-target on held-out parameters. Soft-target fine-tuning therefore reaches the same logit KL with roughly a tenth of the optimizer steps. We did not train hard-target with fewer steps, so we cannot say whether it would also reach this level that quickly.

\begin{table*}[t]
\centering
\footnotesize
\setlength{\tabcolsep}{2.5pt}
\renewcommand{\cifont}{\scriptsize}
\begin{tabular}{ll rr rr rr r}
\toprule
& & \multicolumn{2}{c}{Retention} & \multicolumn{2}{c}{PALOMA} & \multicolumn{2}{c}{NoveltyBench} & Open-gen \\
\cmidrule(lr){3-4}\cmidrule(lr){5-6}\cmidrule(lr){7-8}\cmidrule(lr){9-9}
Model & Condition & gp-IRT & GSM8K & ppl & \shortstack{bits/\\byte} & utility & distinct & \shortstack{median\\first-token\\support} \\
\midrule
Qwen3-1.7B & base & \cis{0.451}{0.41}{0.49} & \cis{0.690}{0.60}{0.78} & \cis{15.50}{10.7}{21.2} & \cis{0.918}{0.823}{0.989} & \cis{2.804}{2.69}{2.92} & \cis{4.07}{3.90}{4.25} & \cis{5.5}{4.0}{8.0} \\[2pt]
 & soft@189 & \cis{0.394}{0.36}{0.43} & \cis{0.480}{0.38}{0.57} & \cis{14.60}{10.6}{19.9} & \cis{0.898}{0.812}{0.972} & \cis{3.143}{3.02}{3.27} & \cis{6.66}{6.48}{6.84} & \cis{466.0}{301.0}{571.7} \\[2pt]
 & soft@2016 & \cis{0.424}{0.39}{0.46} & \cis{0.690}{0.60}{0.78} & \cis{15.01}{10.6}{20.3} & \cis{0.907}{0.814}{0.979} & \cis{2.994}{2.88}{3.12} & \cis{5.06}{4.89}{5.24} & \cis{166.0}{121.0}{211.0} \\[2pt]
 & hard@1988 & \cis{0.450}{0.41}{0.49} & \cis{0.680}{0.59}{0.76} & \cis{14.78}{10.2}{20.2} & \cis{0.902}{0.807}{0.974} & \cis{3.094}{2.97}{3.22} & \cis{5.35}{5.17}{5.53} & \cis{151.0}{91.0}{196.0} \\[2pt]
\midrule
Qwen3-14B & base & \cis{0.682}{0.65}{0.72} & \cis{0.860}{0.79}{0.93} & \cis{9.77}{6.8}{13.0} & \cis{0.763}{0.668}{0.831} & \cis{3.392}{3.28}{3.50} & \cis{3.35}{3.20}{3.50} & \cis{2.0}{2.0}{3.0} \\[2pt]
 & soft@189 & \cis{0.719}{0.68}{0.75} & \cis{0.900}{0.84}{0.95} & \cis{9.54}{7.0}{12.6} & \cis{0.755}{0.674}{0.823} & \cis{4.096}{3.97}{4.23} & \cis{5.32}{5.14}{5.50} & \cis{211.0}{181.0}{271.0} \\[2pt]
 & soft@2016 & \cis{0.739}{0.70}{0.77} & \cis{0.890}{0.83}{0.95} & \cis{9.63}{7.1}{12.6} & \cis{0.758}{0.674}{0.824} & \cis{3.818}{3.70}{3.95} & \cis{4.35}{4.18}{4.54} & \cis{54.0}{44.0}{91.0} \\[2pt]
 & hard@1988 & \cis{0.722}{0.68}{0.75} & \cis{0.880}{0.82}{0.94} & \cis{9.58}{6.8}{12.7} & \cis{0.757}{0.666}{0.823} & \cis{3.849}{3.73}{3.97} & \cis{4.45}{4.26}{4.63} & \cis{91.0}{61.0}{121.0} \\[2pt]
\midrule
Gemma-3-12B-it & base & \cis{0.684}{0.65}{0.72} & \cis{0.900}{0.84}{0.95} & \cis{56.56}{26.5}{114.6} & \cis{1.367}{1.137}{1.569} & \cis{2.173}{2.10}{2.25} & \cis{1.96}{1.89}{2.05} & \cis{3.0}{2.0}{3.0} \\[2pt]
 & soft@189 & \cis{0.709}{0.67}{0.74} & \cis{0.890}{0.82}{0.95} & \cis{36.34}{18.4}{63.6} & \cis{1.217}{1.032}{1.365} & \cis{2.863}{2.77}{2.97} & \cis{2.83}{2.71}{2.96} & \cis{25.0}{17.5}{33.0} \\[2pt]
 & soft@2016 & \cis{0.705}{0.67}{0.74} & \cis{0.900}{0.84}{0.95} & \cis{48.86}{21.3}{93.2} & \cis{1.317}{1.071}{1.495} & \cis{2.486}{2.40}{2.57} & \cis{2.32}{2.22}{2.42} & \cis{10.0}{7.0}{12.0} \\[2pt]
 & hard@1988 & \cis{0.698}{0.66}{0.73} & \cis{0.860}{0.79}{0.92} & \cis{47.99}{21.9}{93.8} & \cis{1.311}{1.079}{1.504} & \cis{2.678}{2.59}{2.77} & \cis{2.50}{2.40}{2.61} & \cis{12.0}{10.0}{16.0} \\[2pt]
\midrule
GPT-OSS-20B & base & \cis{0.463}{0.43}{0.50} & \cis{0.620}{0.53}{0.71} & \cis{103.7}{47.2}{274.3} & \cis{1.491}{1.263}{1.789} & \cis{3.786}{3.66}{3.91} & \cis{5.07}{4.93}{5.24} & \cis{20.0}{17.0}{25.0} \\[2pt]
 & soft@189 & \cis{0.332}{0.30}{0.36} & \cis{0.020}{0.00}{0.05} & \cis{99.95}{47.8}{229.8} & \cis{1.479}{1.265}{1.734} & \cis{2.179}{2.09}{2.28} & \cis{8.56}{8.43}{8.69} & \cis{166.0}{151.0}{241.0} \\[2pt]
 & soft@2016 & \cis{0.384}{0.35}{0.42} & \cis{0.030}{0.00}{0.07} & \cis{100.6}{45.9}{252.7} & \cis{1.481}{1.259}{1.761} & \cis{3.359}{3.24}{3.49} & \cis{6.09}{5.92}{6.28} & \cis{38.5}{32.0}{49.0} \\[2pt]
 & hard@1988 & \cis{0.389}{0.36}{0.42} & \cis{0.030}{0.00}{0.07} & \cis{70.17}{33.2}{155.9} & \cis{1.366}{1.161}{1.599} & \cis{2.615}{2.51}{2.72} & \cis{7.68}{7.53}{7.83} & \cis{151.0}{121.0}{181.0} \\[2pt]
\bottomrule
\end{tabular}
\caption{\textbf{Downstream capability and transfer at a matched optimizer budget.} GSM8K is flexible-match accuracy, which accepts the final number anywhere in the answer; NoveltyBench utility and distinct count are higher-is-better; median support is the median over open-ended prompts of the top-90\% first-token support size of the first-step next-token distribution. gp-IRT is the retention measure; PALOMA is reported separately and only rules out a global flattening of the next-token distribution. Intervals are 95\% bootstrap intervals ($B=2000$), shown in smaller type after the point estimate.}
\label{tab:matched-budget-capability}
\end{table*}

The extra steps leave logit KL unchanged and help only $W_1$ on held-out parameters, but they reduce open-ended diversity (Table~\ref{tab:matched-budget-capability}). Median first-token support of the long soft-target runs falls to roughly the hard-target level or below on all four models, and on the three dense models NoveltyBench utility falls to or below the hard-target level while PALOMA perplexity rises slightly above it. The differences between variants in the main-text transfer results are therefore largely effects of the training budget rather than of the objective. GPT-OSS-20B is the exception on NoveltyBench, where the longer run recovers most of the utility lost by the short one. Aggregate gp-IRT stays within 0.03 of hard-target on the dense models.

\clearpage
\section{Prompt-Template Robustness}
\label{app:prompt-robustness}
\suppressfloats[t]  %

Every training prompt follows a single template (Section~\ref{sec:method}), so the gains in Section~\ref{sec:results} could reflect a response to that wording rather than to the distribution it specifies. We test this by rewording the evaluation prompt while keeping the distribution family and every parameter fixed. We compare four versions of the prompt. The \textit{canonical} version is the prompt used throughout the paper. The \textit{JSON spec} version keeps the instruction words and re-encodes the parameter block as a JSON object. The \textit{distractor} version prepends an unrelated sentence (``The weather here has been mild all week.'') to the canonical prompt. The \textit{question} version rephrases the request as a question (``Could you please sample one number from \ldots? Give me ONLY the number.''). None of the versions adds information: the distractor and question versions only add words around the original prompt, and the JSON version restates the same parameters. We run the four versions on 18 of the 36 held-out prompts, those of the Gaussian and Poisson mixtures, Zipf, and Categorical families. All other settings match the main evaluation (1{,}000 samples per prompt, $T=1$, top-$p=1$).

\begin{table*}[t]
\centering
\scriptsize
\setlength{\tabcolsep}{3pt}
\renewcommand{\cifont}{\tiny}
\begin{tabular}{ll rrrr r}
\toprule
Model & Condition & canonical & JSON spec & distractor & question & Spread \\
\midrule
Qwen3-0.6B & base & \ci{2.783}{2.17}{3.77} & \ci{2.442}{1.92}{3.19} & \ci{2.196}{1.68}{3.04} & \ci{3.388}{2.87}{4.22} & 1.192 \\
 & soft@189 & \ci{0.831}{0.36}{1.57} & \ci{1.090}{0.52}{2.09} & \ci{0.710}{0.45}{1.11} & \ci{0.937}{0.44}{1.81} & 0.380 \\
 & hard@1988 & \ci{0.835}{0.38}{1.57} & \ci{1.055}{0.46}{2.08} & \ci{0.761}{0.39}{1.34} & \ci{0.934}{0.41}{1.85} & 0.294 \\
\midrule
Qwen3-1.7B & base & \ci{9.178}{5.87}{14.56} & \ci{10.140}{6.49}{15.66} & \ci{8.382}{4.82}{13.31} & \ci{9.394}{6.02}{13.54} & 1.758 \\
 & soft@189 & \ci{0.526}{0.33}{0.72} & \ci{0.584}{0.40}{0.79} & \ci{0.613}{0.41}{0.82} & \ci{0.597}{0.31}{0.87} & 0.087 \\
 & hard@1988 & \ci{0.731}{0.36}{1.32} & \ci{1.098}{0.38}{2.44} & \ci{0.651}{0.43}{0.92} & \ci{0.531}{0.34}{0.71} & 0.567 \\
\midrule
Qwen3-4B & base & \ci{5.520}{3.36}{8.56} & \ci{5.353}{3.48}{7.51} & \ci{6.297}{3.78}{9.90} & \ci{3.826}{2.75}{4.77} & 2.471 \\
 & soft@189 & \ci{0.409}{0.23}{0.58} & \ci{0.433}{0.23}{0.63} & \ci{0.388}{0.21}{0.55} & \ci{0.466}{0.28}{0.66} & 0.078 \\
 & hard@1988 & \ci{0.376}{0.19}{0.57} & \ci{0.379}{0.17}{0.60} & \ci{0.506}{0.23}{0.79} & \ci{0.387}{0.19}{0.59} & 0.131 \\
\midrule
Qwen3-8B & base & \ci{4.129}{2.74}{6.35} & \ci{3.896}{2.61}{6.38} & \ci{5.108}{3.47}{7.21} & \ci{3.289}{2.37}{4.24} & 1.819 \\
 & soft@189 & \ci{0.427}{0.23}{0.61} & \ci{0.433}{0.24}{0.63} & \ci{0.456}{0.26}{0.65} & \ci{0.388}{0.20}{0.56} & 0.068 \\
 & hard@1988 & \ci{0.413}{0.18}{0.61} & \ci{0.442}{0.22}{0.67} & \ci{0.438}{0.23}{0.64} & \ci{0.373}{0.17}{0.56} & 0.069 \\
\midrule
Qwen3-14B & base & \ci{4.142}{2.16}{6.96} & \ci{2.964}{2.09}{3.83} & \ci{4.497}{2.44}{7.32} & \ci{3.328}{2.12}{4.42} & 1.533 \\
 & soft@189 & \ci{0.337}{0.15}{0.53} & \ci{0.351}{0.16}{0.54} & \ci{0.363}{0.16}{0.57} & \ci{0.332}{0.13}{0.51} & 0.032 \\
 & hard@1988 & \ci{0.338}{0.14}{0.52} & \ci{0.340}{0.15}{0.51} & \ci{0.358}{0.16}{0.54} & \ci{0.342}{0.15}{0.52} & 0.020 \\
\midrule
Gemma-3-1B-it & base & \ci{9.412}{6.35}{12.98} & \ci{10.876}{7.96}{14.26} & \ci{9.215}{6.14}{12.80} & \ci{10.377}{7.60}{13.43} & 1.661 \\
 & soft@189 & \ci{0.700}{0.54}{0.89} & \ci{0.751}{0.46}{1.01} & \ci{0.679}{0.61}{0.78} & \ci{0.979}{0.68}{1.36} & 0.300 \\
 & hard@1988 & \ci{0.928}{0.65}{1.29} & \ci{0.900}{0.58}{1.19} & \ci{1.056}{0.62}{1.66} & \ci{1.197}{0.73}{1.73} & 0.297 \\
\midrule
Gemma-3-4B-it & base & \ci{10.503}{7.03}{14.42} & \ci{10.474}{7.63}{13.77} & \ci{10.768}{8.33}{13.58} & \ci{10.253}{7.35}{13.57} & 0.516 \\
 & soft@189 & \ci{0.473}{0.27}{0.65} & \ci{0.642}{0.32}{1.09} & \ci{0.441}{0.25}{0.60} & \ci{0.637}{0.29}{1.07} & 0.201 \\
 & hard@1988 & \ci{0.518}{0.31}{0.74} & \ci{0.578}{0.32}{0.92} & \ci{0.497}{0.30}{0.68} & \ci{0.481}{0.29}{0.64} & 0.097 \\
\midrule
Gemma-3-12B-it & base & \ci{6.821}{4.03}{9.85} & \ci{6.430}{3.86}{9.24} & \ci{6.769}{4.12}{9.59} & \ci{7.648}{5.34}{10.36} & 1.219 \\
 & soft@189 & \ci{0.499}{0.37}{0.60} & \ci{0.517}{0.38}{0.66} & \ci{0.459}{0.33}{0.58} & \ci{0.477}{0.35}{0.58} & 0.058 \\
 & hard@1988 & \ci{0.463}{0.31}{0.60} & \ci{0.456}{0.26}{0.63} & \ci{0.446}{0.29}{0.60} & \ci{0.443}{0.29}{0.58} & 0.020 \\
\midrule
Gemma-3-27B-it & base & \ci{11.099}{6.87}{15.78} & \ci{11.419}{7.17}{16.05} & \ci{11.414}{7.17}{15.95} & \ci{12.097}{7.68}{16.83} & 0.999 \\
 & soft@189 & \ci{0.371}{0.19}{0.54} & \ci{0.396}{0.20}{0.57} & \ci{0.403}{0.22}{0.57} & \ci{0.364}{0.17}{0.54} & 0.039 \\
 & hard@1988 & \ci{0.359}{0.18}{0.52} & \ci{0.363}{0.16}{0.55} & \ci{0.368}{0.18}{0.56} & \ci{0.362}{0.18}{0.55} & 0.009 \\
\midrule
Llama-3.2-1B-it & base & \ci{2.541}{1.84}{3.19} & \ci{3.025}{2.38}{3.60} & \ci{2.489}{1.60}{3.33} & \ci{2.243}{1.45}{2.95} & 0.781 \\
 & soft@189 & \ci{0.935}{0.44}{1.55} & \ci{0.851}{0.54}{1.10} & \ci{0.993}{0.49}{1.69} & \ci{0.947}{0.44}{1.63} & 0.142 \\
 & hard@1988 & \ci{1.326}{0.46}{2.60} & \ci{1.368}{0.65}{2.37} & \ci{1.384}{0.49}{2.73} & \ci{1.182}{0.46}{2.30} & 0.202 \\
\midrule
Llama-3.2-3B-it & base & \ci{2.456}{1.70}{3.57} & \ci{2.278}{1.74}{2.86} & \ci{2.330}{1.67}{3.22} & \ci{1.882}{1.32}{2.65} & 0.574 \\
 & soft@189 & \ci{1.089}{0.42}{2.14} & \ci{0.990}{0.43}{1.79} & \ci{1.028}{0.43}{1.86} & \ci{0.927}{0.44}{1.58} & 0.162 \\
 & hard@1988 & \ci{1.330}{0.39}{2.92} & \ci{1.389}{0.40}{3.03} & \ci{1.256}{0.39}{2.67} & \ci{1.307}{0.39}{2.79} & 0.132 \\
\midrule
GPT-OSS-20B & base & \ci{1.646}{1.01}{2.25} & \ci{1.513}{0.84}{2.16} & \ci{1.546}{0.94}{2.13} & \ci{1.839}{1.14}{2.51} & 0.326 \\
 & soft@189 & \ci{0.505}{0.23}{0.78} & \ci{0.519}{0.23}{0.79} & \ci{0.518}{0.24}{0.80} & \ci{0.522}{0.24}{0.79} & 0.017 \\
 & hard@1988 & \ci{0.465}{0.18}{0.75} & \ci{0.475}{0.19}{0.76} & \ci{0.470}{0.18}{0.76} & \ci{0.462}{0.19}{0.74} & 0.013 \\
\bottomrule
\end{tabular}
\caption{\textbf{Prompt-template robustness.} Logit KL under four rewordings of the evaluation prompt that keep the distribution family and every parameter fixed, on the 18 held-out prompts of the Gaussian and Poisson mixtures, Zipf, and Categorical families. Spread is the largest minus smallest value across the four versions of the prompt.}
\label{tab:prompt-robustness}
\end{table*}

The improvement survives every rewording (Table~\ref{tab:prompt-robustness}). For all 12 models, the worst version of either fine-tuned variant stays below the best version of the base model. The fine-tuned models also vary less across versions than the base models do, in absolute terms, for every model. Part of this reflects their lower logit KL, but on the larger checkpoints rewording the prompt barely changes the result: the spread across versions is at most 0.06 for Qwen3-14B, Gemma-3-12B-it, Gemma-3-27B-it, and GPT-OSS-20B under both variants. If the models had only learned the training wording, they would do worse on the reworded prompts, and they do not. The largest sensitivity among fine-tuned models appears on the smallest Qwen checkpoints, where the JSON spec version is the hardest (e.g., 1.10 for Qwen3-1.7B hard-target against 0.53 to 0.73 on the other versions).

\clearpage
\section{String Seed of Thought as an Inference-Time Baseline}
\label{sec:appendix-ssot}
\suppressfloats[t]  %

Tables~\ref{tab:ssot-structured-ood} and~\ref{tab:ssot-structured-unseen} compare the base checkpoint, soft-target Calibration Fine-Tuning, and a structured-sampling baseline based on String Seed of Thought (SSOT) prompting \citep{misaki2025stringseed}. We follow the SSOT protocol, adapting only the prompt wrapper needed to fit the chat format of each model family. SSOT is evaluated on a subset of the held-out families (Bernoulli, Poisson, Maxwell, TruncNorm, Chi, and Weibull), and every column of Table~\ref{tab:ssot-structured-ood}, including Base and Soft, is computed on that subset. Unlike Calibration Fine-Tuning, SSOT does not update model parameters. It instead asks the model to generate an internal random seed string and use that seed through an explicit reasoning trace before emitting the final sampled answer. SSOT is a sample-only inference-time baseline, so we do not report trie-based logit KL. A dash marks a run in which at least one family has no valid sample, so its aggregate \(W_1\) is undefined. We also use 250 generations per prompt rather than 1{,}000, because SSOT changes the generation protocol itself: while our main evaluation disables reasoning traces and asks the model to emit only the final sample, SSOT explicitly requires the model to generate a seed string and reason from it before producing the answer. In practice, these reasoning traces can be verbose and model-dependent, making each sample substantially more expensive to generate.

\begin{table*}[htbp]
\centering
\footnotesize
\resizebox{0.82\textwidth}{!}{%
\begin{tabular}{lcccccc}
\toprule
 & \multicolumn{3}{c}{$W_1$ $\downarrow$} & \multicolumn{3}{c}{Valid rate $\uparrow$} \\
\cmidrule(lr){2-4}\cmidrule(lr){5-7}
Model & Base & Soft & SSOT & Base & Soft & SSOT \\
\midrule
Qwen3-0.6B & 0.462 & 170.423 & 2.05e+30 & 17.3\% & 84.1\% & 37.5\% \\
Qwen3-1.7B & 0.3574 & 0.2311 & 0.3296 & 98.0\% & 94.3\% & 48.4\% \\
Qwen3-4B & 0.3264 & 0.2299 & -- & 100.0\% & 99.4\% & 4.3\% \\
Qwen3-8B & 0.2905 & 0.0893 & -- & 97.3\% & 99.1\% & 7.6\% \\
Qwen3-14B & 0.2459 & 0.0847 & 0.1641 & 100.0\% & 100.0\% & 59.9\% \\
Gemma-3-1B-it & 96.9663 & 0.3443 & 4.21e+223 & 72.1\% & 66.8\% & 44.9\% \\
Gemma-3-4B-it & 0.4666 & 0.3394 & 3.87e+07 & 97.3\% & 90.3\% & 8.5\% \\
Gemma-3-12B-it & 0.2164 & 0.1018 & 1.2646 & 99.8\% & 99.3\% & 90.6\% \\
Gemma-3-27B-it & 0.2178 & 0.1287 & 0.18 & 99.7\% & 99.9\% & 96.5\% \\
Llama-3.2-1B-it & 130.0273 & 18.7111 & 6.46e+16 & 56.5\% & 88.0\% & 30.4\% \\
Llama-3.2-3B-it & 2.8905 & 0.1447 & 1.82e+16 & 61.1\% & 96.1\% & 64.5\% \\
GPT-OSS-20B & 5.47e+03 & 0.1734 & 0.9099 & 99.3\% & 100.0\% & 95.1\% \\
\bottomrule
\end{tabular}
}
\caption{\textbf{Held-out-family structured-sampling comparison between Base, Soft, and the SSOT inference-time baseline}, on a subset of the held-out families (Bernoulli, Poisson, Maxwell, TruncNorm, Chi, and Weibull). All columns are computed on this subset. SSOT uses 250 generations per prompt. Lower $W_1$ and higher valid rate are better; ``--'' indicates undefined aggregate $W_1$.}
\label{tab:ssot-structured-ood}
\end{table*}

\begin{table*}[htbp]
\centering
\footnotesize
\resizebox{0.82\textwidth}{!}{%
\begin{tabular}{lcccccc}
\toprule
 & \multicolumn{3}{c}{Unseen-Param $W_1$ $\downarrow$} & \multicolumn{3}{c}{Valid rate $\uparrow$} \\
\cmidrule(lr){2-4}\cmidrule(lr){5-7}
Model & Base & Soft & SSOT & Base & Soft & SSOT \\
\midrule
Qwen3-0.6B & -- & 4.6399 & 8.36e+10 & 3.1\% & 96.1\% & 30.7\% \\
Qwen3-1.7B & 0.3617 & 0.2868 & 0.3706 & 82.9\% & 96.8\% & 46.5\% \\
Qwen3-4B & -- & 0.175 & -- & 94.2\% & 99.0\% & 0.8\% \\
Qwen3-8B & -- & 0.1147 & -- & 88.1\% & 99.8\% & 3.3\% \\
Qwen3-14B & 0.2266 & 0.074 & 0.1606 & 98.0\% & 99.9\% & 50.5\% \\
Gemma-3-1B-it & -- & 0.3501 & 1.42e+260 & 66.3\% & 86.7\% & 44.8\% \\
Gemma-3-4B-it & 0.3124 & 0.2834 & 13.5034 & 75.3\% & 98.3\% & 10.4\% \\
Gemma-3-12B-it & 0.2692 & 0.149 & 3.2829 & 97.6\% & 99.7\% & 88.2\% \\
Gemma-3-27B-it & 0.2647 & 0.0832 & 0.3394 & 98.4\% & 99.9\% & 95.6\% \\
Llama-3.2-1B-it & 33.6536 & 0.3513 & 5.36e+15 & 63.9\% & 97.8\% & 28.9\% \\
Llama-3.2-3B-it & 1.1658 & 0.2325 & 6.7993 & 68.3\% & 98.5\% & 64.0\% \\
GPT-OSS-20B & 0.7734 & 0.0642 & 0.3066 & 96.6\% & 99.9\% & 95.1\% \\
\bottomrule
\end{tabular}
}
\caption{\textbf{Unseen-parameter structured-sampling comparison between Base, Soft, and the SSOT inference-time baseline.} SSOT uses 250 generations per prompt. Lower $W_1$ and higher valid rate are better; ``--'' indicates undefined aggregate $W_1$.}
\label{tab:ssot-structured-unseen}
\end{table*}

\begin{figure*}[htbp]
\centering
\includegraphics[width=0.92\textwidth]{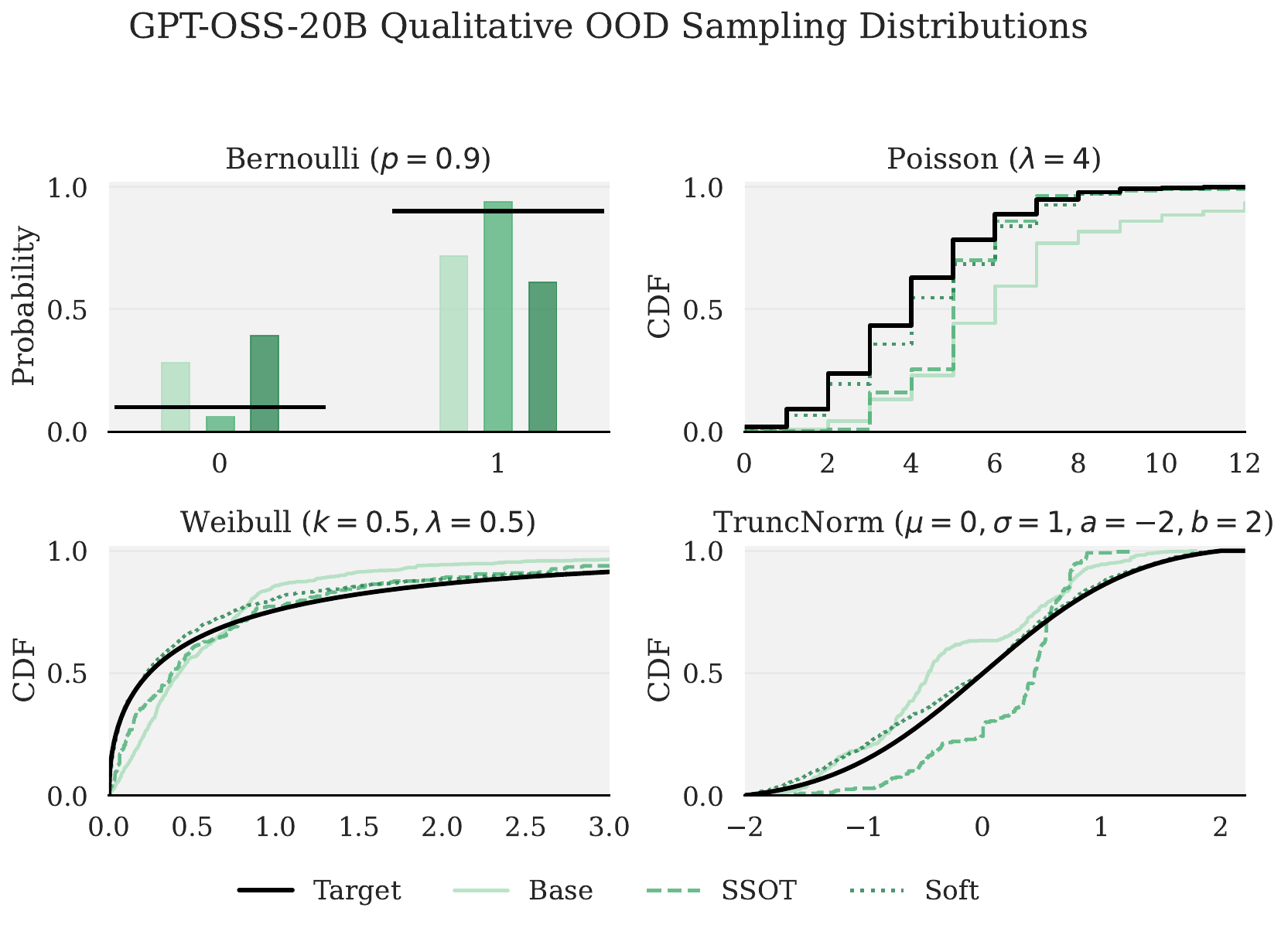}
\caption{\textbf{Qualitative OOD sampling comparison for GPT-OSS-20B.} Each panel overlays the target distribution with empirical samples from the base checkpoint, the SSOT inference-time baseline, and soft-target Calibration Fine-Tuning on the same four held-out OOD configurations used in the main qualitative structured-sampling figure.}
\label{fig:appendix-ssot-qualitative}
\end{figure*}

SSOT helps only some models. SSOT can improve sampling when the model follows the protocol well. The strongest evidence comes from GPT-OSS-20B, which improves over the base checkpoint on both held-out families and unseen parameters; Qwen3-14B also improves on both splits but with lower validity, while Gemma-3-27B-it improves on held-out families but not on unseen parameters. However, this improvement remains smaller than the one obtained by soft-target Calibration Fine-Tuning, as also shown qualitatively in Figure~\ref{fig:appendix-ssot-qualitative}. Across the other checkpoints, the behavior is less consistent. Some models produce finite but weak gains, some preserve high validity but worsen distributional fit, and others fail through invalid outputs, zero-valid distributions, or extreme numeric samples that make the aggregate \(W_1\) undefined or huge.

Overall, prompting alone helps only the models that follow the SSOT protocol reliably, and even then less than fine-tuning. Fine-tuning changes the sampling distribution directly, keeps near-perfect validity on the strongest checkpoints, and needs no reasoning trace per sample, so it is also much cheaper at inference time.

\clearpage
\section{Temperature Scaling as an Inference-Time Baseline}
\label{app:temperature}
\suppressfloats[t]  %

A cheaper alternative to fine-tuning is to change how the base model is sampled. We sweep the sampling temperature of every base checkpoint over $T \in \{0.5, 0.7, 1.0, 1.5, 2.0, 3.0, 5.0\}$ on the 36 held-out prompts, keeping all other settings as in the main evaluation, and compare against both fine-tuned variants sampled at $T=1$. Temperature divides every next-token logit by the same constant, so it preserves the ranking of tokens at every prefix. So temperature can only flatten or sharpen what the base model already prefers. It cannot move mass to outputs the base model ranks low.

\begin{table*}[t]
\centering
\scriptsize
\setlength{\tabcolsep}{1.2pt}
\renewcommand{\cifont}{\tiny}
\begin{tabular}{lrrrrrrrrr}
\toprule
& \multicolumn{7}{c}{Base model at temperature $T$} & \multicolumn{2}{c}{Fine-tuned, $T{=}1$} \\
\cmidrule(lr){2-8}\cmidrule(lr){9-10}
Model & $T{=}0.5$ & $T{=}0.7$ & $T{=}1.0$ & $T{=}1.5$ & $T{=}2.0$ & $T{=}3.0$ & $T{=}5.0$ & Soft & Hard \\
\midrule
Qwen3-0.6B & \cis{6.2}{0.2}{17.5} & \cis{7.1}{0.8}{16.8} & \cis{9.5}{3.4}{18.3} & \cis{8.4}{5.4}{11.8} & \cis{1.4}{0.8}{1.9} & \cis{0.0}{0.0}{0.0} & \cis{0.0}{0.0}{0.0} & \cis{74.9}{53.9}{94.3} & \cis{75.7}{53.5}{93.7} \\[2pt]
Qwen3-1.7B & \cis{84.6}{64.4}{99.7} & \cis{84.9}{63.9}{99.2} & \cis{84.8}{66.2}{98.7} & \cis{82.4}{63.8}{96.5} & \cis{56.8}{43.3}{68.1} & \cis{3.1}{2.1}{4.2} & \cis{0.0}{0.0}{0.0} & \cis{92.9}{85.1}{98.6} & \cis{90.3}{74.1}{99.8} \\[2pt]
Qwen3-4B & \cis{92.7}{77.1}{100.0} & \cis{92.8}{77.4}{100.0} & \cis{92.9}{76.4}{100.0} & \cis{91.9}{76.9}{99.2} & \cis{67.1}{55.4}{75.3} & \cis{3.1}{2.3}{4.0} & \cis{0.0}{0.0}{0.0} & \cis{99.1}{97.7}{99.8} & \cis{99.8}{99.5}{100.0} \\[2pt]
Qwen3-8B & \cis{92.4}{77.2}{100.0} & \cis{92.8}{79.2}{99.9} & \cis{93.3}{81.8}{99.8} & \cis{92.9}{82.9}{99.0} & \cis{82.2}{73.7}{89.4} & \cis{7.5}{3.9}{11.7} & \cis{0.0}{0.0}{0.0} & \cis{99.4}{98.4}{99.9} & \cis{99.7}{99.1}{100.0} \\[2pt]
Qwen3-14B & \cis{95.0}{82.3}{100.0} & \cis{95.2}{83.9}{100.0} & \cis{95.5}{85.1}{100.0} & \cis{95.5}{85.3}{100.0} & \cis{91.6}{81.3}{97.1} & \cis{17.0}{13.2}{21.8} & \cis{0.1}{0.0}{0.1} & \cis{99.8}{99.5}{100.0} & \cis{99.9}{99.7}{100.0} \\[2pt]
Gemma-3-1B-it & \cis{70.0}{44.2}{91.6} & \cis{69.5}{45.5}{91.5} & \cis{69.1}{44.7}{91.0} & \cis{66.3}{43.3}{87.1} & \cis{29.0}{19.7}{37.8} & \cis{0.4}{0.2}{0.7} & \cis{0.0}{0.0}{0.0} & \cis{67.0}{47.8}{86.7} & \cis{60.1}{35.7}{84.1} \\[2pt]
Gemma-3-4B-it & \cis{95.5}{87.7}{100.0} & \cis{95.6}{88.9}{99.7} & \cis{95.0}{89.1}{98.9} & \cis{92.5}{88.4}{95.8} & \cis{69.2}{63.1}{74.9} & \cis{2.3}{1.0}{4.0} & \cis{0.0}{0.0}{0.0} & \cis{89.1}{76.9}{97.8} & \cis{95.9}{89.0}{99.7} \\[2pt]
Gemma-3-12B-it & \cis{95.0}{84.2}{100.0} & \cis{95.0}{86.4}{99.8} & \cis{94.9}{87.3}{99.6} & \cis{92.7}{86.8}{97.1} & \cis{67.0}{52.4}{79.4} & \cis{1.4}{0.6}{2.3} & \cis{0.0}{0.0}{0.0} & \cis{99.6}{98.8}{100.0} & \cis{99.5}{99.0}{99.9} \\[2pt]
Gemma-3-27B-it & \cis{85.9}{65.7}{99.9} & \cis{86.0}{67.1}{99.8} & \cis{87.2}{68.7}{99.6} & \cis{88.3}{74.3}{98.9} & \cis{82.4}{68.0}{93.1} & \cis{29.6}{14.3}{46.4} & \cis{0.0}{0.0}{0.0} & \cis{99.9}{99.8}{100.0} & \cis{99.8}{99.5}{100.0} \\[2pt]
Llama-3.2-1B-it & \cis{8.5}{2.7}{17.1} & \cis{23.8}{10.9}{37.1} & \cis{52.7}{31.7}{72.1} & \cis{33.5}{18.3}{47.6} & \cis{4.9}{2.7}{7.0} & \cis{0.1}{0.0}{0.1} & \cis{0.0}{0.0}{0.0} & \cis{82.8}{66.6}{95.8} & \cis{82.4}{60.9}{98.2} \\[2pt]
Llama-3.2-3B-it & \cis{68.0}{42.2}{91.5} & \cis{65.5}{40.6}{90.2} & \cis{62.6}{39.1}{86.0} & \cis{24.5}{15.0}{33.7} & \cis{2.0}{1.4}{2.8} & \cis{0.0}{0.0}{0.0} & \cis{0.0}{0.0}{0.0} & \cis{89.2}{72.7}{98.8} & \cis{90.8}{73.9}{99.8} \\[2pt]
GPT-OSS-20B & \cis{99.9}{99.8}{100.0} & \cis{99.9}{99.6}{100.0} & \cis{99.0}{98.0}{99.4} & \cis{77.9}{73.5}{82.3} & \cis{17.7}{10.2}{26.3} & \cis{0.1}{0.0}{0.2} & \cis{0.0}{0.0}{0.0} & \cis{99.8}{99.3}{100.0} & \cis{99.8}{99.4}{100.0} \\[2pt]
\bottomrule
\end{tabular}
\caption{\textbf{Temperature sweep on the base model, valid \%, against both fine-tuned variants at $T{=}1$.} Held-out families (12 families, 36 prompts). The $T{=}1$ column is the base model under the main evaluation settings. Logit KL is omitted because it does not depend on the sampling temperature.}
\label{tab:temperature-valid}
\end{table*}

\begin{table*}[t]
\centering
\scriptsize
\setlength{\tabcolsep}{1.2pt}
\renewcommand{\cifont}{\tiny}
\begin{tabular}{lrrrrrrrrr}
\toprule
& \multicolumn{7}{c}{Base model at temperature $T$} & \multicolumn{2}{c}{Fine-tuned, $T{=}1$} \\
\cmidrule(lr){2-8}\cmidrule(lr){9-10}
Model & $T{=}0.5$ & $T{=}0.7$ & $T{=}1.0$ & $T{=}1.5$ & $T{=}2.0$ & $T{=}3.0$ & $T{=}5.0$ & Soft & Hard \\
\midrule
Qwen3-0.6B & \cis{0.244}{0.20}{0.40} & \cis{0.279}{0.24}{0.39} & \cis{0.313}{0.20}{0.61} & \cis{0.464}{0.26}{11.15} & \cis{0.416}{0.26}{0.55} & -- & -- & \cis{0.237}{0.16}{86.12} & \cis{0.209}{0.13}{0.30} \\[2pt]
Qwen3-1.7B & \cis{0.357}{0.29}{0.50} & \cis{0.357}{0.28}{0.49} & \cis{0.356}{0.26}{0.48} & \cis{0.348}{0.26}{0.60} & \cis{0.338}{0.24}{0.64} & \cis{0.341}{0.24}{0.75} & -- & \cis{0.177}{0.11}{0.31} & \cis{0.166}{0.12}{0.31} \\[2pt]
Qwen3-4B & \cis{0.339}{0.26}{0.38} & \cis{0.342}{0.28}{0.41} & \cis{0.332}{0.27}{0.40} & \cis{0.319}{0.25}{0.40} & \cis{0.324}{0.24}{0.42} & \cis{0.252}{0.19}{0.32} & \cis{0.000}{0.00}{0.00} & \cis{0.154}{0.12}{0.22} & \cis{0.163}{0.11}{0.22} \\[2pt]
Qwen3-8B & \cis{0.350}{0.29}{0.53} & \cis{0.339}{0.27}{0.57} & \cis{0.322}{0.25}{0.59} & \cis{0.287}{0.21}{0.57} & \cis{0.262}{0.18}{0.62} & \cis{0.252}{0.17}{0.60} & -- & \cis{0.140}{0.08}{0.19} & \cis{0.120}{0.07}{0.16} \\[2pt]
Qwen3-14B & \cis{0.307}{0.25}{0.35} & \cis{0.277}{0.24}{0.35} & \cis{0.246}{0.23}{0.31} & \cis{0.227}{0.18}{0.32} & \cis{0.229}{0.15}{0.33} & \cis{0.209}{0.14}{0.37} & \cis{0.151}{0.03}{0.32} & \cis{0.097}{0.07}{0.14} & \cis{0.116}{0.07}{0.14} \\[2pt]
Gemma-3-1B-it & \cis{0.409}{0.30}{4.43} & \cis{0.500}{0.32}{5.23} & \cis{0.590}{0.33}{48.47} & \cis{874.9}{0.37}{$1.5{\times}10^{5}$} & \cis{29.187}{0.55}{$2.3{\times}10^{5}$} & \cis{0.487}{0.16}{2.31} & -- & \cis{0.242}{0.17}{0.36} & \cis{0.287}{0.19}{0.44} \\[2pt]
Gemma-3-4B-it & \cis{0.405}{0.31}{0.53} & \cis{0.395}{0.30}{0.53} & \cis{0.370}{0.29}{0.56} & \cis{0.342}{0.27}{0.58} & \cis{0.373}{0.31}{0.71} & \cis{0.319}{0.28}{0.62} & -- & \cis{0.244}{0.14}{0.40} & \cis{0.183}{0.12}{0.29} \\[2pt]
Gemma-3-12B-it & \cis{0.272}{0.23}{0.31} & \cis{0.261}{0.22}{0.30} & \cis{0.247}{0.21}{0.29} & \cis{0.225}{0.18}{0.32} & \cis{0.201}{0.16}{0.40} & \cis{0.260}{0.20}{0.81} & -- & \cis{0.135}{0.08}{0.22} & \cis{0.156}{0.08}{0.24} \\[2pt]
Gemma-3-27B-it & \cis{0.300}{0.23}{0.34} & \cis{0.266}{0.22}{0.33} & \cis{0.241}{0.20}{0.30} & \cis{0.230}{0.19}{0.28} & \cis{0.234}{0.18}{0.27} & \cis{0.213}{0.15}{0.26} & \cis{0.310}{0.00}{0.70} & \cis{0.104}{0.08}{0.28} & \cis{0.120}{0.07}{0.26} \\[2pt]
Llama-3.2-1B-it & \cis{20.872}{1.19}{65.33} & \cis{25.035}{1.89}{94.81} & \cis{63.516}{1.80}{133.1} & \cis{$7.6{\times}10^{8}$}{2.17}{$4.8{\times}10^{12}$} & \cis{720.1}{1.90}{$5.9{\times}10^{3}$} & \cis{41.867}{20.65}{94.90} & -- & \cis{0.339}{0.17}{18.10} & \cis{0.260}{0.17}{0.41} \\[2pt]
Llama-3.2-3B-it & \cis{0.525}{0.29}{1.25} & \cis{1.314}{0.29}{4.50} & \cis{5.168}{0.43}{15.60} & \cis{$3.3{\times}10^{10}$}{1.29}{$9.1{\times}10^{13}$} & \cis{14.015}{0.96}{$4.3{\times}10^{4}$} & \cis{2.283}{0.00}{20.90} & -- & \cis{0.196}{0.11}{0.33} & \cis{0.186}{0.11}{0.29} \\[2pt]
GPT-OSS-20B & \cis{0.286}{0.20}{0.41} & \cis{0.284}{0.19}{0.50} & \cis{0.635}{0.20}{$3.3{\times}10^{3}$} & \cis{$1.8{\times}10^{13}$}{0.53}{$4.0{\times}10^{41}$} & \cis{$4.4{\times}10^{5}$}{0.86}{$1.8{\times}10^{13}$} & \cis{1.470}{0.00}{4.50} & -- & \cis{0.149}{0.08}{0.19} & \cis{0.112}{0.05}{0.14} \\[2pt]
\bottomrule
\end{tabular}
\caption{\textbf{Temperature sweep on the base model, $W_1$, against both fine-tuned variants at $T{=}1$.} Held-out families (12 families, 36 prompts). The $T{=}1$ column is the base model under the main evaluation settings. Logit KL is omitted because it does not depend on the sampling temperature.}
\label{tab:temperature-w1}
\end{table*}

No temperature reaches the combination of validity and distributional fit that fine-tuning obtains (Tables~\ref{tab:temperature-valid} and~\ref{tab:temperature-w1}). For most models, validity is flat up to $T=1.5$ and collapses beyond it. The Llama-3.2 models and GPT-OSS-20B already lose validity at $T=1.5$, and Llama-3.2-1B-it also loses it at low temperature (8.5\% at $T=0.5$). At $T=3$ no model keeps more than 30\% valid outputs, and at $T=5$ almost none remain. On every model except Gemma-3-1B-it, at least one fine-tuned variant is as valid as the base model at its best temperature, within 0.1 percentage points. On the larger checkpoints, raising $T$ lowers $W_1$ moderately (Qwen3-14B from 0.246 at $T=1$ to 0.209 at $T=3$), but both fine-tuned variants stay well below every temperature at which a non-negligible fraction of outputs is valid. Qwen3-0.6B is the closest case: its base model is at most 9.5\% valid at any temperature, and its $W_1$ at $T=0.5$ (0.244) is close to the soft-target value (0.237). The only base cell below the fine-tuned models is Qwen3-4B at $T=5$, where valid outputs are essentially absent (0.0\% after rounding). For the Llama-3.2, Gemma-3-1B-it, and GPT-OSS-20B base models, higher temperatures instead produce extreme in-support samples that inflate $W_1$ by many orders of magnitude, as discussed in Section~\ref{sec:results}.

\begin{table*}[t]
\centering
\footnotesize
\begin{tabular}{l r r c rr}
\toprule
Model & Prompts & Oracle $W_1$ & Modal best $T$ & Soft & Hard \\
\midrule
Qwen3-0.6B & 36 & \ci{0.182}{0.14}{0.22} & 1.0 & \ci{0.237}{0.16}{86.12} & \ci{0.209}{0.13}{0.30} \\
Qwen3-1.7B & 36 & \ci{0.296}{0.22}{0.40} & 2.0 & \ci{0.177}{0.11}{0.31} & \ci{0.166}{0.12}{0.31} \\
Qwen3-4B & 36 & \ci{0.226}{0.17}{0.27} & 3.0 & \ci{0.154}{0.12}{0.22} & \ci{0.163}{0.11}{0.22} \\
Qwen3-8B & 36 & \ci{0.213}{0.17}{0.33} & 3.0 & \ci{0.140}{0.08}{0.19} & \ci{0.120}{0.07}{0.16} \\
Qwen3-14B & 36 & \ci{0.133}{0.08}{0.17} & 3.0 & \ci{0.097}{0.07}{0.14} & \ci{0.116}{0.07}{0.14} \\
Gemma-3-1B-it & 36 & \ci{0.280}{0.20}{0.37} & 3.0 & \ci{0.242}{0.17}{0.36} & \ci{0.287}{0.19}{0.44} \\
Gemma-3-4B-it & 36 & \ci{0.290}{0.24}{0.36} & 3.0 & \ci{0.244}{0.14}{0.40} & \ci{0.183}{0.12}{0.29} \\
Gemma-3-12B-it & 36 & \ci{0.178}{0.14}{0.20} & 2.0 & \ci{0.135}{0.08}{0.22} & \ci{0.156}{0.08}{0.24} \\
Gemma-3-27B-it & 36 & \ci{0.172}{0.13}{0.19} & 3.0 & \ci{0.104}{0.08}{0.28} & \ci{0.120}{0.07}{0.26} \\
Llama-3.2-1B-it & 34 & \ci{18.733}{10.03}{30.34} & 0.5 & \ci{0.339}{0.17}{18.10} & \ci{0.260}{0.17}{0.41} \\
Llama-3.2-3B-it & 36 & \ci{0.429}{0.30}{0.53} & 0.5 & \ci{0.196}{0.11}{0.33} & \ci{0.186}{0.11}{0.29} \\
GPT-OSS-20B & 36 & \ci{0.190}{0.15}{0.25} & 1.0 & \ci{0.149}{0.08}{0.19} & \ci{0.112}{0.05}{0.14} \\
\bottomrule
\end{tabular}
\caption{\textbf{Per-prompt oracle temperature, the strongest form of the temperature baseline.} $W_1$ on the held-out families. Intervals are 95\% bootstrap intervals ($B=2000$), shown in smaller type after the point estimate. The oracle picks the best temperature per prompt with hindsight, so no single temperature can reach a lower $W_1$. Oracle is the median over prompts; Soft and Hard are the $W_1$ aggregates of Table~\ref{tab:main-ood}.}
\label{tab:temperature-oracle}
\end{table*}

Table~\ref{tab:temperature-oracle} gives the baseline its strongest form: for each prompt, we pick the temperature with the lowest $W_1$ in hindsight, which is not possible in practice. The oracle ignores validity, and it most often selects $T=3$, where validity has already fallen below 30\%. Even so, it stays above both fine-tuned variants on ten of the twelve models. The two exceptions are the two smallest models: Qwen3-0.6B, whose oracle (0.182) is below both variants, and Gemma-3-1B-it, whose oracle (0.280) falls between soft-target (0.242) and hard-target (0.287). For Qwen3-0.6B, whose base model is at most 9.5\% valid, the oracle $W_1$ is estimated from few samples. So temperature does not match fine-tuning, even when the best temperature is picked per prompt after the fact.

\clearpage
\section{Comparison with Zhang et al. (2024)}
\label{app:zhang}
\suppressfloats[t]  %

\citet{zhang2024forcing} is the closest training-time method to ours. They fine-tune a language model to minimize the sequence-level forward cross-entropy $-\sum_{y \in \mathcal{T}} p^*(y \mid x) \log p_\theta(y \mid x)$ toward a target distribution $p^*$ over a set $\mathcal{T}$ of valid outputs, and train on enumerated target sets such as numbers, dates, fruits, countries, occupations, and baby names. Their objective and our soft-target loss have the same minimizer. Summing the per-prefix KL of Equation~\ref{eq:cal-loss} over the trie, weighted by the target mass of each prefix, recovers the sequence-level forward KL, and Equation~\ref{eq:cal-loss-full} estimates that sum from a single sampled answer, divided by its length. What separates the two methods is the training data. We compare full recipes, each with its own data: their objective, data, and hyperparameters against our soft-target configuration. We reimplement their method on their released data, computing their objective exactly over the full target set, and evaluate in both directions: their recipe on our benchmark, and ours on theirs.

\subsection{Their Recipe on Our Benchmark}
\label{app:zhang-ourbench}

\begin{table*}[t]
\centering
\footnotesize
\setlength{\tabcolsep}{3pt}
\begin{tabular}{ll r rrr}
\toprule
Model & Condition & Steps & Logit KL $\downarrow$ & $W_1$ $\downarrow$ & Valid (\%) $\uparrow$ \\
\midrule
Qwen3-1.7B & base & -- & \ci{6.524}{4.205}{9.657} & \ci{0.356}{0.260}{0.477} & \ci{84.8}{65.8}{98.7} \\
 & soft@189 & 189 & \ci{0.433}{0.292}{0.589} & \ci{0.177}{0.107}{0.313} & \ci{92.9}{85.1}{98.6} \\
 & soft@2016 & 2016 & \ci{0.536}{0.347}{0.811} & \ci{0.172}{0.121}{0.304} & \ci{95.2}{86.3}{99.7} \\
 & Zhang all & 1300 & \ci{2.251}{1.087}{4.031} & \ci{0.275}{0.198}{0.724} & \ci{91.3}{75.1}{99.9} \\
\midrule
Qwen3-14B & base & -- & \ci{3.272}{2.132}{4.862} & \ci{0.246}{0.225}{0.316} & \ci{95.5}{84.5}{100.0} \\
 & soft@189 & 189 & \ci{0.314}{0.182}{0.442} & \ci{0.097}{0.071}{0.139} & \ci{99.8}{99.5}{100.0} \\
 & soft@2016 & 2016 & \ci{0.323}{0.189}{0.451} & \ci{0.129}{0.058}{0.153} & \ci{99.8}{99.7}{99.9} \\
 & Zhang all & 1300 & \ci{1.799}{1.024}{2.660} & \ci{0.196}{0.120}{0.297} & \ci{95.2}{84.4}{100.0} \\
\midrule
Gemma-3-12B-it & base & -- & \ci{6.236}{4.394}{8.244} & \ci{0.247}{0.206}{0.287} & \ci{94.9}{87.7}{99.4} \\
 & soft@189 & 189 & \ci{0.407}{0.282}{0.514} & \ci{0.135}{0.078}{0.218} & \ci{99.6}{98.8}{100.0} \\
 & soft@2016 & 2016 & \ci{0.351}{0.233}{0.462} & \ci{0.147}{0.078}{0.190} & \ci{98.9}{97.1}{99.9} \\
 & Zhang all & 1300 & \ci{2.229}{1.879}{2.710} & \ci{0.223}{0.154}{0.311} & \ci{89.4}{73.9}{99.8} \\
\midrule
GPT-OSS-20B & base & -- & \ci{1.498}{0.967}{1.948} & \ci{0.635}{0.200}{$3.3{\times}10^{3}$} & \ci{99.0}{98.0}{99.4} \\
 & soft@189 & 189 & \ci{0.497}{0.305}{0.704} & \ci{0.149}{0.083}{0.191} & \ci{99.8}{99.3}{100.0} \\
 & soft@2016 & 2016 & \ci{0.484}{0.284}{0.691} & \ci{0.099}{0.067}{0.186} & \ci{99.9}{99.5}{100.0} \\
 & Zhang all & 1300 & \ci{1.336}{0.828}{1.841} & \ci{0.559}{0.166}{1.459} & \ci{96.4}{94.6}{97.8} \\
\bottomrule
\end{tabular}
\caption{\textbf{Their recipe on our benchmark, 12 held-out families.} Zhang all is their full recipe (objective, data, and hyperparameters) trained on all six of their categories. Point estimate with its 95\% hierarchical bootstrap interval ($B=2000$; families, then distributions within family) in smaller type. Intervals resample evaluation prompts, not training runs.}
\label{tab:zhang-ourbench}
\end{table*}

On our benchmark, their recipe improves on the base model but remains well above ours (Table~\ref{tab:zhang-ourbench}). Logit KL falls from 6.52 to 2.25 on Qwen3-1.7B, from 3.27 to 1.80 on Qwen3-14B, and from 6.24 to 2.23 on Gemma-3-12B-it, but remains 5.2 to 5.7 times above soft-target fine-tuning on these three models. On GPT-OSS-20B their recipe stays within the interval of the base model (1.34 against 1.50), 2.7 times above ours. $W_1$ follows the same ordering, with their recipe between the base model and ours on every model. The gap is not a matter of budget. Their recipe takes one optimizer step per training prompt and receives 1300 steps, about seven times our 189, and our soft-target run at 2016 steps performs similarly to our run at 189. We posit that the gap comes from the training distributions: every target set in their released data is uniform, so their recipe never trains on a shaped target, while our benchmark asks for a specific shape at prompt-specified parameters.

\subsection{Our Recipe on Their Benchmark}
\label{app:zhang-theirbench}

\begin{table*}[t]
\centering
\footnotesize
\setlength{\tabcolsep}{3pt}
\begin{tabular}{ll rrrr}
\toprule
Model & Condition & valid rate & $H_{\mathrm{norm}}$ & $D_{\mathrm{KL}}$ & coverage \\
\midrule
Qwen3-1.7B & base & \ci{0.752}{0.42}{1.00} & \ci{0.242}{0.13}{0.34} & \ci{11.580}{7.90}{15.75} & \ci{0.266}{0.12}{0.43} \\
 & soft@189 & \ci{0.656}{0.35}{0.92} & \ci{0.624}{0.46}{0.77} & \ci{3.701}{1.89}{5.34} & \ci{0.615}{0.37}{0.84} \\
 & soft@2016 & \ci{0.694}{0.40}{0.96} & \ci{0.596}{0.45}{0.74} & \ci{6.522}{3.33}{10.12} & \ci{0.586}{0.35}{0.81} \\
 & hard@1988 & \ci{0.711}{0.39}{0.96} & \ci{0.614}{0.47}{0.77} & \ci{4.510}{2.14}{7.12} & \ci{0.611}{0.36}{0.84} \\
 & Zhang LOO & \ci{0.722}{0.44}{0.96} & \ci{0.696}{0.61}{0.79} & \ci{2.688}{1.28}{4.14} & \ci{0.667}{0.42}{0.89} \\
 & Zhang all & \ci{0.919}{0.79}{0.99} & \ci{0.980}{0.96}{0.99} & \ci{0.164}{0.02}{0.42} & \ci{0.974}{0.92}{1.00} \\
\midrule
Qwen3-14B & base & \ci{0.795}{0.48}{1.00} & \ci{0.297}{0.21}{0.40} & \ci{10.224}{7.17}{12.84} & \ci{0.292}{0.15}{0.45} \\
 & soft@189 & \ci{0.651}{0.35}{0.91} & \ci{0.830}{0.72}{0.94} & \ci{1.942}{0.60}{3.35} & \ci{0.756}{0.49}{0.96} \\
 & soft@2016 & \ci{0.702}{0.40}{0.95} & \ci{0.817}{0.70}{0.93} & \ci{2.200}{0.70}{3.83} & \ci{0.739}{0.48}{0.95} \\
 & hard@1988 & \ci{0.696}{0.40}{0.94} & \ci{0.830}{0.71}{0.94} & \ci{2.030}{0.63}{3.53} & \ci{0.747}{0.48}{0.95} \\
 & Zhang LOO & \ci{0.722}{0.45}{0.94} & \ci{0.810}{0.70}{0.92} & \ci{2.089}{0.71}{3.39} & \ci{0.729}{0.48}{0.93} \\
 & Zhang all & \ci{0.984}{0.97}{1.00} & \ci{0.984}{0.97}{0.99} & \ci{0.050}{0.01}{0.09} & \ci{0.994}{0.98}{1.00} \\
\midrule
Gemma-3-12B-it & base & \ci{0.910}{0.78}{1.00} & \ci{0.195}{0.10}{0.29} & \ci{36.714}{31.22}{42.55} & \ci{0.247}{0.11}{0.39} \\
 & soft@189 & \ci{0.756}{0.45}{0.99} & \ci{0.742}{0.63}{0.85} & \ci{2.770}{1.12}{4.43} & \ci{0.667}{0.43}{0.88} \\
 & soft@2016 & \ci{0.778}{0.46}{1.00} & \ci{0.661}{0.52}{0.80} & \ci{3.842}{1.62}{6.49} & \ci{0.588}{0.33}{0.83} \\
 & hard@1988 & \ci{0.791}{0.49}{1.00} & \ci{0.678}{0.53}{0.84} & \ci{3.460}{1.15}{5.84} & \ci{0.619}{0.37}{0.88} \\
 & Zhang LOO & \ci{0.731}{0.43}{0.98} & \ci{0.815}{0.72}{0.91} & \ci{2.141}{0.72}{3.66} & \ci{0.739}{0.50}{0.93} \\
 & Zhang all & \ci{0.989}{0.98}{1.00} & \ci{0.986}{0.98}{0.99} & \ci{0.033}{0.01}{0.07} & \ci{0.991}{0.98}{1.00} \\
\midrule
GPT-OSS-20B & base & \ci{0.725}{0.41}{0.98} & \ci{0.565}{0.47}{0.67} & \ci{4.889}{2.54}{7.50} & \ci{0.559}{0.33}{0.76} \\
 & soft@189 & \ci{0.649}{0.36}{0.89} & \ci{0.807}{0.69}{0.91} & \ci{4.616}{2.12}{7.26} & \ci{0.742}{0.49}{0.95} \\
 & soft@2016 & \ci{0.556}{0.30}{0.79} & \ci{0.755}{0.65}{0.85} & \ci{9.165}{5.68}{12.24} & \ci{0.696}{0.44}{0.91} \\
 & hard@1988 & \ci{0.604}{0.28}{0.91} & \ci{0.770}{0.63}{0.90} & \ci{10.312}{6.67}{13.68} & \ci{0.647}{0.37}{0.90} \\
 & Zhang LOO & \ci{0.696}{0.39}{0.96} & \ci{0.786}{0.69}{0.89} & \ci{2.229}{0.73}{3.96} & \ci{0.718}{0.44}{0.94} \\
 & Zhang all & \ci{0.814}{0.55}{0.98} & \ci{0.947}{0.89}{0.99} & \ci{0.529}{0.06}{1.32} & \ci{0.892}{0.71}{1.00} \\
\bottomrule
\end{tabular}
\caption{\textbf{Our recipe on their benchmark, averaged over their six categories.} Valid rate is the fraction of samples inside the target set $\mathcal{T}$; $H_{\mathrm{norm}}$ is the entropy of the model restricted to $\mathcal{T}$ divided by $\log|\mathcal{T}|$ (1 is uniform); $D_{\mathrm{KL}}$ is the exact KL from their target to the model; coverage is the fraction of $\mathcal{T}$ sampled at least once. Intervals are 95\% bootstrap intervals ($B=2000$), shown in smaller type after the point estimate. Intervals resample categories, then tasks within category. Zhang all trains and evaluates on the same prompts and only confirms that our reimplementation fits their objective; it is not a comparator. Zhang LOO is the only Zhang condition not trained on the evaluation task.}
\label{tab:zhang-theirbench}
\end{table*}

On their benchmark (Table~\ref{tab:zhang-theirbench}), our checkpoints never train on any of their tasks and still raise normalized entropy well above the base model: from 0.24 to 0.62 on Qwen3-1.7B, from 0.30 to 0.83 on Qwen3-14B, from 0.20 to 0.74 on Gemma-3-12B-it, and from 0.57 to 0.81 on GPT-OSS-20B, and coverage rises with it. The valid rate of each of our fine-tuned conditions falls below the base model, because our models place more mass outside the enumerated target sets. On the two categories whose target lists are incomplete, this extra mass is diversity rather than error. On the four complete ones, it is a real cost. The exact KL from their target to the model falls below the base model everywhere except for GPT-OSS-20B at the longer budgets (soft-target at 2016 steps and hard-target), which exceed the base model. The Zhang all row ($H_{\mathrm{norm}}$ of 0.95 to 0.99) confirms that our reimplementation fits their objective.

\begin{table*}[t]
\centering
\footnotesize
\setlength{\tabcolsep}{3pt}
\begin{tabular}{ll rrrrrr r}
\toprule
Model & Side & numbers & \shortstack{day-\\date} & fruits & \shortstack{continent-\\countries} & \shortstack{occu-\\pations} & \shortstack{baby-\\names} & Mean \\
\midrule
Qwen3-1.7B & Zhang LOO & 0.887 & 0.754 & 0.643 & 0.722 & 0.545 & 0.623 & \ci{0.696}{0.61}{0.79} \\
 & ours, soft@189 & 0.895 & 0.362 & 0.489 & 0.712 & 0.689 & 0.598 & \ci{0.624}{0.46}{0.77} \\
\midrule
Qwen3-14B & Zhang LOO & 0.970 & 0.987 & 0.730 & 0.849 & 0.644 & 0.676 & \ci{0.810}{0.70}{0.92} \\
 & ours, soft@189 & 0.973 & 0.986 & 0.836 & 0.876 & 0.662 & 0.647 & \ci{0.830}{0.72}{0.94} \\
\midrule
Gemma-3-12B-it & Zhang LOO & 0.951 & 0.977 & 0.727 & 0.843 & 0.727 & 0.667 & \ci{0.815}{0.72}{0.91} \\
 & ours, soft@189 & 0.915 & 0.919 & 0.651 & 0.726 & 0.611 & 0.630 & \ci{0.742}{0.63}{0.85} \\
\midrule
GPT-OSS-20B & Zhang LOO & 0.923 & 0.965 & 0.720 & 0.813 & 0.668 & 0.631 & \ci{0.786}{0.69}{0.89} \\
 & ours, soft@189 & 0.933 & 0.990 & 0.667 & 0.881 & 0.735 & 0.634 & \ci{0.807}{0.69}{0.91} \\
\bottomrule
\end{tabular}
\caption{\textbf{Leave-one-out head-to-head per category, $H_{\mathrm{norm}}$ (higher is better).} Each of their folds trains on the five sibling categories and is scored only on the one it held out; ours never trains on any of these tasks. The Mean column carries a 95\% bootstrap interval ($B=2000$; categories, then tasks within category) in smaller type.}
\label{tab:zhang-loo-categories}
\end{table*}

Their own generalization test holds out one category at a time: each fold trains on five of the six categories and is scored on the sixth. This is the only comparison in which neither side has trained on the evaluation task, and the setup favors their side in three ways. Each of their six numbers comes from a separate checkpoint, trained on five sibling categories that share the task type and output format of the held-out one. Each fold uses 900 to 1250 optimizer steps, against our 189. Ours is a single checkpoint trained only on numeric parametric distributions. Under these conditions, mean $H_{\mathrm{norm}}$ favors their recipe on Qwen3-1.7B (0.696 against 0.624) and Gemma-3-12B-it (0.815 against 0.742), and ours on Qwen3-14B (0.830 against 0.810) and GPT-OSS-20B (0.807 against 0.786) (Table~\ref{tab:zhang-loo-categories}). On \textsc{numbers}, the only category adjacent to our training domain, the two sides are within 0.04 on every model. Our largest deficit is \textsc{day-date} on Qwen3-1.7B (0.362 against 0.754), which rises to 0.986 on Qwen3-14B with identical prompts. Taken together, their recipe is the stronger choice for fitting listed sets of answers, and ours for sampling from specified parametric distributions, where theirs falls well short even after seven times as many optimizer steps.

\end{document}